\documentclass[preprint]{article}

\usepackage{neurips_2026}
\usepackage[utf8]{inputenc}
\usepackage{newunicodechar}
\newunicodechar{−}{-}
\usepackage{pifont}

\usepackage[utf8]{inputenc} 
\usepackage[T1]{fontenc}    
\usepackage{xcolor}         
\usepackage{amsmath,amsfonts}
\usepackage{algorithm}
\usepackage{booktabs}
\usepackage{multirow}
\usepackage{algorithmic}
\usepackage{graphicx}
\usepackage{wrapfig}
\usepackage{mdframed}
\usepackage{makecell}
\usepackage{tikz}
\usetikzlibrary{fit,positioning}
\usepackage{arydshln}
\usepackage[most]{tcolorbox}
\tcbuselibrary{skins}
\usepackage{listings}
\usepackage{multicol}
\usepackage{array}
\usepackage{colortbl}
\usepackage{subcaption}
\usepackage{adjustbox}
\usepackage[normalem]{ulem}
\usepackage{booktabs}
\usepackage{caption}
\usepackage{nicefrac}
\usepackage{microtype}
\usepackage{colortbl}
\usepackage[table,xcdraw]{xcolor}
\usepackage{url}
\usepackage{natbib}
\usepackage{hyperref}
\usepackage{enumitem}
\usepackage{tabularx}
\usepackage{booktabs}
\usepackage{makecell}
\usepackage{colortbl}
\definecolor{LightRed}{rgb}{1,0.8,0.8}
\definecolor{LightGreen}{rgb}{0.8,1,0.8}

\newcolumntype{C}[1]{>{\centering\arraybackslash}p{#1}}

\definecolor{lightgray}{gray}{0.88}

\tcbset{
    codebox/.style={
        enhanced,
        colback=white,
        colframe=black,
        boxrule=0.5pt,
        sharp corners,
        fontupper=\ttfamily\footnotesize,
        breakable
    },
    getgraphs/.style={
        enhanced,
        colframe=purple,
        colback=white,
        boxrule=0.7pt,
        borderline={0.7pt}{0pt}{dashed}
    },
    obsbox/.style={
        enhanced,
        colframe=blue,
        colback=white,
        boxrule=0.7pt,
        borderline={0.7pt}{0pt}{dashed}
    },
    actbox/.style={
        enhanced,
        colframe=green!60!black,
        colback=white,
        boxrule=0.7pt,
        borderline={0.7pt}{0pt}{dashed}
    },
    attrbox/.style={
        enhanced,
        colframe=yellow,
        colback=white,
        boxrule=0.7pt,
        borderline={0.7pt}{0pt}{dashed}
    },
    funcbox/.style={
        enhanced,
        colframe=pink,
        colback=white,
        boxrule=0.7pt,
        borderline={0.7pt}{0pt}{dashed}
    }
}
\title{RL Agents Navigating Continuous Latent Actions for Scalable Graph Combinatorial Optimization}
\title{Projection is All You Need: Continuous Action Spaces for Better RL in Graph Combinatorial Optimization}
\title{Bypassing Discrete Actions via Latent Projection in Deep RL for Graph Combinatorial Optimization}
\title{RL Agents Navigating Action Embedding Manifolds for Graph-Based Combinatorial Optimization}
\title{Deep RL Agents Navigating Continuous Action Manifolds for Graph Combinatorial Optimization}
\title{Stop Choosing, Start Projecting: RL-based Latent Actions for Graph Combinatorial Optimization}
\title{Projecting Continuous Latent Actions with Deep RL to Solve Graph Combinatorial Optimization Problems}
\title{Continuous Latent Action Projection in Reinforcement learning for Graph Combinatorial Optimization}
\title{Projecting Latent Actions with Reinforcement Learning for Graph Combinatorial Optimization}
\title{Latent Projections: Unlocking Generalization and Scalability in Graph Combinatorial Optimization}
\title{Latent Projections: Unlocking Generalizable and Scalable Graph Combinatorial Optimization}
\title{Generalizable and Scalable Graph Combinatorial Optimization via Projection  Agents for Reinforcement Learning}

\title{P-Agents: Generalizable and Scalable Graph Combinatorial Optimization for Reinforcement Learning}
\title{Projecting Latent RL Actions: Towards Generalizable and Scalable Graph Combinatorial Optimization}

\author{
  Franco Terranova \\
  Université de Lorraine, CNRS, Inria,
LORIA \\
  \texttt{franco.terranova@inria.fr} \\
  \And
  Guillermo Bernardez \\
  University of California Santa Barbara \\
  \texttt{guillermo\_bernardez@ucsb.edu} \\
  \And
  Albert Cabellos-Aparicio \\
  Universitat Politècnica de Catalunya \\
  \texttt{alberto.cabellos@upc.edu} \\
  \And
  Nina Miolane \\
  University of California Santa Barbara \\
  \texttt{ninamiolane@ucsb.edu} \\
  \And
  Abdelkader Lahmadi \\
  Université de Lorraine, CNRS, Inria,
LORIA \\
  \texttt{abdelkader.lahmadi@loria.fr}
}

\begin{document}

\maketitle

\begin{abstract}
Graph combinatorial optimization (GCO) has attracted growing interest, as many NP-hard problems naturally admit graph formulations, yet their combinatorial explosion renders exact methods computationally intractable.
Recent advances in Reinforcement Learning (RL) combined with Graph Neural Networks (GNNs) have significantly improved learning-based GCO solvers. However, existing approaches face limitations in both generalization across diverse graph instances and computational scalability as action spaces grow. To address both challenges, we introduce \textit{projection agents}, a GCO methodology built on the Wolpertinger framework and adapted to a shared GNN-based latent space for observations and actions. These agents are trained to solve GCO problems by predicting a desired latent action in a single forward pass, which is subsequently decoded into a valid discrete action in the graph using the latent space. 
Across diverse benchmarks, our approach achieves up to 6.41$\times$ faster inference and up to 75\% better generalization than existing solutions using a simple nearest-neighbor decoding, while opening the door to strong RL performance in super-linear decision spaces with multiple interdependent variables. 
Finally, we release \texttt{LaGCO-RL}, a Python library that automates latent action-space construction and supports existing RL-GCO solutions, promoting reproducibility and adaptation to new GCO benchmarks.
\end{abstract}

\section{Introduction}
Solving graph-based combinatorial optimization (GCO) problems has attracted increasing interest, as many NP-hard problems across several domains can naturally be modeled as graphs \citep{darvariu2024graphreinforcementlearningcombinatorial}. In the seminal work of \cite{Karp1972} introducing 21 NP-complete problems (e.g., Traveling Salesman Problem \citep{TSP}), nearly half correspond to versions of graph optimization problems, and most of them can also be formulated on graphs \citep{cappart2022combinatorialoptimizationreasoninggraph}. However, due to the combinatorial nature of these problems, where the number of possible solutions grows rapidly with graph size, exact methods become computationally intractable \citep{DRLCOTSP}.

As a result, heuristic methods are widely employed, trading optimality for computational efficiency \citep{vesselinovaLearingCOGraphsNetworking}. However, traditional heuristics typically rely on domain expertise, handcrafted rules, and iterative trial-and-error \citep{COMBGCNTree}, often demanding substantial human effort to redesign and adapt them for each variation of the problem. More recently, machine learning (ML) approaches have emerged as a promising alternative for data-driven heuristic discovery \citep{bengioMLCO, cappart2022combinatorialoptimizationreasoninggraph, RlforCOSurvey}. In particular, reinforcement learning (RL) methods \citep{cappart2020combiningreinforcementlearningconstraint} have attracted increasing attention because they do not require labeled samples, which are costly to obtain for large instances \citep{Peng2021GraphLearning, variationalAnnealing}, and construct solutions autoregressively through sequential decision-making, thereby mirroring the process employed by traditional GCO solvers \citep{darvariu2024graphreinforcementlearningcombinatorial}. RL has been used both for direct heuristic approximation and to accelerate exact solvers, e.g., by guiding branch-and-bound in constraint programming \citep{cappart2020combiningreinforcementlearningconstraint}. Therefore, improving RL for GCO (RL-GCO) is important to advance both directions.

A key advancement of RL-GCO is the use of Graph Neural Networks (GNNs) \citep{GNNModel} to learn fixed-dimensional, transferable representations of graph structures, which have been widely adopted to enhance the encoding of graph-based observation spaces \citep{graphPooling}.
However, standard RL approaches typically rely on \textit{discrete} action spaces, whose semantics and dimensionality are tightly coupled to individual graph instances. Consequently, learned action mappings are not readily transferable across graphs: the same discrete action identifier may refer to different graph entities \citep{UsingContinuousDiscreteProblems, chen2021learningactiontransferablepolicyaction}, unless the policy is explicitly permutation-invariant and can infer the underlying semantics of actions from the input graph. Moreover, discrete action spaces limit scalability, as the output dimensionality of function approximators grows with the number of available actions \citep{dulacarnold2016deepreinforcementlearninglarge}. 

To address this, recent GNN-based approaches introduce action embeddings derived from graph components and evaluate candidate actions via a Q-function $Q(s,a)$, as in \texttt{S2V-DQN} \citep{learningCOoverGraphs}. However, iterating over all candidate actions incurs a cost that grows linearly with the action space, making runtime dependent on graph size and limiting scalability \citep{GCOMB}.
In particular, this computational cost is exacerbated in real-world GCO tasks with structured, high-dimensional decision variables, where the number of actions can grow super-linearly with graph size. A prominent example is traffic engineering, where decisions over graph paths induce an exponential explosion in the action space \citep{DeepRLforTrafficEngineering}. As a result, evaluating a Deep Neural Network (DNN) a number of times proportional to the action space size quickly becomes computationally prohibitive.
This issue is amplified in dynamic settings, such as GCO applications in communication networks or data center infrastructures, where graph representations evolve continuously, and decisions must be re-computed repeatedly and under strict latency constraints \citep{vesselinovaLearingCOGraphsNetworking}. 

\paragraph{Contributions.}
To enhance the applicability of RL-GCO heuristics to large and diverse graphs across a broader range of tasks, we present the following key contributions:

\textbf{(1) Projection agents for scalable and generalizable RL-GCO.} 
We introduce \textit{projection agents}, which operate directly in continuous GNN-based observation and action embedding spaces. 
Inspired by the Wolpertinger architecture \citep{dulacarnold2016deepreinforcementlearninglarge}, the agent predicts a desired action embedding in a single forward pass and projects it onto a valid discrete action through a Nearest Neighbor (NN) search. This unified approach achieves up to 6.41$\times$ faster inference than iterative Q-value methods, while improving relative generalization by up to 75\% on unseen instances.

 
\textbf{(2) Unlocking RL-GCO for problems with complex and super-linear decision variables.}
We introduce a framework for constructing structured action embeddings via (i) semantic representations of individual decision components and (ii) composite embeddings capturing their interactions. This supports realistic super-linear decision spaces involving multiple interdependent variables, as found in modern real-world GCO tasks (e.g., traffic engineering, where selections involve paths and edges).

\textbf{(3) Self-supervised representation learning framework to better compare RL-GCO.}
We leverage self-supervised GNN representations \citep{GNNUnsupervised} to build aligned observation–action spaces for RL agents. By decoupling representation learning from policy learning, we enable fair comparison across RL methods within a shared embedding space.


\textbf{(4) Unified Python library}. 
We provide \texttt{LaGCO-RL}\footnote{Code:
\url{https://github.com/terranovafr/LaGCO-RL}}, a modular library for automated \textbf{L}atent \textbf{a}ction-space construction in \textbf{GCO} with \textbf{RL}.
The data repository, including model checkpoints, generated scenarios, complete experimental results, and hyperparameter configurations, is publicly available and attached to this paper.\footnote{Data repository:
\url{https://zenodo.org/records/20019625}}
The library enables rapid adaptation to new tasks by requiring only environment-specific logic, while promoting reproducibility of the proposed methodology~\citep{RLOnGraphsSurvey}. The implementation natively supports the proposed projection agent alongside existing RL baselines and is built on standardized Stable-Baselines3 (SB3)~\citep{stable-baselines3} implementations.





\section{Related work}
\label{chap:related_work} 

\begin{figure}
    \centering
    \includegraphics[width=\linewidth]{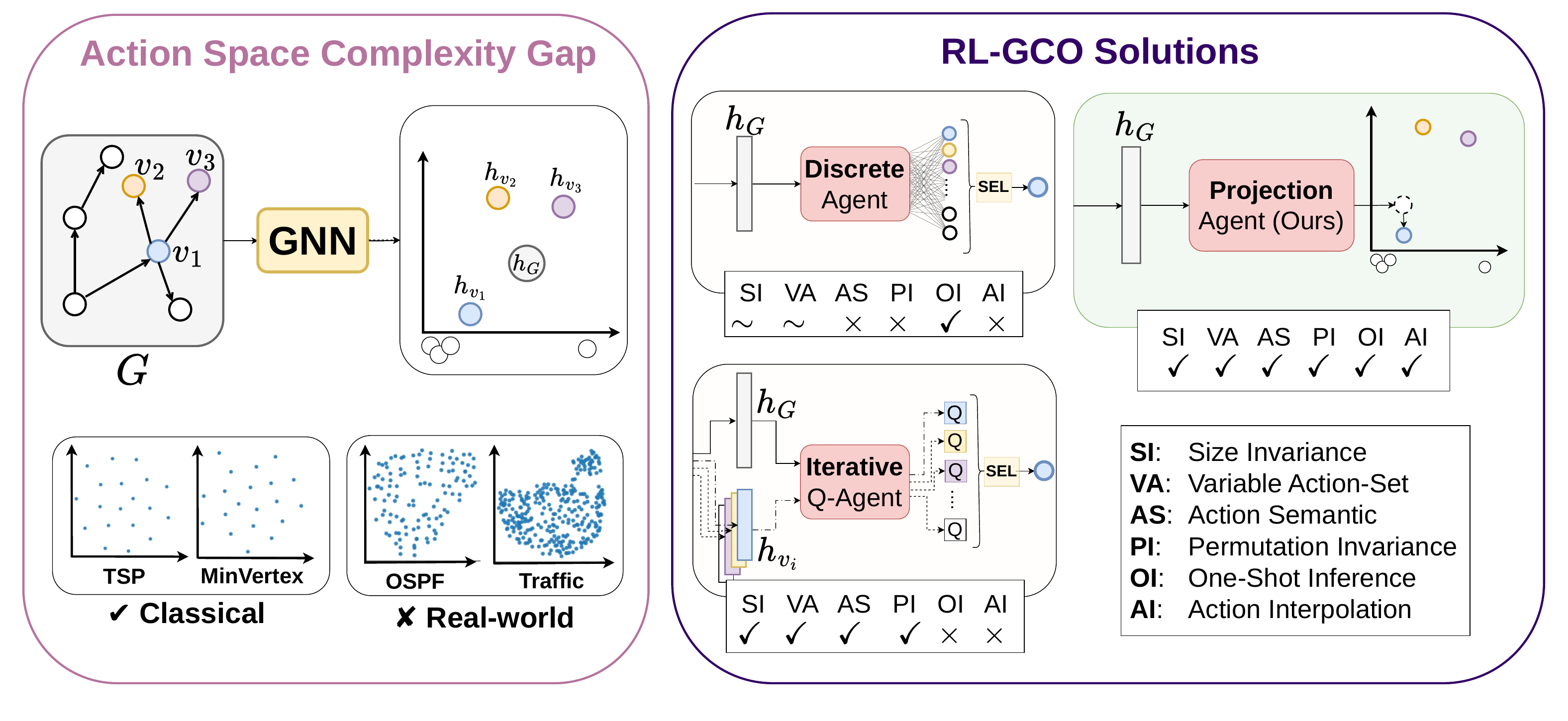}
     \caption{Overview of related work across: (1) action-space complexity, where real-world problems require richer decisions beyond node embeddings \(h_i\); and (2) comparison of our projection method and others based on: \textit{graph-size invariance}, support for \textit{dynamic action sets}, \textit{action semantics} (vs.\ flat indices), \textit{permutation invariance}, \textit{one-shot inference} (one forward pass for a decision), and support for \textit{action interpolation}. \(\checkmark\), \(\sim\), and \(\times\) denote full, partial, and no support, respectively.}
    \label{fig:related_work}
    \vspace{-0pt}
\end{figure}


This section reviews related work along two main axes illustrated in Figure~\ref{fig:related_work}: \textit{(i)} the action space complexity of existing benchmark GCO problems, and \textit{(ii)} the ability of current RL-based solutions to effectively handle the typical action space challenges of GCO. An extended related work against other ML heuristic approaches (including other paradigms beyond RL) is provided in Appendix \ref{chap:extended_related_work}.

\paragraph{Action space complexity gap.} Prior work on RL-GCO has largely focused on simplified benchmarks with limited decision complexity. Seminal studies \citep{learningCOoverGraphs, manchanda2020learningheuristicslargegraphs, COMBGCNTree} consider classical problems such as the Traveling Salesman Problem (\textit{TSP}) and Minimum Vertex Cover (\textit{MinVertex}) \citep{MVC} (see Figure~\ref{fig:related_work}'s action space instances coming from our experimental study), where graph representations remain basic: edge features are often reduced to scalar weights or ignored, and node features are limited to simple indicators (e.g., visited flags).
Moreover, these approaches predominantly treat nodes as the sole decision variables, resulting in action spaces that scale linearly with graph size and exhibit relatively simple structure. Under such settings, GNN-based node embeddings are typically sufficient to produce well-separated manifolds of actions, facilitating efficient RL learning.
In contrast, real-world applications such as \textit{OSPF} engineering \citep{MAGNNETO} or \textit{Traffic} routing \citep{DeepRLforTrafficEngineering} involve more complex decision variables (e.g., paths), resulting in action spaces that grow super-linearly not only in size but also in structural complexity. 
This discrepancy highlights the gap in the resulting action space complexity between commonly used and real-world benchmarks.

\paragraph{RL-GCO solutions.} Existing RL approaches fall short of key operational requirements for generalization and scalability in real-world GCO (Fig.~\ref{fig:related_work}).
\textit{Discrete} methods rely on padding-based action encodings to handle variable graph sizes, assigning one DNN output unit per action and treating graph elements as index-based entities \citep{terranova2025rlAttackPath}. This yields instance-specific policies that do not capture action semantics and transfer poorly across graphs. Such approaches provide only partial size invariance, limited by the padding dimension, and require invalid action masking for varying actions. While they also lack permutation invariance and interpolation, they remain computationally efficient due to a one-shot inference. 
On the other hand, \textit{iterative} latent action-value methods embed both states and actions, typically using GNNs, and evaluate them sequentially via a learned Q-function \citep{learningCOoverGraphs, GCOMB}. This provides semantic representations, permutation and size invariance (from GNN features), and natural support for varying action sets (only valid actions can be iterated). However, this approach requires a separate forward pass per action, resulting in costs that scale with the action space, which itself can grow super-linearly with graph size in some benchmarks. Moreover, the lack of a continuous representation prevents action interpolation.

\paragraph{Adapting Wolpertinger to GCO.} By contrast, our proposed \textit{projection} approach adapts Wolpertinger~\citep{dulacarnold2016deepreinforcementlearninglarge} to predict a target action embedding in a latent space (created leveraging GNNs), with action selection delegated to a dedicated decoding strategy. 
It combines semantic embeddings with size and permutation invariance (via GNNs), naturally handles varying action sets (restricting decoding only to valid ones), and enables one-shot DNN inference followed by a fast decoding (e.g., NN search efficiently indexed for sub-linear retrieval). Unlike other methods, it also supports smooth interpolation across actions by projecting vectors over the full manifold.

Previous applications of Wolpertinger have primarily considered benchmarks such as discretized control, GridWorld, and recommendation systems \citep{dulacarnold2016deepreinforcementlearninglarge,chen2021learningactiontransferablepolicyaction,chandak2019learningactionrepresentationsreinforcement,akkerman2024dynamic}, where action embeddings remain fixed within an episode. In GCO, however, action representations are dynamic: as graph modifications propagate through the GNN, component embeddings evolve, so the same graph entity may have different action embeddings depending on preceding actions. To our knowledge, this is the first application of Wolpertinger to GCO and, more broadly, to a dynamic latent action space, requiring specialized methods for efficient learning.

\section{Methodology}
After a brief introduction of a novel problem formulation, this section introduces our core contributions: the framework for self-supervised embedding learning, the construction of latent spaces, and the design of projection agents. We conclude with an overview of the supporting \texttt{LaGCO-RL} library.



\subsection{Problem formulation}
\label{chap:problem}
We formulate GCO as a sequential learning problem aimed at constructing a solution on a graph \( G = (V, E, V_f, E_f) \), where \(V\) and \(E\) denote the sets of nodes and edges, \(V_f \in \mathbb{R}^{|V|\times d_v}\) encodes node features, and \(E_f \in \mathbb{R}^{|E|\times d_e}\) encodes edge features (with $|V|$ and $|E|$ representing the cardinality of the node and edge sets, respectively, and $d_v$ and $d_e$ their corresponding feature dimensions). Let \(\mathcal{C}(G)\) denote the decision space of admissible graph substructures (e.g., node subsets, edge subsets, paths, or induced subgraphs). This formulation generalizes beyond single-node decisions, enabling the selection of arbitrary substructures and thus yielding a richer and more challenging action space.
In the autoregressive setting, a solution is constructed sequentially as
\(
\mathbf{S} = (S_1, \dots, S_T), 
\text{ where each decision }  S_t \in \mathcal{C}(G)
\) may depend on previous selections. The full sequence \(\mathbf{S}\) defines a feasible solution to the GCO problem with objective:
\begin{equation}
\label{eq:obj_min}
    \min_{\mathbf{S} \in \mathcal{C}(G)^T} F(G, \mathbf{S}) 
\quad \text{s.t.} \quad H(G, \mathbf{S}) \le 0,
\end{equation}
where \(F\) measures the global cost and \(H\) encodes optional feasibility constraints. This formulation captures realistic GCO settings in which both \(F\) and \(H\) may depend on node features, edge features, and higher-order structural interactions. 
This autoregressive formulation naturally induces a Markov Decision Process (MDP) \citep{puterman1994markov} with a discrete action space:
\begin{itemize}[leftmargin=*,itemsep=1pt,parsep=0pt,topsep=0pt]
    \item \textbf{Observation.} Each observation \(o_t \in \mathcal{O}\) represents the intermediate solution at step \(t\), defined as \(o_t = (G, S_1, \dots, S_{t-1})\).
    \item \textbf{Action.} An action \(a_t \in \mathcal{A}(o_t) \subseteq \mathcal{C}(G)\) selects a valid graph substructure that extends the current partial solution.
    \item \textbf{Transition.} The transition deterministically appends the selected substructure to the solution while enforcing constraints \(H\); the episode terminates when a solution is formed, or a cutoff is reached.
    \item \textbf{Reward.} The reward is derived from \(F(G, \mathbf{S})\), so maximizing return is equivalent to minimizing the combinatorial objective, with penalties induced by constraint violations \(H\).
\end{itemize}
For example, in \textit{TSP}, the observation is a city graph with binary indicators of the current partial tour. The action selects an unvisited city, transitions update the tour, and the reward favors shorter tours.

\subsection{Node representation learning}
\label{chap:GNNRepresentation}
We pre-train a GNN encoder via self-supervised learning to produce node-level embeddings that serve as shared blocks for building both observations and actions of a given RL task. 
This design \textit{(i)} decouples representation learning from action selection, enabling fair comparison across RL-GCO methods through a shared embedding space; \textit{(ii)} improves sample efficiency by learning only the observation–action mapping with a lightweight controller \citep{WorldModels}, avoiding joint representation learning; and \textit{(iii)} stabilizes training by keeping action embeddings stable, avoiding additional representational drift on top of the embedding changes induced by graph evolution.

\begin{wrapfigure}{r}{0.5\linewidth}
    \centering
    \vspace{-10pt}
    \includegraphics[width=0.85\linewidth]{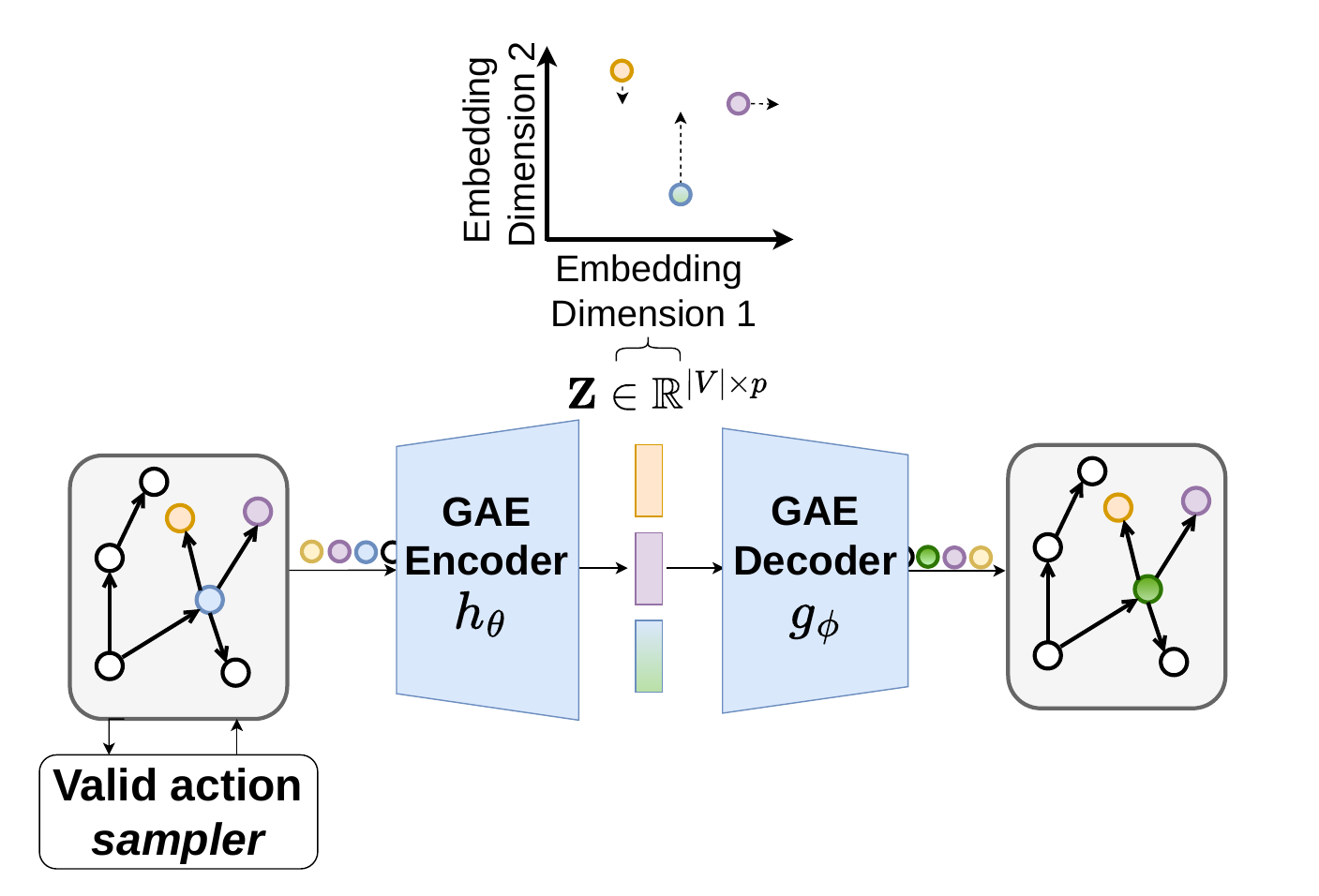}
    \caption{Joint GNN encoder-decoder framework leveraging valid action-sampled snapshots for node representation learning.}
    \label{fig:GAE}
    \vspace{-10pt}
\end{wrapfigure}

We learn node-level representations using a Graph Auto-Encoder (GAE) \citep{kipf2016variational} strategy. A GNN encoder maps each node based on its features and neighborhood to a $p$-dimensional embedding, forming a matrix \(
\mathbf{Z} = h_\theta(G) \in \mathbb{R}^{|V| \times p}
\). A feed-forward decoder $g_\phi$ reconstructs graph information from $\mathbf{Z}$. The decoder is multi-headed, with separate heads for node features \(x_i \in \mathbb{R}^{d_v}\), edge features \(e_{ij} \in \mathbb{R}^{d_e}\) when present, and the adjacency matrix \(A \in \mathbb{R}^{|V| \times |V|}\). 
The encoder–decoder pair is trained jointly (Fig. \ref{fig:GAE}) by minimizing a weighted reconstruction loss over components:
\begin{equation}
\label{eq:loss}
\mathcal{L}(\theta, \phi) =
\alpha \, \frac{1}{|V|} \sum_{i \in V} 
\mathcal{L}_{\text{node}}(\hat{\mathbf{x}}_i, \mathbf{x}_i)
+
\beta \, \frac{1}{|E|} \sum_{(i,j) \in E}
\mathcal{L}_{\text{edge}}(\hat{\mathbf{e}}_{ij}, \mathbf{e}_{ij})
+
\gamma \, \mathcal{L}_{\text{adj}}(\hat{A}, A),
\end{equation}
where $\hat{\mathbf{x}}_i$, $\hat{\mathbf{e}}_{ij}$, and $\hat{A}$ are reconstructions produced by $g_\phi(\mathbf{Z})$, and $\alpha,\beta,\gamma$ control the contribution of each component.

All terms are normalized by the number of reconstructed elements in order to balance gradient magnitudes across heterogeneous components. Additionally, node and edge features are decomposed by type (binary, categorical, continuous) to enable appropriate normalization and dedicated decoder heads with suitable loss functions (e.g., MSE for continuous features).

We pretrain the GAE on trajectories obtained by sampling random valid actions, allowing the graph structure to evolve during learning (e.g., evolving tours in TSP). This corresponds to maximal valid exploration while remaining independent of any learned policy. After convergence, the decoder is discarded, and the encoder is retained to produce meaningful embeddings.

\subsection{Continuous latent spaces}
\label{chap:continuous_spaces}
This subsection describes the methodology for constructing the observation and action spaces from the set of node embeddings \(\mathbf{Z}\) introduced in Section \ref{chap:GNNRepresentation}. By sharing the same GNN representation, changes in graph observations induce semantically consistent shifts in action embeddings, enabling Wolpertinger to adapt to GCO's dynamic action space, where the embedding of a given action entity evolves with the graph state, as discussed in Section \ref{chap:related_work}.

\paragraph{Observation space.}
We represent each observation as a graph \(G_t\) describing the current solution state. A fixed-dimensional observation vector \(o_t\) is obtained by applying \(K\) permutation-invariant graph aggregation operators \(\mathcal{P}_k\) (e.g., sum, mean, or handcrafted graph statistics) over the node embeddings \(\mathbf{Z}_t \in \mathbb{R}^{|V| \times p}\), and concatenating the resulting representations:
\begin{equation}
\label{eq:observation}
o_t = \Phi_{\text{obs}}(G_t) = \bigoplus_{k=1}^{K} \mathcal{P}_k(\mathbf{Z}_t),
\end{equation}
where \(\Phi_{\text{obs}}\) denotes the observation mapping function and \(\bigoplus\) the concatenation operator. This yields a fixed-dimensional, order-invariant graph embedding that summarizes the agent's observation.


\paragraph{Action space.}
Actions may target graph components of varying granularity, including nodes, edges, paths, subgraphs, or combinations thereof (e.g., edge pairs). 
An action $a$ is represented as a set of \textit{components} $\mathcal{C}(a)={c_1,\dots,c_m}$, where each component corresponds to a subset of nodes $\mathcal{V}(c_i) \subseteq V$.
\begin{wrapfigure}{l}{0.55\linewidth}
    \vspace{0pt}
    \includegraphics[width=1.1\linewidth]{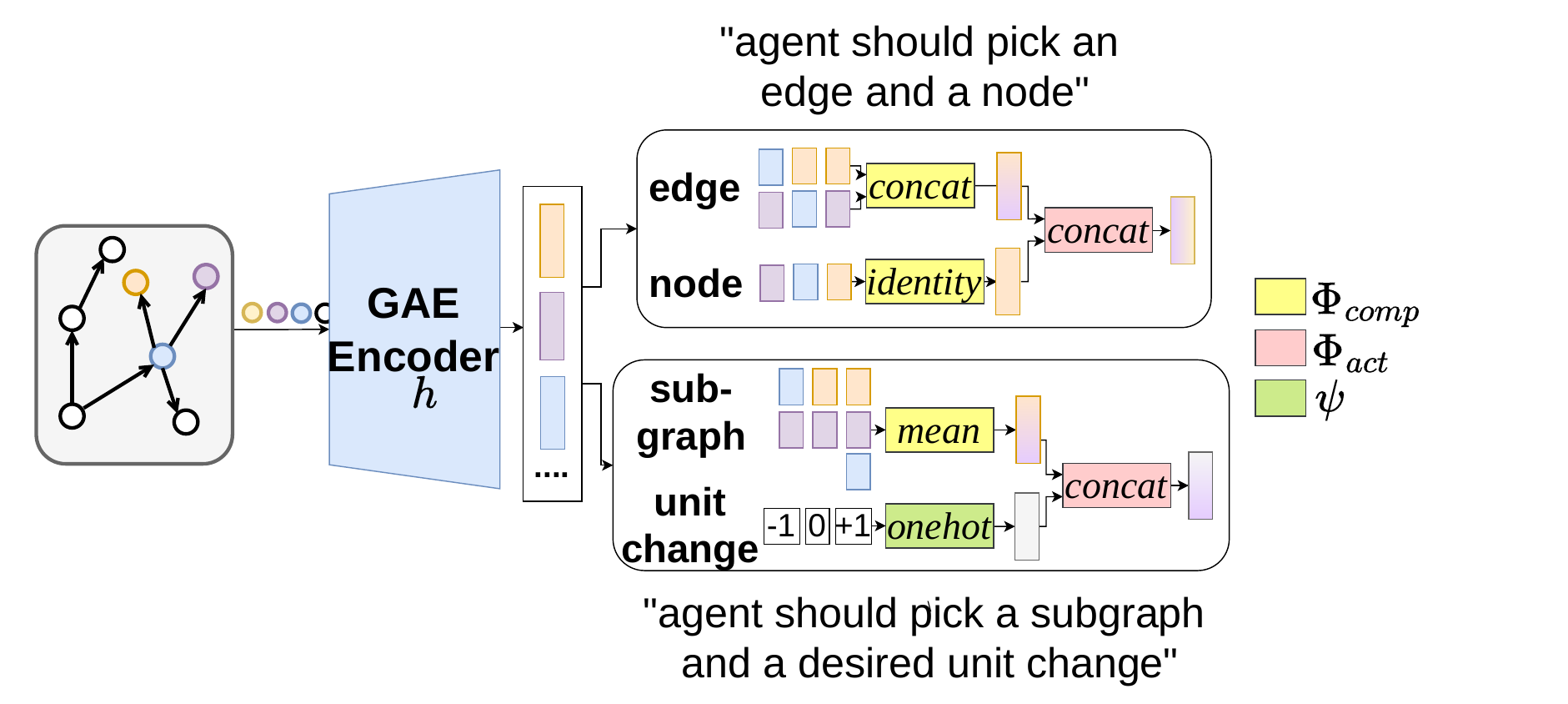}
    \caption{Action points' construction from node embeddings, illustrating two application examples and possible operator choices for \(\Phi_{comb}, \Phi_{act}, \text{ and } \psi\).}
    \label{fig:action_space}
    \vspace{-0pt}
\end{wrapfigure}

A component embedding is obtained by aggregating the embeddings of its nodes with a pooling $\Phi_{\text{comp}}$:
\begin{equation}
\label{eq:pooling_components}
h(c_i) = \Phi_{\text{comp}}(z_j \mid j \in \mathcal{V}(c_i)).
\end{equation}
Depending on the application, this operator may be permutation-invariant (e.g., for subgraphs) or permutation-variant (e.g., directed edges). The action embedding is then obtained by combining embeddings of its components through an aggregation operator \(\Phi_{\text{act}}\) (Fig. \ref{fig:action_space}), which again can be permutation-invariant depending on whether component ordering is relevant:
\begin{equation}
\label{eq:pooling_act}
\mathbf{u}(a) = \Phi_{\text{act}}(h(c_1), \ldots, h(c_m)), \quad c_i \in \mathcal{C}(a),
\end{equation}

In many applications, actions involve not only graph components but also additional non-graph attributes \(\alpha_i\), such as discrete options or features associated with graph elements (e.g., in the cybersecurity scenario of Section \ref{chap:experiments}, vulnerabilities are attached to a node and described by text). Depending on the setting, these attributes may either be independent or conditionally determined by previously selected components.
These are encoded via a task-specific function $\psi(\cdot)$ and combined as:
\begin{equation}
\label{eq:ua}
\mathbf{u}(a) =
\Phi_{\text{act}}\big(h(c_1), \ldots, h(c_m),
\psi(\alpha_1 \mid c_1,\ldots,c_m), \ldots, \psi(\alpha_k \mid c_1,\ldots,\alpha_{k-1})
\big).
\end{equation}
This unified formulation enables heterogeneous actions typical of real-world applications by embedding both graph components and custom attributes in a shared space.



\subsection{Projection agent}
\label{chap:projection}
\begin{wrapfigure}{r}{0.4\textwidth} 
      \vspace{-20pt}
      \centering
      \includegraphics[width=0.4\textwidth]{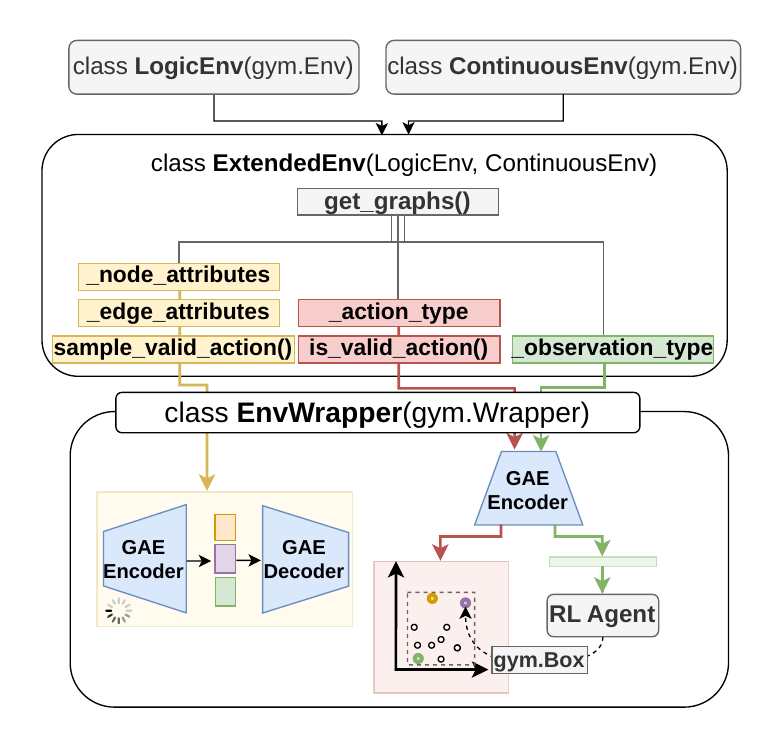}
      \caption{User-defined logic, attributes, and methods provided for GAE pre-training (left) and RL integration (right).}
      \label{fig:LaGCORL}
      \vspace{-10pt}
\end{wrapfigure}

The learned action embeddings enable the proposed projection agent by defining a bounded continuous latent action space over all possible action embeddings, onto which the agent projects its decisions. The latent action space $\mathbf{U} = \{\mathbf{u}(a_1), \ldots, \mathbf{u}(a_n)\}$ is integrated into the agent through a three-step pipeline: (1) optional preprocessing, such as normalization or dimensionality reduction, while preserving the geometric structure; (2) empirical estimation of the bounds of $\mathbf{U}$ across graph instances (min/max per dimension); and (3) alignment of its dimensionality and bounds with the agent's \texttt{gym.Box} action space (see RL agent in Figure \ref{fig:LaGCORL}).


Given an observation \(o_t\), the policy \(\pi\) outputs a continuous proto-action in this space 
\(
\tilde{a}_t = \pi(o_t) \in \mathbf{U},
\)
where $\tilde{a}_t$ represents a "desired" target point in latent space rather than a valid discrete action. This proto-action is then decoded by retrieving its \(k\) nearest embeddings, forming a candidate set \(\mathcal{A}_k(o_t)\). A final single action is selected from this set using a proper decoding strategy: e.g., simple NN (\(k{=}1\)), stochastic or rule-based selection, or a learned scoring function over the \(k\)-NN candidates. This results in a region-based learning signal in \(\mathbf{U}\), where the decoding induces a space partition, and all points within a region map to a discrete action. As a result, the projection is piecewise constant, and the gradients act on the geometry of these regions rather than on individual actions.

\subsection{RL-GCO automated framework}
\label{chap:python}

This subsection presents our \texttt{LaGCO-RL} library and its primary research contribution: a framework for studying graph representation learning, latent action-space design, and Wolpertinger-based RL for GCO (Sections~\ref{chap:GNNRepresentation}--\ref{chap:projection}). 
At the environment level, users define a standard Gym environment encoding the task logic and extend it through a class inheriting from both \texttt{ContinuousEnv} (Fig.~\ref{fig:LaGCORL}) and a task-specific base class.
The framework core idea is to minimize task-specific engineering, requiring only a small set of high-level attributes and methods leveraged by a \texttt{gym.Wrapper} module. The graph structure should be provided through \textbf{\textcolor{gray}{get\_graphs()}}, with step-varying attributes, while node and edge encodings are defined via \textbf{\textcolor{gray}{\_node\_attributes}} and \textbf{\textcolor{gray}{\_edge\_attributes}}, enabling automatic construction of decoder heads and loss functions for GNN self-supervised learning, supported by \textbf{\textcolor{gray}{sample\_valid\_action()}}. Observations are then derived through \textbf{\textcolor{gray}{\_observation\_type}}, using pooling-based or custom invariant functions. The action space is specified via \textbf{\textcolor{gray}{\_action\_type}} as combinations of proper decision elements; the framework generates the full action set via Cartesian products and optionally filters infeasible actions using \textbf{\textcolor{gray}{is\_valid\_action(action)}}, allowing time-varying action spaces. The resulting continuous space is exposed as a \textbf{\textcolor{gray}{gym.Box}}, aligned with the projection agent’s action space, or appropriately wrapped for compatibility in case of other RL strategies. Examples of instantiations are provided in Appendix~\ref{chap:LAGCoRL}.

\section{Experiments} \label{chap:experiments}


\paragraph{Benchmarks.} We evaluate our approach on seven GCO benchmarks. This includes three standard problems with node-level decisions and linearly growing action spaces—TSP, MinVertex, and Maximum Cut (\textit{MaxCut}) \citep{MaxCut}; the latter is of particular importance, as more than half of the 21 Karp's canonical CO problems are reducible to MaxCut \citep{barrett2020exploratory}.
In addition, we consider four application-driven tasks drawn from delay-sensitive domains (networking and cybersecurity): Virtual Machine Placement (\textit{Placement}) \citep{DeepRLVMP}, Cyber-Attack Path Prediction (\textit{Cyber-Path}) \citep{terranova2025rlAttackPath}, \textit{OSPF}, and \textit{Traffic}.
Unlike classical benchmarks, these tasks involve structured actions combining graph and non-graph entities, while involving super-linear growth in the action space.
Table~\ref{tab:envs} summarizes their action components, the expected growth of the action space, and the structured action representations used for each benchmark. 
Additional details on the logic of each environment, the episode structure, the score function, and the graph features are provided in Appendix~\ref{chap:benchmark_envs}.

To generate scenarios, we perform an extended empirical sweep inspired by \citep{learningCOoverGraphs} (e.g., 10k sweeps for \textit{TSP}) to approximate worst- and best-case GCO solutions via standard heuristic-guided exploration for each instance (see Appendix \ref{chap:scenario_generation} for details). This produces a normalized reference score in $[0,1]$, which mitigates scale effects (e.g., distance-dependent costs in \textit{TSP}), enables per-step relative rewards to alleviate the challenges of sparse episodic rewards \citep{barrett2020exploratory} (see Appendix \ref{chap:benchmark_envs}), and ensures comparability and aggregation across instance scores.

\begin{figure*}[t]
    \centering

    \begin{subfigure}[t]{0.135\textwidth}
        \centering
        \includegraphics[width=\linewidth]{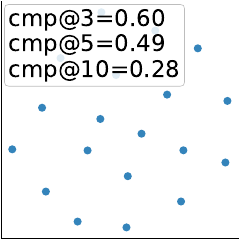}
        \caption{TSP}
    \end{subfigure}
    \hfill
    \begin{subfigure}[t]{0.135\textwidth}
        \centering
        \includegraphics[width=\linewidth]{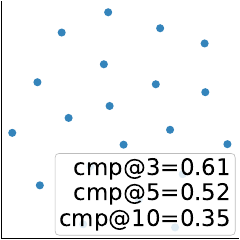}
        \caption{MinVertex}
    \end{subfigure}
    \hfill
    \begin{subfigure}[t]{0.135\textwidth}
        \centering
        \includegraphics[width=\linewidth]{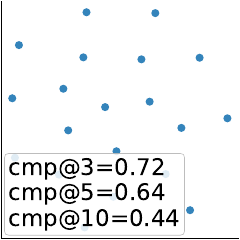}
        \caption{MaxCut}
    \end{subfigure}
    \hfill
    \begin{subfigure}[t]{0.135\textwidth}
        \centering
        \includegraphics[width=\linewidth]{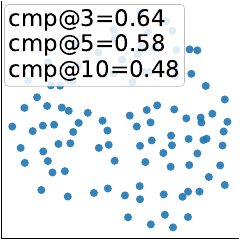}
        \caption{Placement}
    \end{subfigure}
    \hfill
    \begin{subfigure}[t]{0.135\textwidth}
        \centering
        \includegraphics[width=\linewidth]{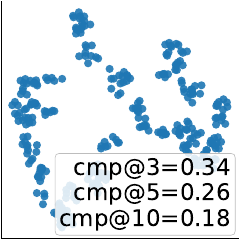}
        \caption{Cyber-Path}
    \end{subfigure}
    \hfill
    \begin{subfigure}[t]{0.135\textwidth}
        \centering
        \includegraphics[width=\linewidth]{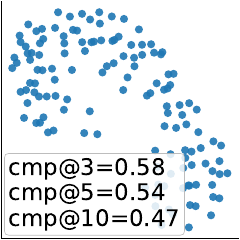}
        \caption{OSPF}
    \end{subfigure}
    \hfill
    \begin{subfigure}[t]{0.135\textwidth}
        \centering
        \includegraphics[width=\linewidth]{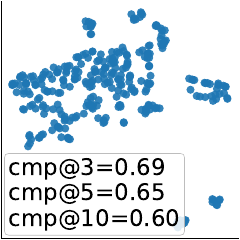}
        \caption{Traffic}
    \end{subfigure}

    \caption{UMAP projection of benchmark action spaces for a 20-node graph instance, illustrating local geometry and neighborhood proximity with the \texttt{cmp@k} (\(k \in {3,5,10}\)) (computed on 20 instances).}
    \label{fig:tiny_umaps_subfig}
\end{figure*}

\uline{\textit{Observation:}} Figure~\ref{fig:tiny_umaps_subfig} shows 2D UMAP \citep{UMAP} projections of action spaces for 20-node benchmarks. We quantify complexity using compactness, $\mathrm{cmp}@K$, defined as the ratio of the mean pairwise distance to the $k$-NN distance ($k \in {3,5,10}$), averaged over 20 instances. Real-world benchmarks exhibit higher $\mathrm{cmp}$, indicating denser neighborhoods and greater overlap.

\begin{table*}[t]
\centering
\small
\caption{Action-space structure across benchmarks, including worst-case growth and the chosen representation function (\(h\) denotes the node embedding function).}
\label{tab:envs}

\begin{tabular}{p{2cm} p{4.1cm} p{1.5cm} p{4.2cm}}
\specialrule{1pt}{2pt}{2pt}
\rowcolor{gray!20}
Benchmark & Action Components & Worst-Case Growth & Representation (\(h: \mathcal{V} \rightarrow \mathbb{R}^{16}\))\\
\midrule

TSP
& node $v$
& $O(|V|)$
& $h(v)$ \\
\specialrule{0.1pt}{0pt}{0pt}

\rowcolor{gray!10}
MinVertex 
& node $v$ (not visited)
& $O(|V|)$
& $h(v)$ \\
\specialrule{0.1pt}{0pt}{0pt}

MaxCut 
& node $v$
& $O(|V|)$
& $h(v)$ \\

\specialrule{1pt}{2pt}{2pt}

\rowcolor{gray!10}
Placement
& \makecell[l]{source node $u$ (type=VM) \\ target node $v$ (type=PM)}
& $O(|V|^2)$
& concat$(h(u),h(v))$ \\
\specialrule{0.1pt}{0pt}{0pt}

Cyber-Path
& \makecell[l]{source node $u$ (status=owned) \\ target node $v$ (status=discovered) \\ vuln $\alpha \in v$}
& $O(k|V|^2)$
& concat$(h(u),h(v),\texttt{BERT}(\alpha_\text{text}))$ \\
\specialrule{0.1pt}{0pt}{0pt}

\rowcolor{gray!10}
OSPF 
& \makecell[l]{edge $(u,v)$ \\ weight value $w \in \{1, ..., N\}$}
& $O(k|V|^2)$
& concat$(h(u),h(v),\texttt{onehot}(w))$ \\
\specialrule{0.1pt}{0pt}{0pt}

Traffic
& \makecell[l]{edge $(u,v)$ (type=traffic) \\ path $(u_1,\ldots,u_n)$ (type=link)}
& $O(|V|^2 2^{|V|})$
& pad(concat$(h(u),h(v),...,h(u_n))$) \\

\specialrule{1pt}{2pt}{2pt}
\end{tabular}
\end{table*}

\paragraph{Solutions.}
We compare the three policy parameterizations introduced in Section \ref{chap:related_work}: \textit{Discrete}, \textit{Iterative}, and \textit{Projection}. The discrete baselines use two observation encodings: padding (P-, permutation-variant) and GNN-based pooling (G-, permutation-invariant), each evaluated with and without action masking (-M representing "with"). We implement these baselines using \texttt{PPO} and \texttt{MaskablePPO} from SB3, respectively. Our projection method likewise uses \texttt{PPO}, with a z-score-normalized action space and $(K{=}1)$-NN decoding, representing the most challenging high-precision setting. The iterative solution is implemented by extending SB3's fitted Q-iteration \texttt{DQN}, as this approach is not available in existing Python RL libraries. Both latent-space methods operate over valid actions only through action masking.

We also distinguish between the \textit{non-revisiting} and \textit{revisiting} interaction strategies as orthogonal axes. By default, we follow the non-revisiting strategy inspired by \texttt{S2V-DQN} \cite{learningCOoverGraphs}, where each selected action constrains the remaining action space. For example, in \textit{Placement}, assigning a VM to a PM makes all other actions involving that VM allocation invalid, progressively shrinking the set of feasible actions and yielding a one-shot construction process.
We additionally evaluate a \textit{revisiting} strategy inspired by \texttt{ECO-DQN} \cite{barrett2020exploratory}, which allows previously selected actions to be reconsidered (e.g., reassigning a previously placed VM), rather than removing action points conditional to prior selections. Since revisitation is only meaningful for non-sequential problems, other benchmarks (TSP and Cyber-Path) are evaluated exclusively under the non-revisiting strategy. Details of the revisiting procedure are provided in Appendix \ref{chap:benchmark_envs}.

\uline{\textit{Observation:}} We focus our main comparisons on RL-based parameterizations within a shared methodology, ensuring a consistent autoregressive and data-driven setting. We therefore exclude supervised and unsupervised methods that rely on pre-existing labels or do not enforce the feasibility constraints required by our benchmarks (Appendix~\ref{chap:extended_related_work}). Since optimal solvers are not uniformly available across benchmarks, we instead use a heuristic-guided empirical reference for min-max score normalization in [0,1] (details in Appendix~\ref{chap:benchmark_envs}).

Solution-specific hyperparameters, along with those that govern the GAE and its architecture, are optimized using the Tree-structured Parzen Estimator (TPE) \citep{tpe} for 25 trials each. The objective is task-specific performance evaluated on a held-out validation set, considering only the \textit{TSP} benchmark for simplicity. Complete search spaces, controller details, seeds, and hyperparameters are provided in Appendix~\ref{chap:hyperparams}, with observation-space design in Appendix~\ref{chap:observation_space}.

\subsection{How well do RL-GCO approaches generalize across graph distributions?}
\label{chap:generalization}

\paragraph{Experimental design.} For each benchmark, we generate 101 diverse instances by sampling key parameters from typical benchmark distributions to evaluate out-of-distribution generalization \citep{extrapolation} across instance sizes and training configurations.
Scenarios are then ordered by size and evaluated under four transfer-oriented training strategies. In the first three, agents train on a single scenario based on their size—\textit{smallest} (S), \textit{medium} (M), or \textit{largest} (L)—and are tested on all remaining instances; each strategy is repeated \(K=5\) times with different seeds and scenarios (removing those already selected from the selection pool). The fourth, \textit{varied} (V), adopts a K-fold-inspired approach: agents are trained on five non-overlapping subsets of scenarios (20\%), with training instances rotated every 100 episodes. For each run, performance is assessed over 5 episodes per unseen instance (80\% for V, 100 instances for S, M, and L), with the best result retained; aggregated outcomes across 20 runs per solution–benchmark pair are reported in Table~\ref{tab:generalization_light}.
We report the Interquartile Mean (IQM) over these 100 best scores, following the guidelines on suitable RL indicators from \cite{EdgeStatisticalPrecipe}.
Additionally, the \(\Delta\) columns report the test-to-training gap (averaged across strategies) for each method. Training time, learning curves, extended generalization results, and additional metrics (including evaluation without best-per-environment selection) are provided in Appendices~\ref{chap:training_time}, \ref{chap:learning_curves}, and \ref{chap:extended_generalization}. Appendix~\ref{chap:ablation_GNN} compares GAE loss variants against plain graph features and random embeddings, assessing the contribution of the learned graph representations.

\newcommand{\bestblock}[1]{\dashuline{\textbf{#1}}} 

\begin{table}[t]
\scriptsize
\centering
\caption{IQM of normalized generalization scores across benchmarks and training regimes (S/M/L/V) under the non-revisiting strategy. \(\Delta\) denotes the average train–test generalization gap. \bestblock{Underlined bold} indicates the best method per benchmark.} 

\setlength{\tabcolsep}{2pt}
\begin{tabular}{l|cccc:c|cccc:c|cccc:c}
\toprule
& \multicolumn{5}{c|}{TSP} & \multicolumn{5}{c|}{MinVertex} & \multicolumn{5}{c}{MaxCut} \\
Method & S & M & L & V & \(\Delta\) & S & M & L & V & \(\Delta\) & S & M & L & V & \(\Delta\) \\
\midrule
P-Discrete & 0.00 & 0.00 & 0.00 & 0.00 & +0.00 & 0.00 & 0.00 & 0.00 & 0.16 & -0.62 & 0.91 & 0.43 & 0.06 & 0.87 & +0.00 \\
P-Discrete-M & 0.50 & 0.47 & 0.47 & 0.48 & -0.05 & 0.39 & 0.36 & 0.39 & 0.39 & -0.44 & 0.91 & 0.89 & 0.88 & 0.87 & -0.02 \\
G-Discrete & 0.00 & 0.00 & 0.00 & 0.00 & -0.05 & 0.00 & 0.00 & 0.13 & 0.03 & -0.28 & 0.07 & 0.89 & 0.90 & 0.88 & -0.24 \\
G-Discrete-M & 0.51 & 0.50 & 0.50 & 0.50 & -0.19 & 0.32 & 0.35 & 0.36 & 0.39 & -0.44 & 0.07 & 0.90 & 0.91 & 0.90 & -0.23 \\
Iterative & 0.90 & 0.79 & 0.97 & \bestblock{0.99} & -0.04 & 0.40 & 0.05 & 0.00 & 0.16 & -0.19 & 0.91 & 0.23 & 0.33 & 0.21 & -0.15 \\
Projection (ours) & 0.78 & 0.76 & 0.60 & 0.71 & -0.13 & \bestblock{0.70} & 0.01 & 0.03 & 0.02 & -0.18 & 0.93 & \bestblock{0.94} & 0.91 & 0.93 & -0.03 \\
\bottomrule
\end{tabular}


\setlength{\tabcolsep}{2pt}

\begin{tabular}{l|cccc:c|cccc:c|cccc:c|cccc:c}
\toprule
& \multicolumn{5}{c|}{Placement} & \multicolumn{5}{c|}{Cyber-Path} & \multicolumn{5}{c|}{OSPF} & \multicolumn{5}{c}{Traffic} \\
Method & S & M & L & V & \(\Delta\) & S & M & L & V & \(\Delta\) & S & M & L & V & \(\Delta\) & S & M & L & V & \(\Delta\) \\
\midrule
P-Discrete & 0.67 & 0.58 & 0.58 & 0.60 & -0.36 & 0.19 & 0.18 & 0.19 & 0.18 & -0.05 & 0.08 & 0.11 & 0.03 & 0.00 & -0.33 & 0.11 & 0.48 & 0.16 & 0.37 & -0.35 \\
P-Discrete-M & 0.72 & 0.74 & 0.70 & 0.71 & -0.30 & 0.61 & 0.63 & 0.61 & \bestblock{0.67} & -0.20 & 0.86 & 0.85 & 0.84 & 0.81 & -0.04 & 0.81 & 0.79 & 0.80 & 0.81 & -0.13 \\
G-Discrete & 0.42 & 0.58 & 0.66 & 0.69 & -0.33 & 0.18 & 0.19 & 0.18 & 0.17 & -0.05 & 0.36 & 0.20 & 0.19 & 0.59 & -0.55 & 0.24 & 0.57 & 0.49 & 0.58 & -0.42 \\
G-Discrete-M & \bestblock{0.79} & 0.76 & 0.74 & 0.72 & -0.24 & 0.45 & 0.45 & 0.51 & 0.50 & -0.28 & 0.51 & 0.58 & 0.74 & 0.76 & -0.28 & 0.69 & 0.72 & 0.78 & 0.77 & -0.16 \\
Iterative & 0.26 & 0.02 & 0.21 & 0.29 & -0.19 & 0.34 & 0.24 & 0.19 & 0.24 & -0.22 & 0.93 & 0.93 & 0.91 & 0.92 & +0.00 & 0.66 & 0.54 & 0.49 & 0.61 & -0.10 \\
Projection (ours) & \bestblock{0.79} & 0.78 & 0.76 & \bestblock{0.79} & -0.12 & 0.60 & 0.64 & 0.60 & \bestblock{0.67} & -0.17 & \bestblock{0.95} & 0.92 & 0.89 & 0.90 & -0.03 & 0.79 & \bestblock{0.85} & 0.83 & 0.84 & -0.09 \\
\bottomrule
\end{tabular}


\label{tab:generalization_light}

\end{table}

\begin{table*}[t]
\centering
\caption{Effect of revisiting (R) vs.\ non-revisiting (NR) interaction on revisitable tasks.}
\label{tab:interaction_strategy}
\scriptsize
\setlength{\tabcolsep}{2pt}
\begin{tabular}{l|cccc|cccc|cccc|cccc|cccc}
\toprule
& \multicolumn{4}{c|}{MinVertex}
& \multicolumn{4}{c|}{Maxcut}
& \multicolumn{4}{c|}{Placement}
& \multicolumn{4}{c|}{OSPF}
& \multicolumn{4}{c}{Traffic} \\
& S & M & L & V 
& S & M & L & V 
& S & M & L & V 
& S & M & L & V
& S & M & L & V  \\
\midrule

P-Discrete-M (NR)
& 0.39 & 0.36 & 0.39 & 0.39 & 0.91 & 0.89 & 0.88 & 0.87 & 0.72 & 0.74 & 0.70 & 0.71 & 0.86 & 0.85 & 0.84 & 0.81 & 0.81 & 0.79 & 0.80 & 0.81 \\
P-Discrete-M (R)
& 0.00 & 0.00 & 0.00 & 0.00 & 0.91 & 0.47 & 0.80 & 0.17 & 0.61 & 0.64 & 0.66 & 0.65 & 0.25 & 0.22 & 0.70 & 0.01 & 0.36 & 0.74 & 0.77 & 0.73 \\

\midrule
Iterative (NR)
& 0.40 & 0.05 & 0.00 & 0.16  & 0.91 & 0.23 & 0.33 & 0.21  & 0.26 & 0.02 & 0.21 & 0.29  & 0.93 & 0.93 & 0.91 & 0.92 & 0.66 & 0.54 & 0.49  & 0.61  \\
Iterative (R)
& 0.13 & 0.01 & 0.01 & 0.00 & 0.69 & 0.86 & 0.78 & 0.86 & 0.55 & 0.02 & 0.07 & 0.15  & 0.00 & 0.04 & 0.00 & 0.02  & 0.01 & 0.01 & 0.00 & 0.01 \\

\midrule
Projection (NR)
& \bestblock{0.70} & 0.01 & 0.03 & 0.02 & 0.93 & \bestblock{0.94} & 0.91 & 0.93 & 0.79 & 0.78 & 0.76 & 0.79 & \bestblock{0.95} & 0.92 & 0.89 & 0.90 & 0.79 & \bestblock{0.85} & 0.83  & 0.84 \\
Projection (R)
& 0.13 & 0.38 & 0.07 & 0.07 & 0.84 & 0.91 & 0.91 & 0.93 & 0.87 & 0.87 & 0.86 & \bestblock{0.92} & 0.67 & 0.79 & 0.84 & 0.88 & 0.80 & 0.82 & 0.81 & 0.81 \\

\bottomrule

\end{tabular}
\label{tab:revisiting}
\end{table*}







\paragraph{Analysis.} Table \ref{tab:generalization_light} reports results under the non-revisiting strategy, aiming to compare the different policy parameterization strategies. The iterative approach performs particularly well on TSP, matching 10k-sweep heuristic optima with only five trials per instance. This demonstrates that RL can match strong heuristic baselines under a highly limited evaluation budget.
On the remaining benchmarks, our projection method achieves the best (or tied-best) performance, with particularly pronounced gains on real-world problems with larger and more complex action spaces. The two approaches face different scaling effects. As action-space dimensionality increases, the projection agent benefits from a richer latent space for representing the target action before decoding. In contrast, the iterative approach becomes less effective as the candidate action set grows, since \(\arg\max\) selection requires accurately ranking an increasing number of actions, making it more sensitive to Q-value estimation errors. The masked discrete approach remains more robust by restricting selection to valid actions, achieving the second-best overall performance.

Table \ref{tab:revisiting} further examines the effect of the interaction strategy by evaluating the different policy parameterizations (using the best discrete variant) under both non-revisiting (NR) and revisiting (R) strategies. This comparison evaluates whether RL agents can exploit an unrestricted action space that allows revisiting previous decisions, at the cost of more actions per step.
Revision substantially degrades the discrete and iterative approaches, whereas projection remains comparatively more stable, suggesting greater robustness to larger action spaces. In terms of solution quality, revisiting provides limited benefits, improving \textit{Placement} while yielding similar or slightly worse results elsewhere.

We additionally demonstrate competitive generalization against classical, unsupervised, and RL-based solvers in Appendix~\ref{chap:beyond_solvers}, including MaxCut, where we achieve a $+0.03$ mean score over physics-informed GNNs \cite{Schuetz2022PhysicsInformed} and $95\%$ of simulated annealing performance.

\subsection{How does inference time scale with graph size?}
\label{chap:efficiency} 

\begin{wraptable}{r}{0.48\linewidth}
\vspace{-10pt}
\centering


\begin{minipage}{0.99\linewidth}
    \centering
    \includegraphics[width=\linewidth]{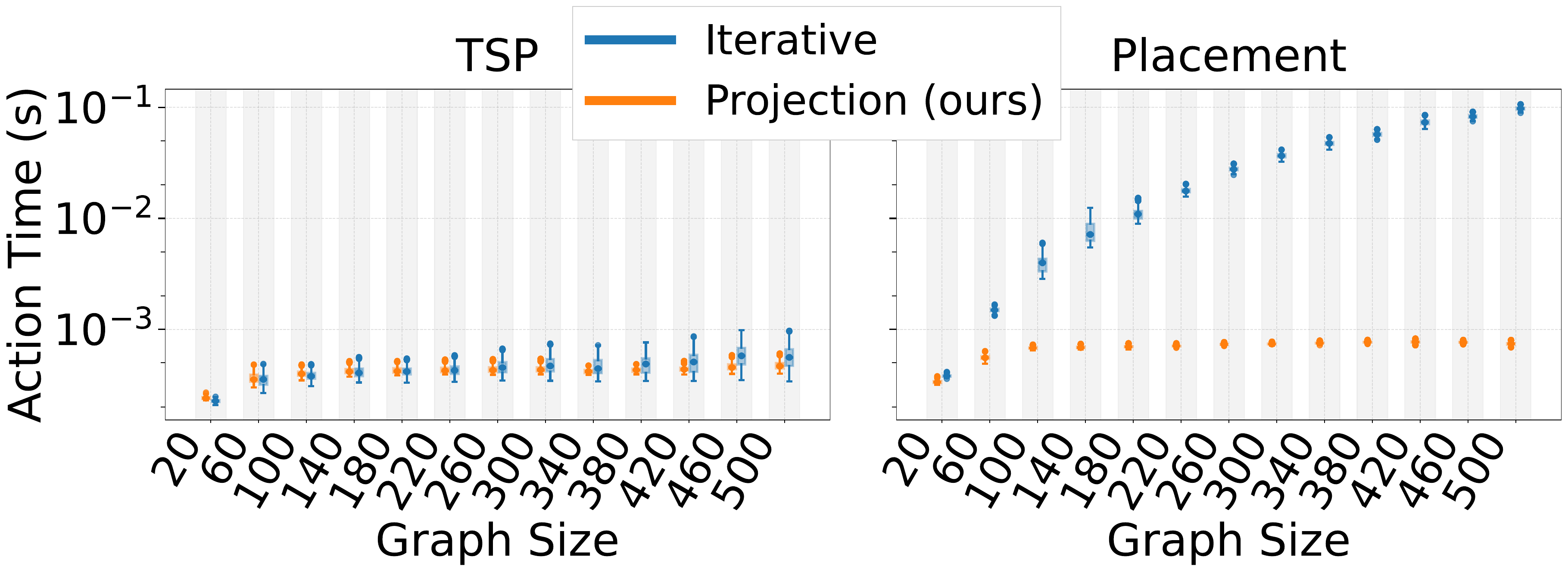}
    \captionof{figure}{Evolution of inference-time action-selection latency across graph sizes for the representative TSP and Placement benchmarks.}
\label{fig:travel_action}
\end{minipage}

\vspace{6pt}

\small
\setlength{\tabcolsep}{3pt}
\begin{tabular}{l cc|cc|cc}
\toprule
& \multicolumn{2}{c|}{P-Discrete} 
& \multicolumn{2}{c|}{\shortstack{Projection\\(ours)}} 
& \multicolumn{2}{c}{Iterative} \\
Env & $\alpha$ & $R^2$ 
    & $\alpha$ & $R^2$ 
    & $\alpha$ & $R^2$ \\
\midrule
TSP        & 0.07 & 0.89 & \textbf{0.17} & 0.85 & 0.25 & 0.94 \\
MinVertex  & 0.16 & 0.69 & \textbf{0.30} & 0.88 & 0.32 & 0.95 \\
MaxCut     & 0.03 & 0.78 & \textbf{0.22} & 0.94 & 0.25 & 0.92 \\
\midrule
Placement  & 0.09 & 0.85 & \textbf{0.22} & 0.85 & 1.81 & 0.99 \\
Cyber-Path & 0.09 & 0.75 & \textbf{0.17} & 0.73 & 1.46 & 0.95 \\
OSPF       & 0.20 & 0.76 & \textbf{0.49} & 0.94 & 1.44 & 0.99 \\ 
Traffic    & 0.05 & 0.70 & \textbf{0.73} & 0.71 & 4.68 & 0.99 \\
\bottomrule
\end{tabular}

\caption{Scaling of action-selection time with fitted power-law exponent \(\alpha\). Lower \(\alpha\) indicates better scaling with graph size. }

\label{tab:action_time_scaling}
\vspace{-20pt}
\end{wraptable}
\paragraph{Experimental design.} We evaluate computational scalability by analyzing how action-selection time scales with graph size across three approaches: the proposed GNN-based \textit{projection} method, which uses a FAISS Flat Index for NN lookup with linear retrieval complexity \citep{FAISS}; the \textit{discrete} baseline, which requires only a single DNN inference without additional computation; and the \textit{iterative} method, which repeatedly evaluates the DNN. All experiments are conducted on a benchmark machine.\footnote{\textit{Benchmark Machine:} AMD Ryzen 7 PRO 7840U w/ Radeon 780M Graphics, 30 GiB RAM, Ubuntu 22.04.4.}
Specifically, we fit a power-law model \(T(n) = c \cdot n^\alpha\) where \(T(n)\) is the median action-selection time for a graph of size \(n\), and \(\alpha\) is the scaling exponent. 

\paragraph{Analysis.} Table~\ref{tab:action_time_scaling} reports the scaling exponent \(\alpha\) and corresponding \(R^2\) values. As expected, \textit{discrete} scales best due to its single DNN evaluation, while \textit{projection} substantially improves over \textit{iterative} by replacing per-action DNN evaluations with a single inference and NN search. This advantage is strongest in delay-sensitive benchmarks with superlinear action spaces; Figure~\ref{fig:travel_action} shows TSP and Placement, with additional scaling and approximate-NN results in Appendices~\ref{chap:scalability_curves} and~\ref{chap:approximateNN}.

\section{Discussion} 
\label{chap:conclusion}

\paragraph{Conclusion.}
We introduce an end-to-end framework for latent-space RL-GCO and a Wolpertinger adaptation that improves generalization and scalability across parameterizations. We further propose realistic benchmarks with structured action spaces and demonstrate consistent gains for our solution. For reproducibility, we release \texttt{LaGCO-RL}, a modular library for integrating new GCO benchmarks.

\paragraph{Limitations.}
Our methodology has several limitations. (1) Encoding the evolving partial solution as state-dependent node features requires repeated GNN recomputation; future work could decouple graph structure from solution state to reduce this overhead. (2) We use a simple NN decoder, deliberately chosen as a conservative worst-case setting, though learned or heuristic ranking could improve performance. (3) We use a single encoding per action-space setting, leaving richer multi-encoding representations for future work. (4) Self-supervised embeddings are not jointly fine-tuned with the RL policy, enabling fair comparisons but potentially limiting performance. 



\bibliographystyle{plainnat}
\bibliography{bibliography}  

\appendix

\newpage
\section{Related work: beyond RL solutions}
\label{chap:extended_related_work}

Several ML paradigms have been proposed for data-driven heuristic discovery in GCO. This section reviews the most relevant approaches according to their underlying learning paradigm.

\paragraph{Supervised learning.}
Supervised approaches have been extensively studied in the literature, with recent advances including end-to-end methods that directly generate complete solutions—for example, diffusion-based models \citep{DIFUSCO}—as well as neural improvement heuristics \citep{neuralImprovementHeuristics}. The latter departs from full solution construction by adopting an iterative refinement strategy, where existing solutions are progressively improved through local decisions. This shift simplifies the learning problem while still achieving competitive performance.

However, these methods inherently rely on labeled solutions, which are often costly or infeasible to obtain with exact methods for large-scale instances \citep{biLevelFramework}. As a consequence, in practice, labels are typically generated using existing manually defined heuristics, introducing an inherent performance ceiling: models are trained to mimic potentially suboptimal solutions and cannot surpass this upper bound \citep{cappart2022combinatorialoptimizationreasoninggraph, variationalAnnealing}.

\paragraph{Unsupervised learning.}
Unsupervised methods, similar to RL, do not require labeled data and instead optimize the problem objective directly \citep{Peng2021GraphLearning}. A common strategy is to relax these discrete optimization problems into continuous formulations via concave surrogate losses, which are optimized in an unsupervised manner \citep{variationalAnnealing, unsupervisedCORelaxation}. These approaches have been explored using various architectures, including GNNs \citep{karalias2021erdosgoesneuralunsupervised}, hypergraph neural networks \citep{distributedConstrainedCO}, physics-informed neural networks \citep{Schuetz2022PhysicsInformed}, and variational annealing \citep{variationalAnnealing}. 


Despite their promise, unsupervised approaches face several challenges. Their generalization capabilities remain insufficiently explored, complex constraints are difficult to incorporate \citep{biLevelFramework}, and models may converge to suboptimal local minima \citep{distributedConstrainedCO}. Additionally, reliance on continuous relaxations introduces approximation gaps that can degrade solution quality and complicate optimization \citep{cappart2022combinatorialoptimizationreasoninggraph}.

\paragraph{Reinforcement learning.}
In contrast to other approaches, RL optimizes the objective through sequential decision-making, relying on step-wise evaluative feedback rather than fixed supervision over entire trajectories, as in supervised learning or imitation learning \citep{imitationLearning}. Instead of using static labels for complete solutions, RL leverages reward signals to assess the quality of actions over time. This framework enables agents to go beyond the limitations imposed by predefined datasets and known reference solutions.

However, the key challenge of RL lies in the inherently sparse reward structure of GCO problems, where meaningful evaluations are typically only available at the end of an episode, making training difficult \citep{variationalAnnealing, biLevelFramework}. Prior work has addressed this limitation by introducing intermediate reward signals to alleviate sparsity and improve learning efficiency \citep{barrett2020exploratory}. A similar idea is adopted in this work, which leverages the worst- and best-known solutions to compute step-wise improvement signals, thereby providing a denser reward for more effective learning.

\paragraph{Hybrid approaches.}
Hybrid methods have been proposed to combine learning-based and classical optimization techniques for solving GCO problems. In learning-assisted optimization, several works have proposed neural models able to guide traditional solvers, such as learning heuristics for A* search \citep{GraphEditDistance} or branching strategies in branch-and-bound \citep{ExactCOGCN, learningBranchAAAI}. Similar ideas have also been explored using RL \citep{cappart2020combiningreinforcementlearningconstraint}. Another study proposed a bi-level framework in which RL reduces the problem space before applying a fast heuristic for refinement \citep{biLevelFramework}.

More recently, hybrid learning approaches combine multiple ML paradigms to further improve the learning-based solution. For instance, supervised learning can be used to prune the search space explored by RL policies \citep{GCOMB}, improving scalability to large graphs. However, such methods reintroduce the limitations of supervised learning, as their effectiveness depends on the quality of the initial labels. Nevertheless, these approaches remain complementary and can be naturally integrated into all RL-based frameworks, including the one proposed in this work.



\section{Benchmark environments}
\label{chap:benchmark_envs}
This section describes the benchmark environments used in the experimental study (Section \ref{chap:experiments}), focusing on their logical structure and dynamics, graph representations, and the scoring criteria used to evaluate solutions. During training, the best-known score is additionally employed as a termination criterion to accelerate convergence. This mechanism is removed at test time to avoid the potential drawbacks associated with access to supervised solutions, as discussed in Appendix \ref{chap:extended_related_work}.

\subsection{Traveling Salesman Problem}
\textit{Logic:} Construct a minimum-length tour that visits each city exactly
once and returns to the starting city (hard constraint).

\begin{itemize}
\item \textbf{Parameters:} Number of cities $N$ and coordinate range
defining the $(x,y)$ positions of the cities.

\item \textbf{Graph:} Nodes represent cities, and edges encode distances
between cities. The underlying graph is fully connected, but is sparsified
by retaining the $K=10$ nearest neighbors per node to avoid a fully
connected graph and the resulting issues with GNN embeddings.

\item \textbf{Node features:} A binary visited flag
$\in \{0,1\}$ indicating whether a city has already been included in the
tour, together with its spatial coordinates $(x,y)$.

\item \textbf{Edge features:} Euclidean distances between cities.

\item \textbf{Reset:} A starting city is randomly selected and marked as
visited, while all remaining cities are initialized as unvisited. The
current city is set to the selected starting city and the tour is
initialized from this node.

\item \textbf{Step helper:} At each step, the agent selects the next city
to visit. The selected city is appended to the current tour, marked as
visited, and removed from the action space. The distance from the current
city to the selected city is added to the accumulated tour length, and
the selected city becomes the new current city.

\item \textbf{Objective and reward:} The objective is to minimize the total tour
length, given by the sum of the Euclidean distances along the constructed
tour, including the final return to the starting city. The reward is
normalized as a relative score using the empirically known best and
worst tour lengths. When the tour is
incomplete, the remaining cities are incorporated using a worst-case
completion heuristic to estimate the final tour cost, encouraging the
agent to make decisions that also facilitate completion of the tour.

\item \textbf{Termination:} The episode terminates once all $N$ cities
have been visited. For a complete tour, the distance from the final city
back to the starting city is then added to obtain the total tour length.

\item \textbf{Score:} The score is based on the
constructed tour length, normalized relative to the empirically known
best and worst solutions.

\item \textbf{Revisiting strategy:} Not supported by this sequential
benchmark, as each action permanently extends the tour and consumes the
selected city from the available action space.

\end{itemize}


\subsection{Minimum Vertex Cover}

\textit{Logic:} Select the smallest subset of nodes such that every edge is
incident to at least one selected node (hard constraint).

\begin{itemize}
    \item \textbf{Parameters:} Number of nodes $N$ and edge probability $p$.

    \item \textbf{Graph:} An Erd\H{o}s--R\'enyi random graph. A solution is a
    subset of nodes whose selected nodes collectively cover every edge.

    \item \textbf{Node features:} A binary selected flag
    $\in \{0,1\}$ indicating whether a node currently belongs to the cover.

    \item \textbf{Edge features:} A binary covered flag
    $\in \{0,1\}$ indicating whether an edge is currently covered by at
    least one of its incident selected nodes.

    \item \textbf{Reset:} All nodes are initialized as unselected and all
    edges as uncovered.

    \item \textbf{Step helper (non-revisiting):} Selecting an unselected node
    adds it permanently to the cover, removes it from the action space, and
    marks all newly covered incident edges as covered. 

    \item \textbf{Step helper (revisiting):} When revisiting is enabled, an
    action can be applied to any node, including nodes that have already been
    selected. The action \emph{toggles} the node state: selecting an
    unselected node adds it to the cover, while selecting an already selected
    node removes it from the cover. 

    \item \textbf{Objective and reward:} At each step, the objective value is computed both before and after the node toggle. The reward is defined as the difference in the size of the cover relative to the previous timestep. Consequently, a positive reward indicates an improvement in the current solution, whereas a negative reward indicates a deterioration.

    \item \textbf{Termination:} In the constructive setting,
    the episode terminates as soon as all edges are covered, since a feasible
    vertex cover has been constructed, or when a predefined cutoff is
    reached. In the revisiting setting, the agent may continue toggling nodes to search for a smaller feasible
    cover. Therefore, termination is controlled only by the predefined cutoff.

    \item \textbf{Score:} The score is zero if no feasible cover is found;
    otherwise, the final score is based on the size of the best feasible
    cover found, normalized relative to the empirically known best and worst
    solutions.
\end{itemize}

\subsection{Maximum Cut}

\textit{Logic:} Partition nodes into two sets to maximize the total weight of
edges crossing between partitions.

\begin{itemize}
    \item \textbf{Parameters:} Number of nodes $N$ and edge weight range.

    \item \textbf{Graph:} A weighted graph with edge weights randomly sampled
    across node pairs. The graph is sparsified by retaining $K=10$ NNs per node to avoid a fully connected graph.

    \item \textbf{Node features:} A binary partition assignment
    $\in \{0,1\}$ indicating the set to which each node belongs.

    \item \textbf{Edge features:} Edge weight as a scalar value.

    \item \textbf{Reset:} All nodes are initially assigned to partition $0$.

    \item \textbf{Step helper (non-revisiting):} A node is assigned to a
    selected partition and then removed from the action space. Each node is
    therefore assigned at most once, resulting in a constructive generation
    of the partition. The reward corresponds to the resulting change in cut
    value.

    \item \textbf{Step helper (revisiting):} When revisiting is enabled, any
    node can be selected repeatedly. Each action toggles the node between
    the two partitions. The cut value is updated according to the change in
    the crossing edges caused by the move, allowing previously made
    assignments to be reversed.

    \item \textbf{Objective and reward:} The objective is the current cut
    value $C_t$, and the reward is the incremental change with respect to the previous timestep.
    Thus, a positive reward corresponds to an improvement in the cut, while
    a negative reward corresponds to a deterioration. With relative rewards,
    this change is normalized by the difference between the empirically known
    best and worst cut values.

    \item \textbf{Termination:} In the constructive setting, the episode
    terminates once all nodes have been assigned, since a complete partition
    has been constructed, or when the empirically known best cut is reached.
    In the revisiting setting, node assignments are reversible and the agent
    can continue modifying the partition to improve the cut. Therefore,
    termination is controlled by the search horizon, unless the empirically
    known best cut is reached.

    \item \textbf{Score:} The score is based on the best cut value obtained
    during the episode, normalized relative to the empirically known best and
    worst solutions.
\end{itemize}

\subsection{Virtual Machine Placement}
\textit{Logic:} Assign virtual machines (VMs) to physical machines (PMs) to
jointly optimize utilization, energy efficiency, packing efficiency, load
balancing, and security risk (representing the attack surface), subject to
capacity constraints.

\begin{itemize}
    \item \textbf{Parameters:} Number of VMs, PMs, tenants; VM resource
    demands (CPU, MIPS, RAM, storage); PM capacities; traffic, latency,
    energy, and security characteristics.

    \item \textbf{Graph:} Nodes represent VMs and PMs; edges encode VM--PM
    allocations, VM-to-VM traffic, and PM-to-PM latency.

    \item \textbf{Node features:} VM nodes encode resource requirements,
    while PM nodes encode utilization and power consumption.

    \item \textbf{Edge features:} Traffic demands and latency values.

    \item \textbf{Reset:} The system is initialized with the empirically
    known worst allocation to encourage exploration of action sequences
    that improve the allocation.

    \item \textbf{Step helper (non-revisiting):} A $(VM,PM)$ assignment is
    validated against capacity constraints and applied if feasible. Once a
    VM has been moved, it is removed from the action space and cannot be
    selected again, resulting in a constructive allocation process.

    \item \textbf{Step helper (revisiting):} When revisiting is enabled, a VM
    remains in the action space after being moved and can subsequently be
    migrated to another PM. Each valid migration updates the allocation and
    the resulting system metrics, allowing previously made placement
    decisions to be reversed or refined.

    \item \textbf{Objective and reward:} The reward is based on the variation
    of the weighted normalized system metrics between consecutive
    allocations, with an additional penalty for migration cost. Thus, a
    positive reward corresponds to an improvement in the current allocation,
    while a negative reward corresponds to a deterioration or costly
    migration.

    \item \textbf{Termination:} In the constructive setting, the episode
    terminates when no valid VM assignments remain or when the empirically
    known best score is reached. In the revisiting setting, VMs remain
    available for migration, allowing the agent to iteratively improve the
    allocation. Therefore, termination is controlled by the exploration
    horizon, unless the empirically known best score is reached.

    \item \textbf{Score:} The score is the weighted sum of the final
    normalized metric values, with the best allocation encountered during
    revisiting retained as the solution.
\end{itemize}

\subsection{Cyber-Attack Path Prediction}
\textit{Logic:} Predict critical multi-step attack paths in a networked
environment by sequentially exploiting vulnerabilities distributed across
interconnected host nodes. The agent models an attacker operating under
partial observability, incrementally discovering the network topology as the
attack progresses. Formulated as a Partially Observable MDP, this setting
captures the uncertainty inherent to reconnaissance and lateral movement.

\begin{itemize}
    \item \textbf{Parameters:} Number of nodes, vulnerabilities per node,
    communication probability distribution (modeling possible remote
    exploitation), visibility probability distribution, detection
    probability distribution (modeling failure rates), and attack-related
    parameters.

    \item \textbf{Graph:} A dynamic attack graph in which nodes and
    connections become progressively available to the agent. Edges represent
    attack actions that have already been executed, including the
    vulnerabilities exploited between source and target nodes.

    \item \textbf{Node features:} Aggregated service/vulnerability BERT
    embeddings using order- and size-invariant poolings, together with
    visibility, compromise status, privilege level, data presence,
    exfiltration status, persistence, DoS, and defense-evasion flags.

    \item \textbf{Edge features:} Vulnerability BERT embeddings representing
    the vulnerabilities used by the selected attack actions. Multiple
    vulnerabilities exploited between the same source--target pair are
    pooled under the corresponding edge.

    \item \textbf{Reset:} The episode starts from a randomly compromised node
    with partial visibility of its neighborhood, representing an attacker
    with incomplete knowledge of the network.

    \item \textbf{Step helper:} At each step, the agent selects a
    $(source,target,vulnerability)$ action. The action is validated against the current exploitability and
    privilege constraints. If successful, the corresponding node state is
    updated and the attack graph and available action space may expand as
    new information is discovered.

    \item \textbf{Reward:} Outcome-driven rewards reflect the attacker's
    objective under a "control" threat model. Successful outcomes such as
    credential access or lateral movement provide positive rewards, while
    unsuccessful or defense-related outcomes can provide negative rewards.
    Repeating an outcome that has already been achieved provides no additional reward.

    \item \textbf{Termination:} The episode terminates according to the
    attack-specific termination condition, such as compromising all target
    nodes, or when a predefined cutoff is reached.

    \item \textbf{Score:} The score is the percentage of nodes successfully
    compromised during the attack trajectory.

    \item \textbf{Revisiting strategy:} Not supported as a separate
    revisiting strategy because the benchmark is inherently sequential.
    Each action represents the next step of an evolving attack path under
    the current network state and partial observability, rather than a
    reversible modification of a previously constructed solution.
\end{itemize}


\subsection{OSPF Engineering}
\textit{Logic:} Adjust link weights to minimize congestion by reducing maximum
link utilization under routing constraints.

\begin{itemize}
    \item \textbf{Parameters:} Number of nodes, capacity range, traffic
    range, weight bounds, graph distribution, and Equal-Cost Multi-Path
    (ECMP) routing flag.

    \item \textbf{Graph:} A spanning graph with undirected communication
    edges and directed traffic flows.

    \item \textbf{Node features:} Aggregated incoming and outgoing traffic.

    \item \textbf{Edge features:} Capacity, utilization, weight, and traffic
    load.

    \item \textbf{Reset:} The system is initialized with the empirically
    known worst-case weight configuration to encourage exploration of action
    sequences that improve the routing configuration.

    \item \textbf{Step helper (non-revisiting):} A link is selected and
    assigned a final weight within the allowed bounds. Once modified, the
    link is marked as visited and cannot be selected again, resulting in a
    constructive weight configuration. Routing and link utilization are
    recomputed after each modification, according to the selected ECMP
    setting.

    \item \textbf{Step helper (revisiting):} When revisiting is enabled, a
    link can be selected repeatedly and its weight can be increased or
    decreased by a specified variation. The resulting weight is kept within
    the allowed bounds, after which routing and link utilization are
    recomputed. This allows previously made weight adjustments to be
    reversed or refined.

    \item \textbf{Objective and reward:} The reward is based on the change in
    maximum link utilization and, when applicable, the change in routing
    feasibility. For relative rewards,
    \[
        r_t =
        \alpha\left(F_t-F_{t-1}\right)
        +\beta\frac{U_{t-1}-U_t}
        {U_{\mathrm{initial}}-U_{\mathrm{best}}},
    \]
    where $F_t$ is the feasible traffic percentage and $U_t$ is the
    aggregated link utilization. Thus, a positive reward corresponds to an
    improvement in feasibility and/or a reduction in link utilization.

    \item \textbf{Termination:} In the constructive setting, the episode
    terminates once all communication links have been assigned a weight or
    when the empirically known best utilization is reached. In the
    revisiting setting, links remain available for further weight
    adjustments, allowing the agent to iteratively improve the routing
    configuration. Termination therefore occurs when the target
    configuration is reached or according to the predefined exploration
    horizon.

    \item \textbf{Score:} The score is based on the reduction in maximum link
    utilization relative to the initial configuration, while accounting
    for routing feasibility.
\end{itemize}

\subsection{Traffic Engineering}
\textit{Logic:} Route traffic demands through feasible paths to minimize
congestion while respecting capacity constraints. This represents a variant
of OSPF engineering without relying on OSPF as an intermediary helper.

\begin{itemize}
    \item \textbf{Parameters:} Number of nodes, capacity range, traffic
    characteristics, graph distribution, and maximum path length, which
    restricts the set of communication paths considered as candidate
    solutions.

    \item \textbf{Graph:} A spanning graph with communication and traffic
    edges.

    \item \textbf{Node features:} Aggregated incoming and outgoing traffic.

    \item \textbf{Edge features:} Capacity, utilization, and traffic load.

    \item \textbf{Reset:} The system is initialized with the empirically
    known worst-case allocation configuration to encourage exploration of
    action sequences that improve the routing configuration.

    \item \textbf{Step helper (non-revisiting):} A traffic demand is assigned
    to a feasible communication path. Once selected, the traffic demand is
    marked as visited and cannot be selected again, resulting in a
    constructive routing process.

    \item \textbf{Step helper (revisiting):} When revisiting is enabled, a
    traffic demand can be selected repeatedly and assigned to a different
    communication path. The previously allocated path is deallocated before
    the new path is assigned, allowing previously made routing decisions to
    be reversed or refined.

    \item \textbf{Objective and reward:} The reward is based on the change in
    routing feasibility and link utilization between consecutive
    allocations. For relative rewards,
    \[
        r_t =
        \alpha
        \frac{F_t-F_{t-1}}{F_{t-1}}
        +
        \beta
        \frac{U_{t-1}-U_t}
        {U_{\mathrm{initial}}-U_{\mathrm{best}}},
    \]
    where $F_t$ is the feasible traffic percentage and $U_t$ is the
    aggregated link utilization. Thus, a positive reward corresponds to
    improved feasibility and/or reduced link utilization.

    \item \textbf{Termination:} In the constructive setting, the episode
    terminates once all traffic demands have been assigned or when the
    empirically known best routing is reached. In the revisiting setting,
    traffic demands remain available for rerouting, allowing the agent to
    iteratively improve the allocation. Termination therefore occurs when
    the target routing quality is reached or according to the predefined
    exploration horizon.

    \item \textbf{Score:} The score is based on the reduction in maximum
    link utilization relative to the initial configuration, while
    accounting for routing feasibility.
\end{itemize}




\section{Hyper-parameter optimization}
\label{chap:hyperparams}
\begin{table*}[h]
\centering
\scriptsize
\setlength{\tabcolsep}{3pt}
\begin{tabular}{p{2cm}p{1cm}p{0.6cm}p{0.6cm}p{0.7cm}p{0.6cm}p{0.65cm}p{0.55cm}p{0.9cm}p{0.8cm}p{0.9cm}p{0.9cm}p{1.05cm}}

\toprule
 & Learning Rate & Batch Size & $\gamma$ & N Steps & Ent Coef & Max Grad Norm & Tau & Target Update Interval & Num Layers & NN Channels & Out Channels & Activation \\
\midrule

\textbf{Discrete PPO}
& 0.00001, 0.0001, \textbf{0.001}
& 32, \textbf{64}, 128
& \textbf{0.9}, 0.95, 0.99
& 512, 1024, \textbf{2048}
& \textbf{0.01}, 0.1, 0.2
& 0.1, 0.3, \textbf{0.5}
& --
& --
& --
& --
& --
& -- \\
\midrule

\textbf{Projection PPO}
& 0.00001, 0.0001, \textbf{0.001}
& 32, \textbf{64}, 128
& \textbf{0.9}, 0.95, 0.99
& 512, 1024, \textbf{2048}
& \textbf{0.01}, 0.1, 0.2
& 0.1, 0.3, \textbf{0.5}
& --
& --
& --
& --
& --
& -- \\
\midrule

\textbf{Iterative IDQN}
& 0.00001, 0.00005, \textbf{0.0001}
& \textbf{32}, 64, 128
& 0.9, \textbf{0.95}, 0.99
& 1, 3, \textbf{5}
& --
& --
& 0.01, \textbf{0.05}, 0.1
& 2000, 5000, \textbf{10000}
& --
& --
& --
& -- \\
\midrule

\textbf{GAE}
& 0.00005, 0.0001, 0.001, \textbf{0.01}, 0.05
& 16, \textbf{32}, 64
& --
& --
& --
& --
& --
& --
& \textbf{2}, 3
& 16, \textbf{32}
& \textbf{16}, 32, 64
& LeakyReLU, \textbf{ReLU}, null \\
\bottomrule

\end{tabular}
\caption{Hyperparameter ranges explored for each method with best configurations in \textbf{bold}.}
\label{tab:hyperopt_updated}
\end{table*}

Table \ref{tab:hyperopt_updated} reports the hyperparameter search space explored for each optimized model, along with the best configuration identified. The hyperparameter optimization was conducted exclusively on the \textit{TSP} benchmark for simplicity, using 25 trials for each model. Regarding discrete solutions, this process has been performed only on the P-Discrete-M variant for simplicity, and hyperparameters were used across all other discrete variants.

The following additional hyperparameters were tuned manually and selected based on empirical evaluation:

\begin{itemize}

\item \textbf{Node embeddings dimensionality:} 16

\item \textbf{Nearest neighbor search (action space):} FAISS Flat Index

\item \textbf{Distance metric (action space search):} Cosine similarity

\item \textbf{Action vector processing:}
\begin{itemize}
    \item No normalization in the iterative approach
    \item Z-score normalization in the projection approach (ensures projection outputs remain within a suitable range for policy-based methods)
\end{itemize}

\item \textbf{Action space bounds:} Extended by a margin of $(-1, +1)$ (to the z-score bounds) relative to the empirical latent space range discovered across the training instances per benchmark

\item \textbf{Episode length:} To maintain consistency across varying problem scenarios, the episode length is scaled proportionally to the instance complexity. Concretely, it is defined as a coefficient multiplied by the number of decision variables (e.g., number of VMs in \textit{Placement}), ensuring that agents are given sufficient interaction steps with respect to scenario size:
\begin{itemize}
    \item Default: $1 \times$ scenario size
    \item Cyber-Path / traffic scenarios: $3 \times$ scenario size
    \item OSPF engineering scenarios: $2 \times$ scenario size
    \item Maxcut scenarios: $2 \times$ scenario size
\end{itemize}

\item \textbf{Reward normalization:} Min-max normalization (Stable-Baselines3 default)

\item \textbf{Feature normalization:}
\begin{itemize}
    \item Node features are normalized independently using min-max scaling based on their respective ranges within the graph, ensuring feature scale independence
    \item L1 normalization applied to language model embeddings when used as features, as ranges can hardly be determined
\end{itemize}

\item \textbf{Scenario switching during training (only for \textit{V} experiments):}
\begin{itemize}
    \item Interval: every 50 episodes
    \item Strategy: random selection
\end{itemize}

\item \textbf{Policy architecture:}
\begin{itemize}
    \item Hidden layers: [128, 64]
    \item Subsequent output layer proportional to the action space and agent formulation used 
    \item Activation: LeakyReLU
    \item Optimizer: Adam ($\epsilon = 10^{-7}$, weight decay $= 10^{-4}$, AMSGrad disabled)
\end{itemize}

\item \textbf{Graph Autoencoder (GAE):}
\begin{itemize}
    \item Final layer: no activation function, with normalization enabled to ensure properly scaled embeddings
    \item Loss functions (all adjusted to have the same scale, and weights all set to 1):
    \begin{itemize}
        \item Continuous features: Mean Squared Error (MSE)
        \item Adjacency matrix: Contrastive loss ($\tau = 0.5$)
        \item Binary features: Binary Cross-Entropy with logits
        \item Multi-categorical features: Cross-Entropy loss
    \end{itemize}
\end{itemize}

\end{itemize}

\noindent

\paragraph{Seeds.} The random seeds used in the experiments are defined as follows. A seed of 42 is used for the scenario generation, initial experimental setup, and splitting strategies. For the generalization study, five independent runs per strategy and RL solution are performed with the following seeds used during training: 42, 100, 123, 200, and 300. For the testing phase, seeds are generated based on the number of episodes per scenario instance, ensuring a one-to-one correspondence and symmetry across test episode identifiers of several scenario instances. The resulting sequence of test seeds is: 42, 100, 123, 200, followed by increments of 100 up to 5000 (i.e., 400, 500, 600, \dots, 5000).
The same sequence is used for assessing the distribution of action-selection time.

All other hyperparameters were set to their respective library default values and are provided in the accompanying configuration files.

\section{Scenario generation}
\label{chap:scenario_generation}
As described in Section~\ref{chap:generalization}, the generalization study relies on 101 generated instances obtained by randomly sampling the parameters of the environment within predefined ranges. These ranges are designed to induce variability across instances and are reported below; all other parameters are held constant as indicated.

\setlength{\columnsep}{0.5cm}
\setlength{\parskip}{0pt}

\begin{multicols}{3}

\textbf{Traveling Salesman Problem}\\
\texttt{num\_cities}:[10,100]\\
\texttt{max\_coord}:[100,1000]

\columnbreak

\textbf{Maximum Cut}\\
\texttt{num\_nodes}:[10,100]\\
\texttt{max\_weight}:[10,100]

\columnbreak

\textbf{Minimum Vertex Cover}\\
\texttt{num\_nodes}:[10,50]\\
\texttt{edge\_prob}:[0.1,0.4]

\end{multicols}

\begin{multicols}{2}

\textbf{OSPF Engineering}\\
\texttt{num\_nodes}:[10,30]\\
\texttt{communication\_edge\_ratio}:[0.1,0.3]\\
\texttt{non\_zero\_traffic\_ratio}:[0.1,0.3]\\
\texttt{min\_capacity}:[10,100]\\
\texttt{max\_capacity}:[500,1000]\\
\texttt{max\_traffic}:[25,50]\\
\texttt{graph\_distribution}: \texttt{spanning tree}\\
\texttt{min\_weight}: 1\\
\texttt{max\_weight}: 5

\columnbreak

\textbf{Traffic Engineering}\\
\texttt{num\_nodes}:[10,25]\\
\texttt{communication\_edge\_ratio}:[0.1,0.2]\\
\texttt{non\_zero\_traffic\_ratio}:[0.1,0.2]\\
\texttt{min\_capacity}:[10,100]\\
\texttt{max\_capacity}:[500,1000]\\
\texttt{max\_traffic}:[25,50]\\
\texttt{max\_path\_len}: 4\\
\texttt{graph\_distribution}: \texttt{spanning tree}

\end{multicols}

\begin{multicols}{2}

\textbf{Virtual Machine Placement}\\
\texttt{n\_vms}:[10,50]\\
\texttt{n\_pms}:[10,50]\\
\texttt{n\_tenants}:[2,5]\\
\texttt{vm\_vuln\_prob\_min}:[0.01,0.1]\\
\texttt{vm\_vuln\_prob\_max}:[0.2,1.0]\\
\texttt{pm\_escape\_prob\_min}:[0.01,0.1]\\
\texttt{pm\_escape\_prob\_max}:[0.2,0.33]\\
\texttt{pm\_capacity\_memory\_min}:[32,64]\\
\texttt{pm\_capacity\_memory\_max}:[128,512]\\
\texttt{vm\_demand\_memory\_min}:[1,8]\\
\texttt{vm\_demand\_memory\_max}:[16,24]\\
\texttt{pm\_capacity\_storage\_min}:[100,500]\\
\texttt{pm\_capacity\_storage\_max}:[1000,5000]\\
\texttt{vm\_demand\_storage\_min}:[10,50]\\
\texttt{vm\_demand\_storage\_max}:[60,100]\\
\texttt{pm\_capacity\_pe\_min}:[8,16]\\
\texttt{pm\_capacity\_pe\_max}:[32,128]\\
\texttt{vm\_demand\_pe\_min}:[2,4]\\
\texttt{vm\_demand\_pe\_max}:[8,16]\\
\texttt{pm\_capacity\_mips\_min}:[1000,5000]\\
\texttt{pm\_capacity\_mips\_max}:[10000,50000]\\
\texttt{vm\_demand\_mips\_min}:[100,500]\\
\texttt{vm\_demand\_mips\_max}:[750,1000]\\
\texttt{latency\_min}:[0.1,1.0]\\
\texttt{latency\_max}:[2,8]\\
\texttt{p\_idle}:[50,100]\\
\texttt{p\_peak}:[150,300]\\
\texttt{coefficients}: 1\\
\texttt{min\_traffic}: 1\\
\texttt{max\_traffic}: 10\\
\texttt{traffic\_density}: 0.8\\

\columnbreak

\textbf{Cyber-Attack Path Prediction}\\
\texttt{n\_nodes}:[10,20]\\
\texttt{n\_vulns\_per\_node}:5\\
\texttt{vulns\_overlap}:[0.0,0.1]\\
\texttt{p\_data\_present}:[0.6,0.9]\\
\texttt{p\_feature\_visible}:[0.5,0.7]\\
\texttt{p\_recon}:[0.2,0.4]\\
\texttt{p\_detection}:[0.05,0.2]\\
\texttt{goal}: control\\
\texttt{fully\_connected}: true\\
\texttt{outcome\_selection}: false\\

\end{multicols}

Scalability experiments present an evolving set of values for the main dimension (number of nodes) and the same ranges for secondary features.

Scenario instances are exhaustively evaluated using heuristic-based exploration to estimate empirical upper and lower bounds. The number of sweeps is set based on preliminary experiments (10000 for TSP, 2000 for MaxCut, 10000 for MVC, 5000 for Placement, 2000 for OSPF, and 2000 for Traffic). The Cyber-Path scenario is omitted as its bounds can be directly inferred from the number of nodes (which becomes the maximum relative score). The heuristics guiding exploration were designed using a language model (gpt-5.4) and refined through manual validation.

\begin{itemize}
    \item \textit{TSP}: We generate random tours and iteratively refine them using stochastic 2-opt moves. Improvements (or degradations) are sampled from candidate swaps, biased by their impact on tour length. This allows simultaneous estimation of both best and worst-case tour costs. 
    \item \textit{MinVertex}: A greedy-stochastic strategy is used to iteratively select nodes covering the largest number of uncovered edges. At each step, selection is randomized among the top-k candidates to maintain diversity. A pruning phase removes redundant nodes to refine solutions. The best cover size is tracked across multiple sweeps. The worst is set to the overall number of nodes.
    \item \textit{MaxCut}: Random graph partitions are initialized and improved via stochastic local search. Node flips that increase the cut value are identified, and one is selected probabilistically based on gain. This process continues until no further improvements are possible. The maximum cut value observed across samples defines the upper bound. The worst value is set to 0.
    \item \textit{Placement}: Allocations are constructed using heuristics that either maximize or balance resource utilization across physical machines. Initial solutions are built by prioritizing high-demand VMs, followed by local refinements through stochastic reassignment. Both best and worst allocations are identified through repeated sweeps.
    \item \textit{OSPF}: OSPF link-weight settings are explored through stochastic search over candidate configurations. Each sweep starts either from a random solution or from a perturbed elite solution sampled from small best/worst pools, allowing intensified search without losing diversity. Neighboring configurations are obtained by randomly modifying a subset of link weights, and are selected with a bias toward lower and higher utilization to estimate both worst and best. Previously explored configurations are excluded.
    \item \textit{Traffic}: Same idea as \textit{OSPF} but optimizing path allocations and not edge weights.
\end{itemize}

\section{Embedding action space plots}
\label{chap:action_space}
Figure \ref{fig:action_space_appendix} presents the UMAP projection of the latent action spaces for seven representative 20-node scenarios from each real-world benchmark, highlighting both their variability and the common structural patterns that emerge in the action space manifolds. The scenarios were manually selected as the most representative manifolds based on exploratory analysis. Classical benchmarks were excluded due to their limited diversity in terms of manifolds, with respect to the already provided plots in Section \ref{chap:experiments}.
\begin{center}
\captionsetup{type=figure}
\setlength{\tabcolsep}{2pt}

\textbf{Virtual Machine Placement}\\[0.3em]
\begin{tabular}{ccccccc}
\includegraphics[width=0.13\textwidth]{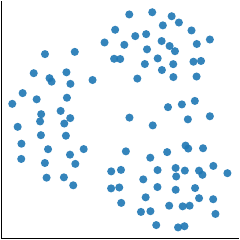} &
\includegraphics[width=0.13\textwidth]{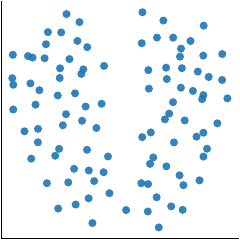} &
\includegraphics[width=0.13\textwidth]{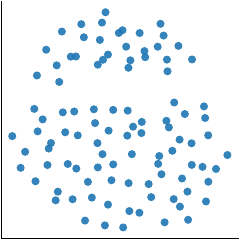} &
\includegraphics[width=0.13\textwidth]{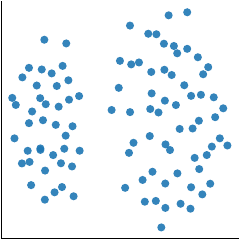} &
\includegraphics[width=0.13\textwidth]{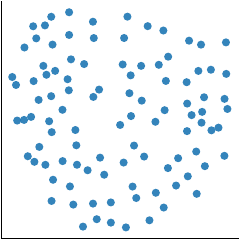} &
\includegraphics[width=0.13\textwidth]{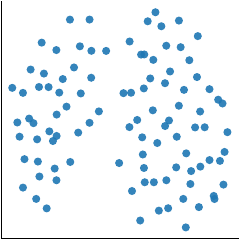} &
\includegraphics[width=0.13\textwidth]{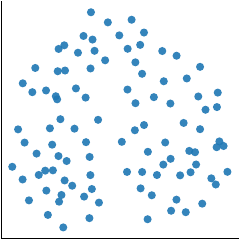} \\
\end{tabular}

\textbf{Cyber Attack Path Prediction}\\[0.3em]
\begin{tabular}{ccccccc}
\includegraphics[width=0.13\textwidth]{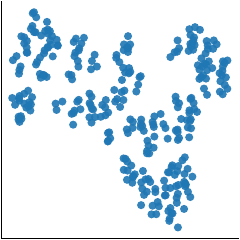} &
\includegraphics[width=0.13\textwidth]{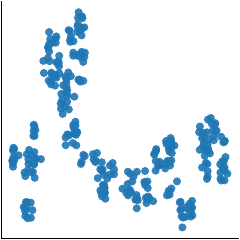} &
\includegraphics[width=0.13\textwidth]{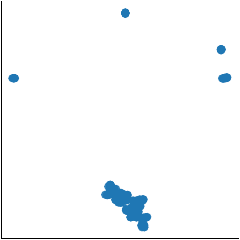} &
\includegraphics[width=0.13\textwidth]{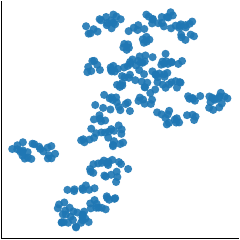} &
\includegraphics[width=0.13\textwidth]{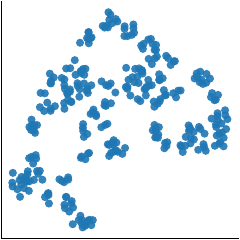} &
\includegraphics[width=0.13\textwidth]{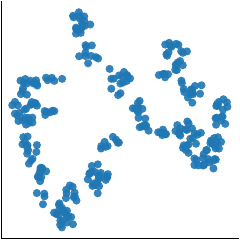} &
\includegraphics[width=0.13\textwidth]{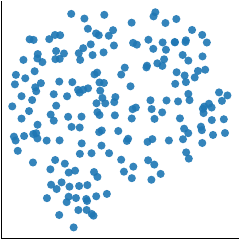} \\
\end{tabular}

\textbf{OSPF Engineering}\\[0.3em]
\begin{tabular}{ccccccc}
\includegraphics[width=0.13\textwidth]{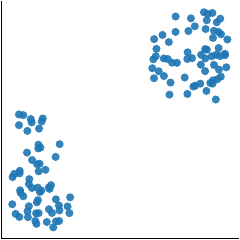} &
\includegraphics[width=0.13\textwidth]{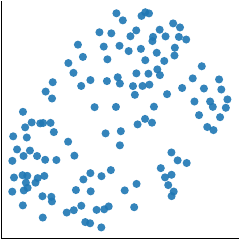} &
\includegraphics[width=0.13\textwidth]{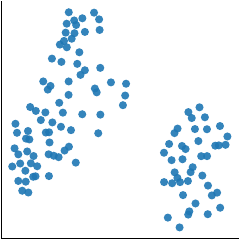} &
\includegraphics[width=0.13\textwidth]{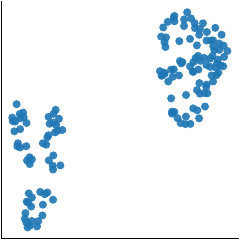} &
\includegraphics[width=0.13\textwidth]{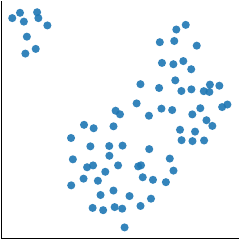} &
\includegraphics[width=0.13\textwidth]{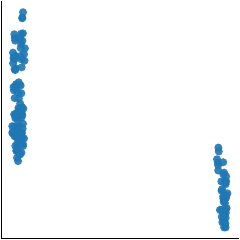} &
\includegraphics[width=0.13\textwidth]{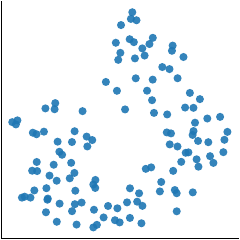} \\
\end{tabular}

\textbf{Traffic Engineering}\\[0.3em]
\begin{tabular}{ccccccc}
\includegraphics[width=0.13\textwidth]{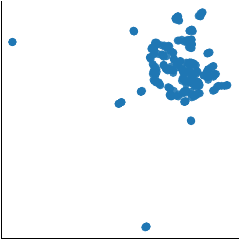} &
\includegraphics[width=0.13\textwidth]{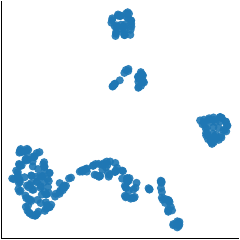} &
\includegraphics[width=0.13\textwidth]{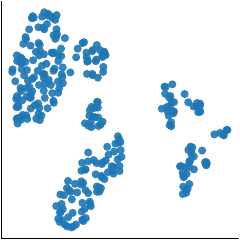} &
\includegraphics[width=0.13\textwidth]{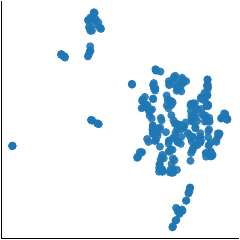} &
\includegraphics[width=0.13\textwidth]{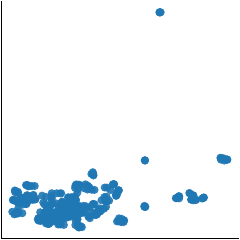} &
\includegraphics[width=0.13\textwidth]{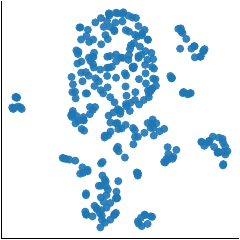} &
\includegraphics[width=0.13\textwidth]{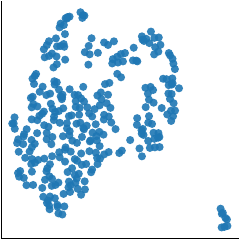} \\
\end{tabular}

\captionof{figure}{UMAP projections of action spaces for seven 20-node graph instances across the four real-world benchmarks, with up to 300 actions (sampled by maintaining a representative approximation of the underlying action distribution).}
\label{fig:action_space_appendix}
\end{center}

\section{Graph Representation Learning Ablations}
\label{chap:ablation_GNN}

We conduct ablations to assess the role of GNN-based representations in the action embedding space used by the projection agent. 

\subsection{Action Space Ablation}

We study how the representation of the action space affects the projection agent. We compare three representations: randomly initialized vectors, the raw feature vector of each node, and GNN-derived embeddings for benchmarks with nodes as actions (TSP, MinVertex, MaxCut). All variants use the same embedding dimensionality and projection-agent architecture, and are evaluated under the smallest training regime (\(S\)) using the same protocol as in Section~\ref{chap:generalization}.

Table~\ref{tab:ablation_action_embeddings} reports the results. We report \(\Delta_{\mathrm{train}}\), the change from training to test performance, and \(\Delta_{\mathrm{GNN}}\), the difference relative to the GNN-based representation. The random-vector variant performs substantially worse across all problems, showing that the projection agent benefits from semantically meaningful action representations rather than merely from placing action anchors in a continuous space. Using raw node features recovers much of this performance, demonstrating that node semantics alone already provide useful information. However, the GNN representation performs better, indicating that there is advantage in enriching node features with information about the surrounding graph structure for these benchmarks.

\subsection{Ablation of GAE Loss Weights}

We further investigate the contribution of the individual reconstruction objectives used to pre-train the GAE. As described in Section~\ref{chap:GNNRepresentation}, the self-supervised objective jointly reconstructs node features, edge features, and the adjacency matrix. Since these objectives are optimized jointly within a limited-dimensional embedding space, they may impose competing constraints on the learned representation. We therefore conduct a two-stage ablation to first verify that each objective captures its intended graph information, and then assess whether this information benefits downstream RL-GCO.

\paragraph{Reconstruction ablation.}
We first evaluate whether each reconstruction objective effectively contributes to learning its corresponding graph information, independently of the RL downstream task. We train four GAE variants: the full model and three ablated versions, each removing one reconstruction objective (adjacency, edge, or node reconstruction). Preliminary results on TSP are reported below, with normalized reconstruction scores. 
Models are pre-trained on 50 graph instances (25 for training and 25 for testing) and evaluated over 50 test episodes. Reconstruction scores are normalized such that 0 corresponds to the average loss of 50 randomly initialized GAEs and 1 to perfect reconstruction (loss = 0).

\begin{table}[t]
\centering
\caption{Projection-agent performance with different action embeddings. 
Random embeddings are compared against embeddings initialized with the 
feature vector of the corresponding node.}
\label{tab:ablation_action_embeddings}
\begin{tabular}{llccc}
\toprule
Problem & Embedding & Score & $\Delta_{\mathrm{train}}$ & $\Delta_{\mathrm{GNN}}$ \\
\midrule
TSP       & Random      & 0.36 & -0.37 & -0.42 \\
          & Node feature & 0.64 & -0.32 & -0.12 \\
\midrule
MinVertex & Random      & 0.36 & -0.57 & -0.34 \\
          & Node feature & 0.21 & -0.19 & -0.59 \\
\midrule
MaxCut    & Random      & 0.36 & -0.57 & -0.57 \\
          & Node feature & 0.79 & -0.19 & -0.14 \\
\bottomrule
\end{tabular}
\end{table}
\begin{table}[t]
\centering
\caption{Ablation of GAE reconstruction objectives on TSP. 
The left block reports normalized reconstruction scores for each objective,
while the right block reports downstream projection-agent performance under
the \(S\) training regime.}
\label{tab:ablation_gae_loss}
\begin{tabular}{lccc|cccc}
\toprule
& \multicolumn{3}{c|}{Self-supervised reconstruction}
& \multicolumn{4}{c}{Downstream optimization} \\
Model & Adj. & Edge & Node
& Mean & IQM & Std & 95\% BCI \\
\midrule
Full model
& 0.04 & 0.97 & 0.48
& 0.571 & 0.566 & 0.115 & [0.561, 0.581] \\
$-$ Adjacency
& 0.03 & 0.97 & 0.48
& 0.593 & 0.601 & 0.109 & [0.584, 0.603] \\
$-$ Edge
& 0.15 & 0.12 & 0.47
& 0.566 & 0.558 & 0.099 & [0.558, 0.575] \\
$-$ Node
& 0.07 & 0.96 & 0.01
& 0.519 & 0.504 & 0.117 & [0.508, 0.529] \\
\bottomrule
\end{tabular}
\end{table}

Results in Table \ref{tab:ablation_gae_loss} (self-supervised reconstruction column) shows how removing each objective primarily degrades reconstruction of the corresponding graph component, confirming that the three objectives encode distinct and relevant aspects of the graph structure. The slight improvements observed in some components (e.g., adjacency after removing edge reconstruction) highlights the trade-off between competing objectives during optimization. 

\paragraph{Impact on RL-GCO.}
We next assess whether these learned signals translate into improved policy performance. Using the smallest training regime (\(S\)), we follow the experimental protocol of Section~\ref{chap:generalization} and evaluate each variant over five runs on TSP. For each ablation, one reconstruction objective is removed while all other components remain unchanged. Results are reported in Table~\ref{tab:ablation_gae_loss} (downstream optimization column).

Removing node or edge reconstruction reduces downstream performance, indicating that the corresponding information is useful for policy learning. Interestingly, removing adjacency reconstruction improves performance. This highlights that some graph information is not necessarily beneficial for the downstream RL objective: jointly reconstructing multiple graph properties can introduce competing constraints within the limited-dimensional latent space. Thus, the reconstruction objective should be viewed as a representation-design choice rather than assuming that incorporating every available graph signal necessarily produces the best policy representation.

\section{Comparison with Other Solvers}
\label{chap:beyond_solvers}
To assess the value of learning heuristics with our RL-GCO methodology and further validate our experimental framework, we compare the projection agent against both classical solvers and alternative learning-based heuristics. We consider two representative benchmarks, MaxCut and TSP, and use the same instances and generalization regimes as in Section~\ref{chap:generalization} to ensure a direct and consistent comparison.

\subsection{Maximum Cut: Beyond RL Solvers}

For MaxCut, we compare the projection agent with PI-GNN~\cite{Schuetz2022PhysicsInformed}, a physics-inspired, unsupervised GNN that provides a representative non-RL, non-autoregressive approach to MaxCut, as well as simulated annealing (SA) as a classical optimization baseline. Since PI-GNN was originally evaluated on unweighted MaxCut, we adapt its methodology to the weighted setting considered in our experiments.

\paragraph{Validation of the normalization reference.}
Our evaluation normalizes solution quality relative to the best solution obtained from a heuristic-driven sweep of 20,000 candidate solutions. To assess whether this provides a reliable reference, we compare the resulting solutions with those obtained by simulated annealing on the same instances. Across 101 scenarios, the two procedures identified identical best solutions in 98 cases. In the remaining three cases, the difference in normalized solution quality was only 0.0002. This close agreement indicates that the heuristic sweep provides a reliable approximation to the SA reference on our MaxCut instances. Thus, the normalized scores reported throughout this section can also be interpreted as relative performance with respect to a strong classical optimization baseline.

\paragraph{Comparison with PI-GNN.}
We evaluate PI-GNN and the projection agent on identical graph instances and train/test splits, using the \(S\), \(M\), and \(L\) generalization regimes defined in Section~\ref{chap:generalization}. We do not consider the \(V\) regime because PI-GNN assumes learning from a single instance. For both methods, we use the same best-of-five evaluation protocol and normalization procedure, allowing for a direct comparison between the two learning paradigms.

PI-GNN is known to be sensitive to hyperparameter selection. We therefore report two configurations: (i) the hyperparameters specified in the original work and (ii) a configuration obtained through hyperparameter optimization on a held-out validation set from our benchmark. This distinction separates the performance of the published configuration from the performance achievable with problem-specific tuning.

\begin{table}[t]
\centering
\caption{Comparison between PI-GNN and the projection agent on MaxCut under the \(S\), \(M\), and \(L\) generalization regimes. PI-GNN is evaluated both with the hyperparameters reported in the original work and after hyperparameter optimization on a held-out validation set.}
\label{tab:maxcut_solver_comparison}
\begin{tabular}{lccc}
\toprule
Method & $S$ & $M$ & $L$ \\
\midrule
PI-GNN (reported) & 0.52 & 0.61 & 0.22 \\
PI-GNN (tuned)    & 0.92 & 0.91 & 0.92 \\
Projection (LaGCO-RL)      & \bestblock{0.95} & \bestblock{0.95} & 0.92 \\
\bottomrule
\end{tabular}
\end{table}

The projection agent achieves the strongest overall performance, outperforming the tuned PI-GNN configuration in the \(S\) and \(M\) regimes and matching it in the \(L\) regime. The substantial improvement obtained by tuning PI-GNN also highlights the sensitivity of this approach to hyperparameter selection. Importantly, the close agreement between the heuristic-sweep reference and simulated annealing provides additional context for interpreting these results: a normalized score of approximately 0.95 corresponds to achieving roughly 95\% of the solution quality obtained by the strong classical SA reference on these instances.

Overall, these results indicate that the projection agent is competitive with a specialized non-RL, non-autoregressive solver on MaxCut, while requiring no autoregressive decoding and remaining robust across the considered generalization regimes.

\subsection{TSP: Beyond RL Solvers}

For TSP, we compare the projection agent with POMO~\cite{POMO}, an RL-based solver that represents a prominent approach for this benchmark. We additionally use Concorde as a reference solver for assessing solution quality against high-quality, and where applicable optimal, solutions.

\paragraph{Validation of the normalization reference.}
As in the other benchmarks, our initial evaluation normalizes solution quality relative to the best solution obtained from a heuristic-driven sweep, using 10,000 candidate solutions per instance. Unlike MaxCut, however, we found that the solutions obtained by this procedure frequently differed from those produced by Concorde. This indicates that, for TSP, the heuristic-sweep reference does not provide a sufficiently reliable proxy for the reference solution quality and makes direct comparison with Concorde less meaningful under the original normalization.

We therefore perform an additional retraining and evaluation using an empirical Concorde-based reference. This rerun ensures that the reported results are directly comparable to the solutions produced by the reference solver and avoids conflating differences in normalization with differences in solver performance.

\paragraph{Comparison with POMO.}
We evaluate POMO and the projection agent on identical graph instances and train/test splits, using the \(S\), \(M\), \(L\), and \(V\) generalization regimes defined in Section~\ref{chap:generalization}. Both methods are evaluated using the same best-of-five protocol and the same normalization procedure within the comparison, replacing only the upper bound estimate.

\begin{table}[t]
\centering
\caption{Comparison between POMO and the projection agent on TSP under the \(S\), \(M\), \(L\), and \(V\) generalization regimes.}
\label{tab:tsp_solver_comparison}
\begin{tabular}{lcccc}
\toprule
Method & $S$ & $M$ & $L$ & $V$ \\
\midrule
POMO    & \bestblock{0.81} & \bestblock{0.81} & \bestblock{0.81} & \bestblock{0.81} \\
Iterative (LaGCO-RL) & 0.72 & 0.69 & 0.67 & 0.71 \\
Projection (LaGCO-RL) & 0.67 & 0.52 & 0.42 & 0.44 \\
\bottomrule
\end{tabular}
\end{table}

In contrast to the MaxCut results, the projection agent does not reach the performance of the specialized POMO-RL solver on TSP. For a comparison within our framework, however, the relevant reference is our iterative baseline, which is the strongest of the considered parameterization approaches on this benchmark. The iterative agent remains below the POMO baseline, with a minimum gap of \(0.04\) in the \(V\) regime. Overall, the comparison with POMO provides a useful indication of the remaining gap to a solver specifically designed for TSP.

\section{Library and data}
\label{chap:LAGCoRL}
The \textit{LaGCO-RL} library is released under the GPL-3.0 License alongside this paper. The associated data repository contains the simulated graph instances, trained models, dataset splits, and hyperparameter configurations. The data is distributed under the CC BY 4.0 open license.

To facilitate reproducibility, the repository includes detailed \texttt{README} and \texttt{REPRODUCIBILITY} documents describing how to use the tool, reproduce the experiments, and access the data.

Figures~\ref{fig:traffic_engineering_env} and~\ref{fig:cyberattack_env} illustrate two concrete instantiations of the framework enabled by the \textit{LaGCO-RL} library, showcasing two examples, respectively \textit{cyber-attack path prediction} and \textit{traffic engineering}, and how the corresponding environments can be instantiated in our framework. These examples provides pseudo-code demonstrating how to define key components, including graph construction, observation and action spaces, attribute specifications, and supporting functions.

Figure~\ref{fig:cyberattack_env} presents a potential choice of framework instantiation for the cybersecurity benchmark. The observation space combines graph-level representations with auxiliary features (e.g., number of nodes), while the action space is compositional, involving the selection of a source node, a target node, and an associated vulnerability. This example highlights the framework’s ability to support structured actions over graph elements and its components in a conditional manner, as well as the integration of learned feature extractors (e.g., BERT-based embeddings) for semantic attributes such as vulnerabilities.
In contrast, Figure~\ref{fig:traffic_engineering_env} demonstrates a more complex setting in which multiple graph views are used to represent different aspects of the same environment instance. Specifically, a communication graph and a traffic graph are jointly defined, enabling the agent to reason over both topological connectivity and traffic flow dynamics from different graph representations. The observation space aggregates information independently from each graph, while the action space combines elements from both representations to form each action (i.e., selecting traffic links and communication paths). This highlights the flexibility of the framework in handling multi-graph environments and cross-graph interactions.

Additionally, the traffic engineering example illustrates the use of pooling operators for different graph components, as well as normalization strategies (e.g., min–max scaling) applied to node attributes to ensure consistent feature scaling across different instances. 

\begin{figure*}
\begin{minipage}[t]{0.99\textwidth}
\textbf{Traffic Engineering}
\begin{tcolorbox}[codebox]
class TrafficEngineeringExt(TrafficEngineering, ContinuousEnv):
\begin{tcolorbox}[
    getgraphs,
    enhanced,
    sharp corners,
    boxrule=0.7pt,
    colback=white,
    top=1pt, bottom=1pt, left=2pt, right=2pt, 
    boxsep=0pt,
    fontupper=\ttfamily\scriptsize
]
\begin{lstlisting}[basicstyle=\ttfamily\scriptsize, aboveskip=0pt, belowskip=0pt, breaklines=true]
def get_graphs(...):
    comm_G = ...
    traffic_G = ...
    return {"comm_G": comm_G, "traffic_G": traffic_G}
\end{lstlisting}
\end{tcolorbox}

\begin{tcolorbox}[
    obsbox,
    enhanced,
    sharp corners,
    boxrule=0.7pt,
    colback=white,
    top=1pt, bottom=1pt, left=2pt, right=2pt, 
    boxsep=0pt,
    fontupper=\ttfamily\scriptsize
]
\begin{lstlisting}[basicstyle=\ttfamily\scriptsize, aboveskip=0pt, belowskip=0pt, breaklines=true]
self._observation_type = { 
    "comm": Graph(poolings=[Mean],
                   graph="comm_G"), 
    "traffic": Graph(poolings=[Mean],
                   graph="traffic_G"), 
    ...
}
\end{lstlisting}
\end{tcolorbox}

\begin{tcolorbox}[
    actbox,
    enhanced,
    sharp corners,
    boxrule=0.7pt,
    colback=white,
    top=1pt, bottom=1pt, left=2pt, right=2pt, 
    boxsep=0pt,
    fontupper=\ttfamily\scriptsize
]
\begin{lstlisting}[basicstyle=\ttfamily\scriptsize, aboveskip=0pt, belowskip=0pt, breaklines=true]
self._action_type = {
    "traffic_link": Edge(graph="traffic_G", pooling=Concat),
    "communication_path": Path(graph="comm_G", pooling=Concat) 
    ...
}
\end{lstlisting}
\end{tcolorbox}

\begin{tcolorbox}[
    attrbox,
    enhanced,
    sharp corners,
    boxrule=0.7pt,
    colback=white,
    top=1pt, bottom=1pt, left=2pt, right=2pt, 
    boxsep=0pt,
    fontupper=\ttfamily\scriptsize
]
\begin{lstlisting}[basicstyle=\ttfamily\scriptsize, aboveskip=0pt, belowskip=0pt, breaklines=true]
self._node_attributes = {
    "outgoing_traffic": Attribute(norm="min_max"),
    ...
}
self._edge_attributes = { ... }
\end{lstlisting}
\end{tcolorbox}

\begin{tcolorbox}[
    funcbox,
    enhanced,
    sharp corners,
    boxrule=0.7pt,
    colback=white,
    top=1pt, bottom=1pt, left=2pt, right=2pt, 
    boxsep=0pt,
    fontupper=\ttfamily\scriptsize
]
\begin{lstlisting}[basicstyle=\ttfamily\scriptsize, aboveskip=0pt, belowskip=0pt, breaklines=true]
def sample_valid_action(...): ...
def is_valid_action(...): ...
\end{lstlisting}
\end{tcolorbox}
\end{tcolorbox}
\end{minipage}

\caption{An instantiation of the automated framework with an example of the definition of the observation and action space, attribute specification, and proper methods to support the traffic engineering benchmark.}
\label{fig:traffic_engineering_env}
\end{figure*}

\begin{figure*}[t]
\centering

\begin{minipage}[t]{1\textwidth}
\textbf{Cyber-Path}
\begin{tcolorbox}[codebox]
class CyberAttackExt(CyberAttack, ContinuousEnv):
\begin{tcolorbox}[
    getgraphs,
    enhanced,
    sharp corners,
    boxrule=0.7pt,
    colback=white,
    top=1pt, bottom=1pt, left=2pt, right=2pt, 
    boxsep=0pt,
    fontupper=\ttfamily\scriptsize
]
\begin{lstlisting}[basicstyle=\ttfamily\scriptsize, aboveskip=0pt, belowskip=0pt, breaklines=true]
def get_graphs(...):
    G = ...
    return {"attack_G": G}
\end{lstlisting}
\end{tcolorbox}

\begin{tcolorbox}[
    obsbox,
    enhanced,
    sharp corners,
    boxrule=0.7pt,
    colback=white,
    top=1pt, bottom=1pt, left=2pt, right=2pt, 
    boxsep=0pt,
    fontupper=\ttfamily\scriptsize
]
\begin{lstlisting}[basicstyle=\ttfamily\scriptsize, aboveskip=0pt, belowskip=0pt, breaklines=true]
self._observation_type = { 
    "graph": Graph(poolings=[Mean, Sum], graph="attack_G"), 
    "nodes_num": Function(get_num_nodes, graph="attack_G"),
    ...
}
\end{lstlisting}
\end{tcolorbox}
\begin{tcolorbox}[
    actbox,
    enhanced,
    sharp corners,
    boxrule=0.7pt,
    colback=white,
    top=1pt, bottom=1pt, left=2pt, right=2pt, 
    boxsep=0pt,
    fontupper=\ttfamily\scriptsize
]
\begin{lstlisting}[basicstyle=\ttfamily\scriptsize, aboveskip=0pt, belowskip=0pt, breaklines=true]
self._action_type = {
    "source": Node(graph="attack_G", spec={"controlled": true}),
    "target": Node(graph="attack_G"),
    "vuln": Object(extractor='bert', reference="target", key="vulns")
}
\end{lstlisting}
\end{tcolorbox}

\begin{tcolorbox}[
    attrbox,
    enhanced,
    sharp corners,
    boxrule=0.7pt,
    colback=white,
    top=1pt, bottom=1pt, left=2pt, right=2pt, 
    boxsep=0pt,
    fontupper=\ttfamily\scriptsize
]
\begin{lstlisting}[basicstyle=\ttfamily\scriptsize, aboveskip=0pt, belowskip=0pt, breaklines=true]
self._node_attributes = {
    "vulns": Attribute(extractor='bert', type='continuous')
    ...
}
self._edge_attributes = { ... }
\end{lstlisting}
\end{tcolorbox}


\begin{tcolorbox}[
    funcbox,
    enhanced,
    sharp corners,
    boxrule=0.7pt,
    colback=white,
    top=1pt, bottom=1pt, left=2pt, right=2pt, 
    boxsep=0pt,
    fontupper=\ttfamily\scriptsize
]
\begin{lstlisting}[basicstyle=\ttfamily\scriptsize, aboveskip=0pt, belowskip=0pt, breaklines=true]
def sample_valid_action(...): ... 
def is_valid_action(...): ...
\end{lstlisting}
\end{tcolorbox}
\end{tcolorbox}

\end{minipage}
\caption{An instantiation of the automated framework with an example of the definition of the observation and action space, attribute specification, and proper methods to support the cyber-attack path prediction benchmark.}
\label{fig:cyberattack_env}
\end{figure*}

\section{Observation space}
\label{chap:observation_space}

\paragraph{Latent observation.}
The observation vector used in Section~\ref{chap:experiments} for latent observation agents (G-Discrete, G-Discrete-M, Projection, Iterative) is defined as a fixed-dimensional representation that combines graph-level information and, for the projection agent, a summarized view of the action space. This design ensures sufficient expressiveness while maintaining a compact representation of both the environment structure and the decision space. The observation is constructed as follows:

\begin{itemize}
\item \textbf{Graph embedding:} Node embeddings are aggregated using multiple pooling strategies to capture complementary structural information:
\begin{itemize}
\item Mean pooling
\item Max pooling
\item Min pooling
\item Sum pooling
\end{itemize}

\item \textbf{Structural descriptors:} Additional invariant features are included to explicitly encode global graph properties:
\begin{itemize}
    \item Number of nodes
    \item Number of edges
    \item Average node degree
    \item Graph density
\end{itemize}

\item \textbf{Action-space summarization (projection agent only):}  
For benchmarks with structured action spaces that cannot be fully captured through node embeddings alone (which are already aggregated in the graph embedding), the action space is further embedded and aggregated using:
\begin{itemize}
    \item Mean pooling
    \item Max pooling
    \item Min pooling
    \item Sum pooling
\end{itemize}
The resulting pooled representation is concatenated to the observation vector, providing a compact summary of the latent action space.
This additional vector is provided only to the projection agent, as the iterative agent inherently iterates over all embeddings, resulting in inherent visibility over the latent action space.
\end{itemize}
\paragraph{Padding observation.}
For the padding-based discrete baselines, observations are explicitly constructed to encode all relevant information in a fixed-size vector. These representations are benchmark-specific and capture features at the node, edge, graph, or path level, depending on the task, and are chosen to align with the graph features used for GNN-based embeddings.

The observation design for each benchmark is detailed below:

\begin{itemize}
\item \textbf{TSP.}
The observation concatenates:
\begin{itemize}
\item Flattened node coordinates
\item Binary visited status for each city
\item Flattened upper-triangular distance matrix (pairwise distances)
\end{itemize}

\item \textbf{MinVertex.}  
The observation includes:
\begin{itemize}
    \item Node-level binary indicators specifying whether each node is selected
    \item Edge-level binary indicators (flattened upper triangle) denoting whether edges are covered
\end{itemize}

\item \textbf{MaxCut.}  
The observation consists of:
\begin{itemize}
    \item Node partition assignments
    \item Flattened upper-triangular edge weights
\end{itemize}
This representation captures both the graph structure and the current partitioning.

\item \textbf{Placement.}  
The observation concatenates:
\begin{itemize}
    \item Physical machine (PM) features (e.g., resource capacities and usage)
    \item Virtual machine (VM) features (e.g., resource demands)
\end{itemize}

\item \textbf{Cyber-Path:}  
The observation is constructed as a concatenation of per-node feature vectors, encoding:
\begin{itemize}
    \item Visibility status (feature-level and graph-level)
    \item Compromise state and privilege level
    \item Service and vulnerability counts
    \item Security-relevant flags (e.g., persistence, defense evasion, data exfiltration)
    \item Aggregated vulnerability outcomes
    \item Privilege requirements
\end{itemize}
Features are partially masked depending on visibility, reflecting a partially observable setting, which is also done in the graph representation of the latent approaches.

\item \textbf{OSPF.}  
The observation is built from edge-level features:
\begin{itemize}
    \item For communication links: capacity, routing weight, and used capacity
    \item For traffic demands: traffic volume on each demand edge
\end{itemize}

\item \textbf{Traffic.}  
The observation encodes:
\begin{itemize}
    \item Communication edge features: capacity and used capacity
    \item Traffic demand values for each traffic edge
\end{itemize}

\end{itemize}


\section{Learning curves}
\label{chap:learning_curves}
\begin{figure}[H]
    \centering

    \begin{subfigure}{0.32\textwidth}
        \centering
        \includegraphics[width=\linewidth]{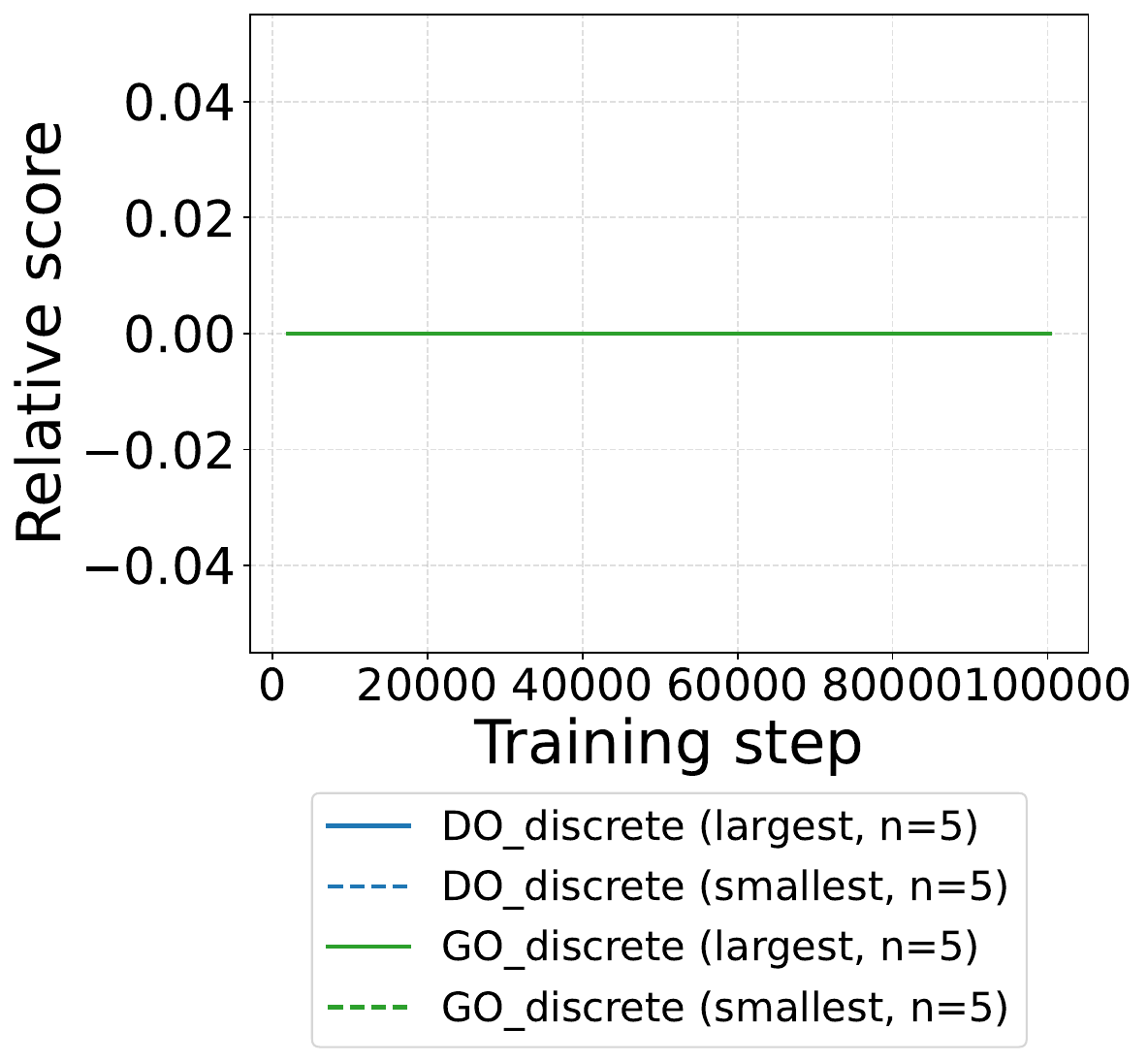}
        \caption{TSP Discrete S/L}
    \end{subfigure}
    \hfill
    \begin{subfigure}{0.32\textwidth}
        \centering
        \includegraphics[width=\linewidth]{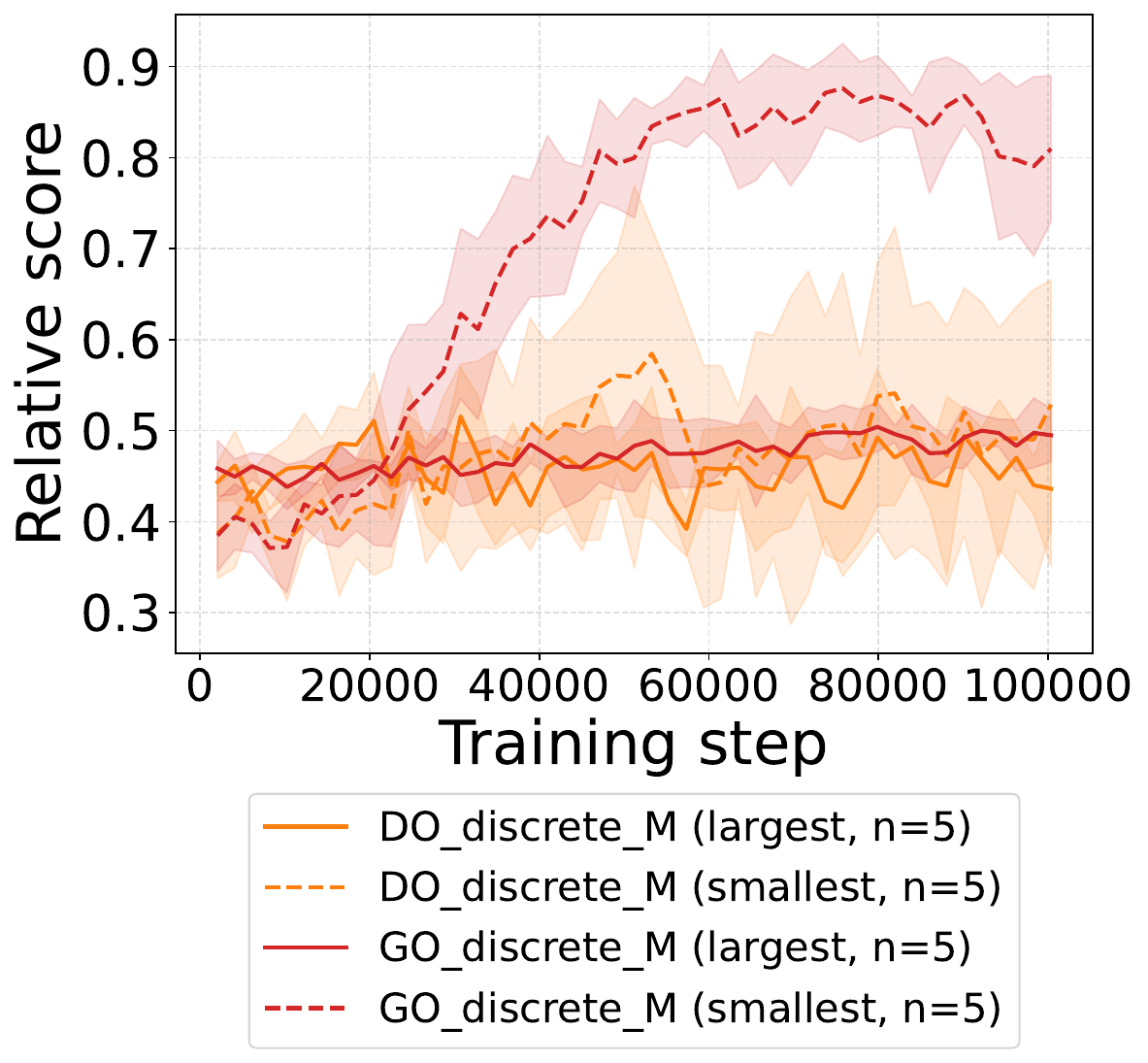}
        \caption{TSP Discrete-M S/L}
    \end{subfigure}
    \hfill
    \begin{subfigure}{0.32\textwidth}
        \centering
        \includegraphics[width=\linewidth]{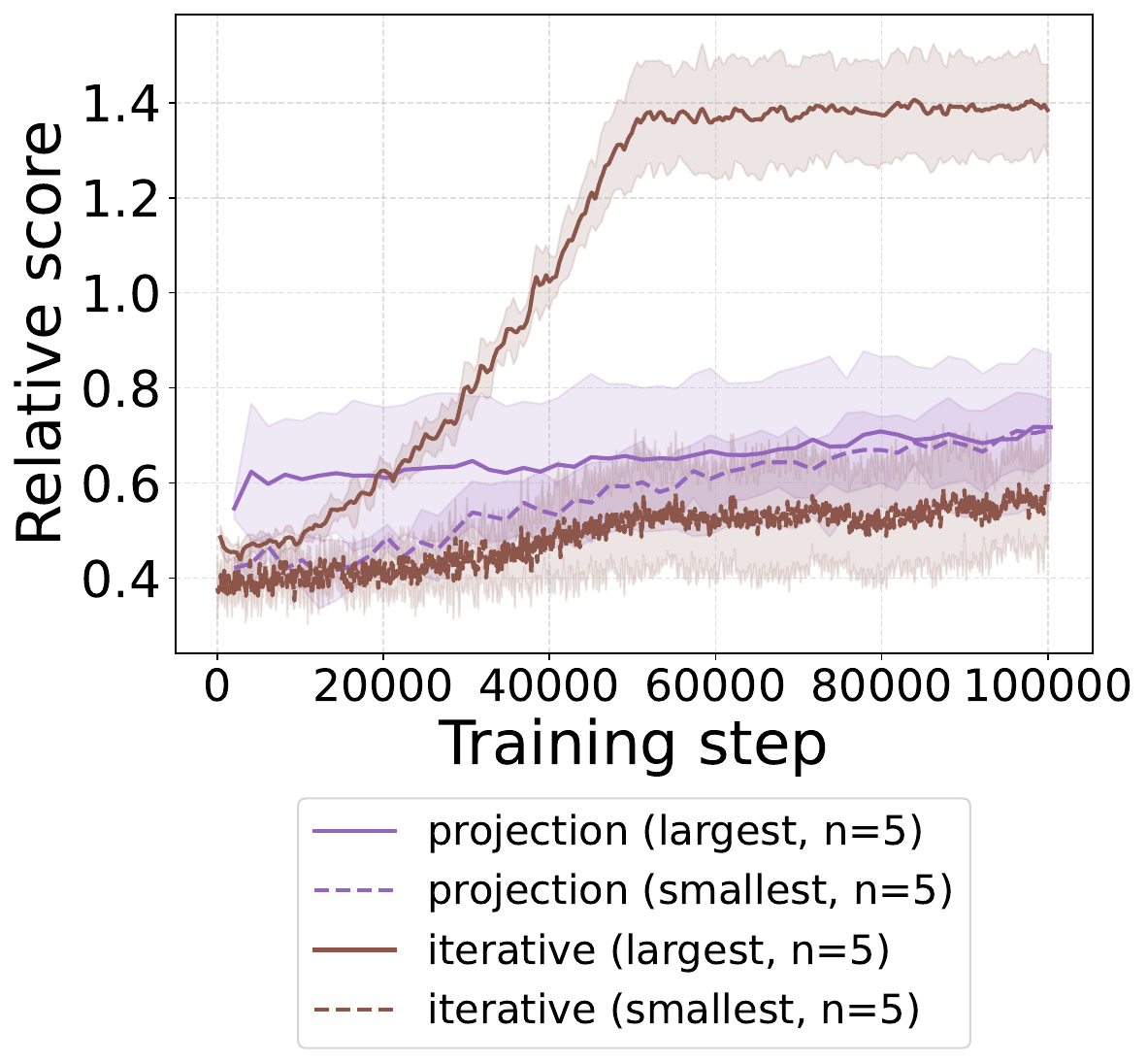}
        \caption{TSP Latent S/L}
    \end{subfigure}

    \begin{subfigure}{0.32\textwidth}
        \centering
        \includegraphics[width=\linewidth]{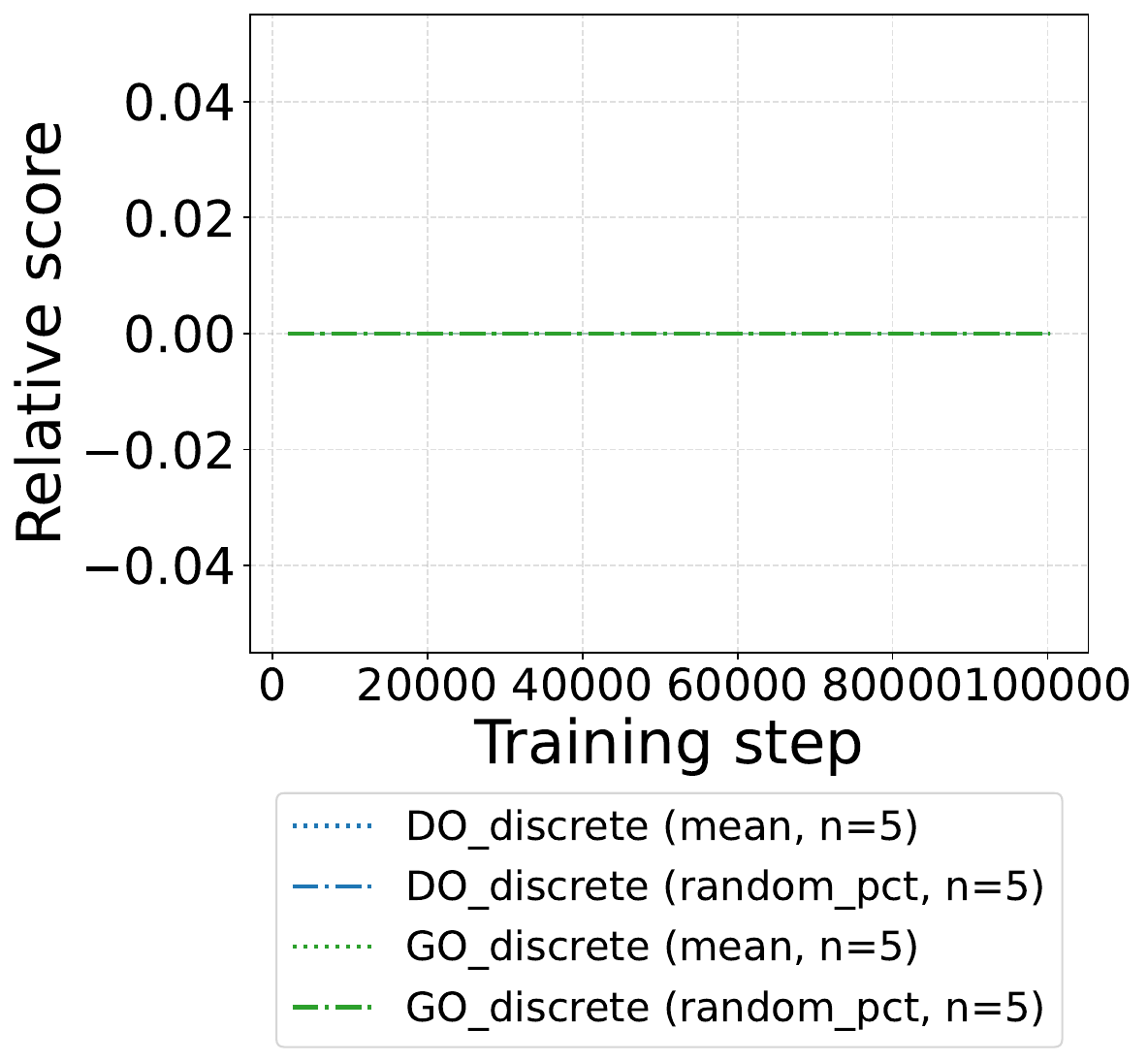}
        \caption{TSP Discrete M/V}
    \end{subfigure}
    \hfill
    \begin{subfigure}{0.32\textwidth}
        \centering
        \includegraphics[width=\linewidth]{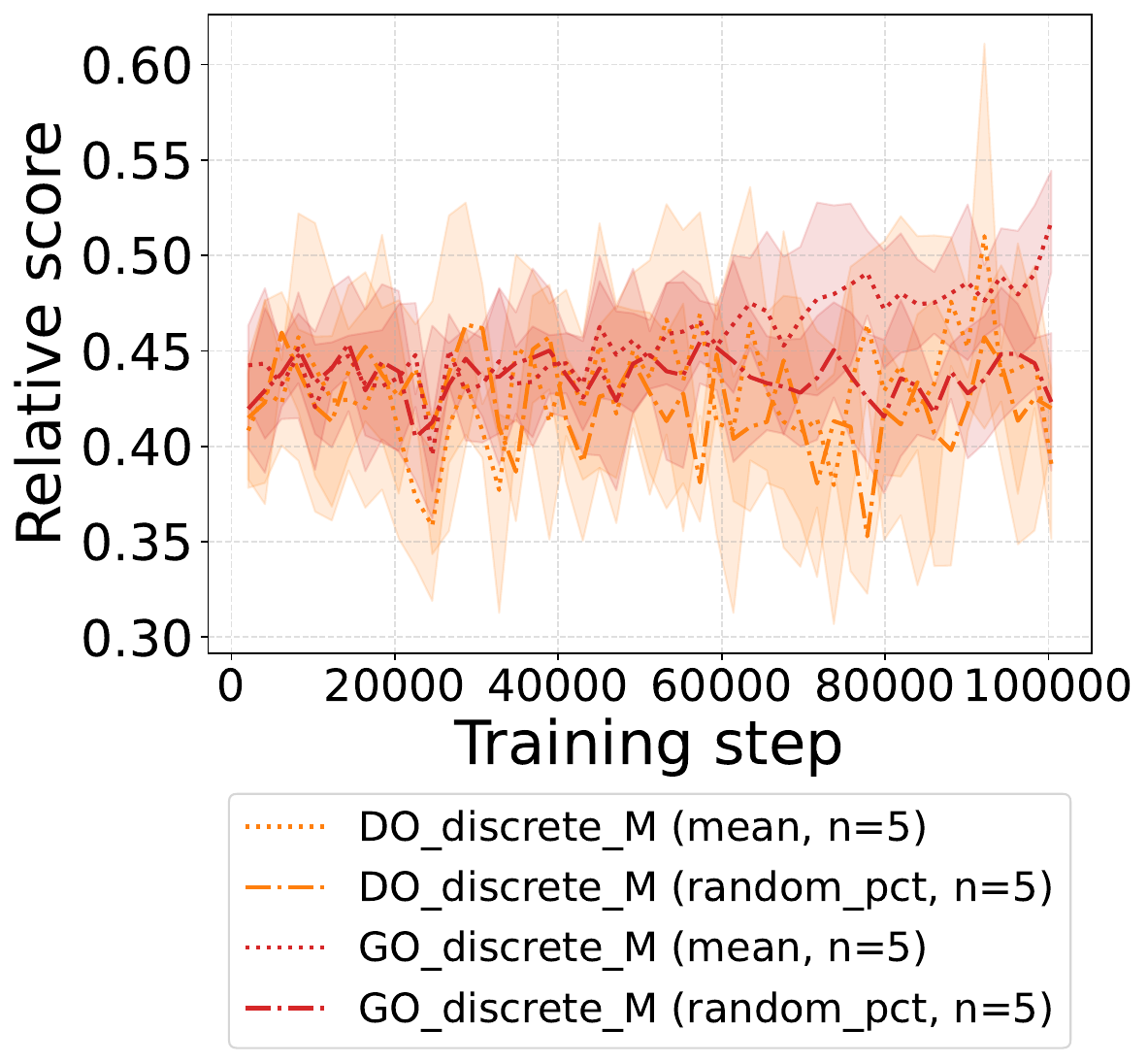}
        \caption{TSP Discrete-M M/V}
    \end{subfigure}
    \hfill
    \begin{subfigure}{0.32\textwidth}
        \centering
        \includegraphics[width=\linewidth]{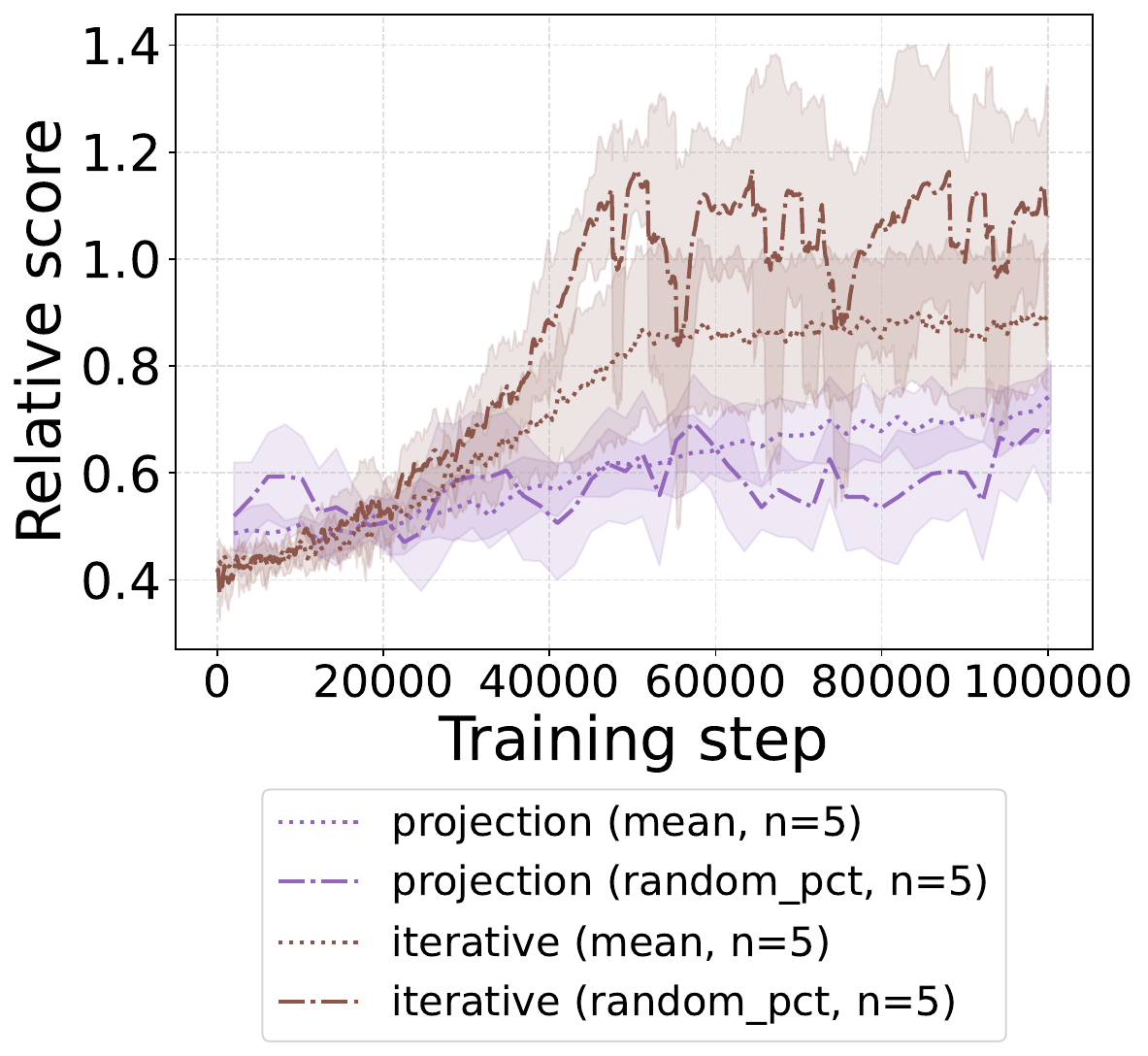}
        \caption{TSP Latent M/V}
    \end{subfigure}

  \caption{Average learning curves (mean ± standard deviation) across TSP benchmark environments under all training configurations.}
    \label{fig:all_learning_curves_travel}
\end{figure}

\begin{figure}[H]
    \centering

    \begin{subfigure}{0.32\textwidth}
        \centering
        \includegraphics[width=\linewidth]{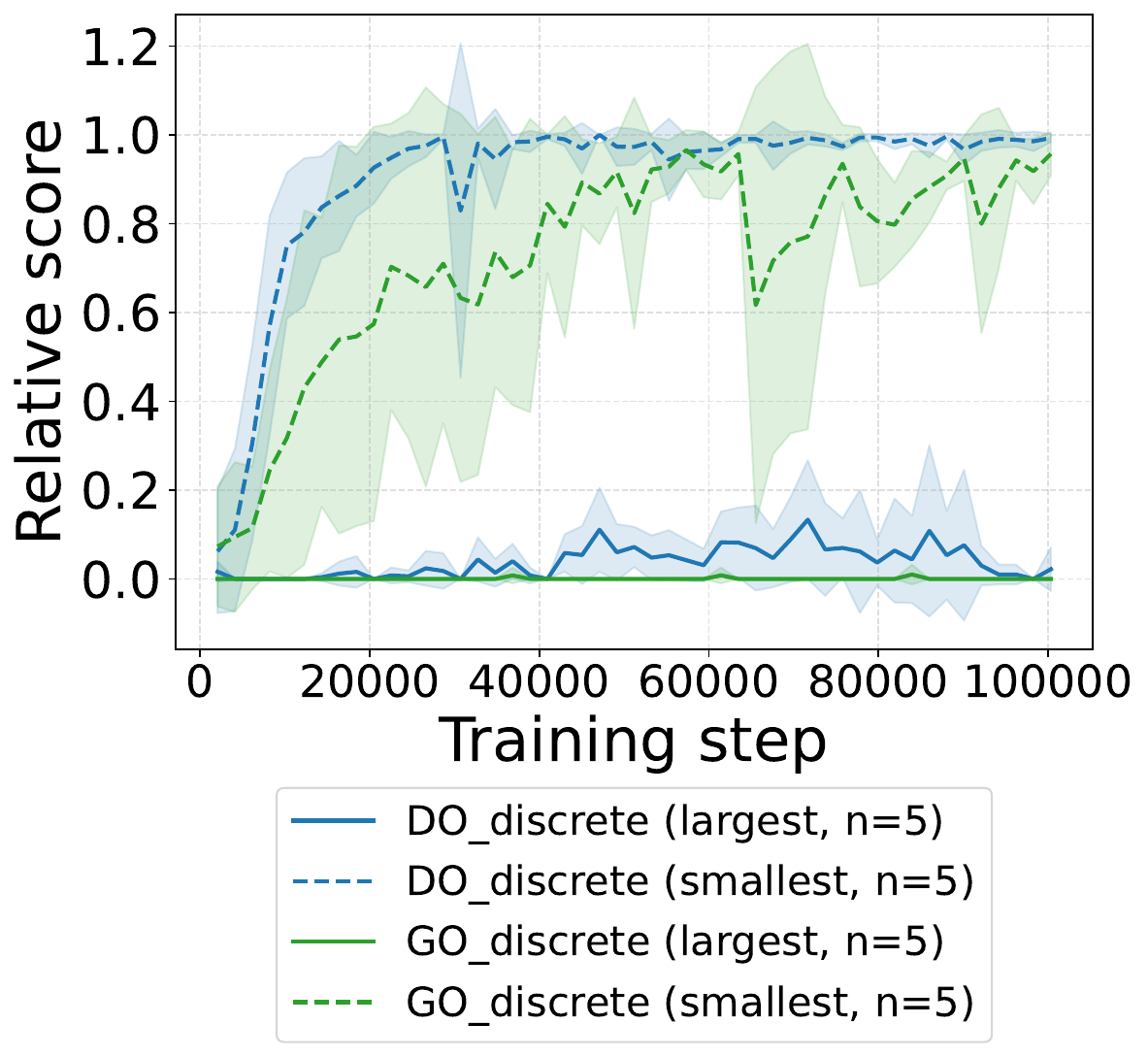}
        \caption{MinVertex Discrete S/L}
    \end{subfigure}
    \hfill
    \begin{subfigure}{0.32\textwidth}
        \centering
        \includegraphics[width=\linewidth]{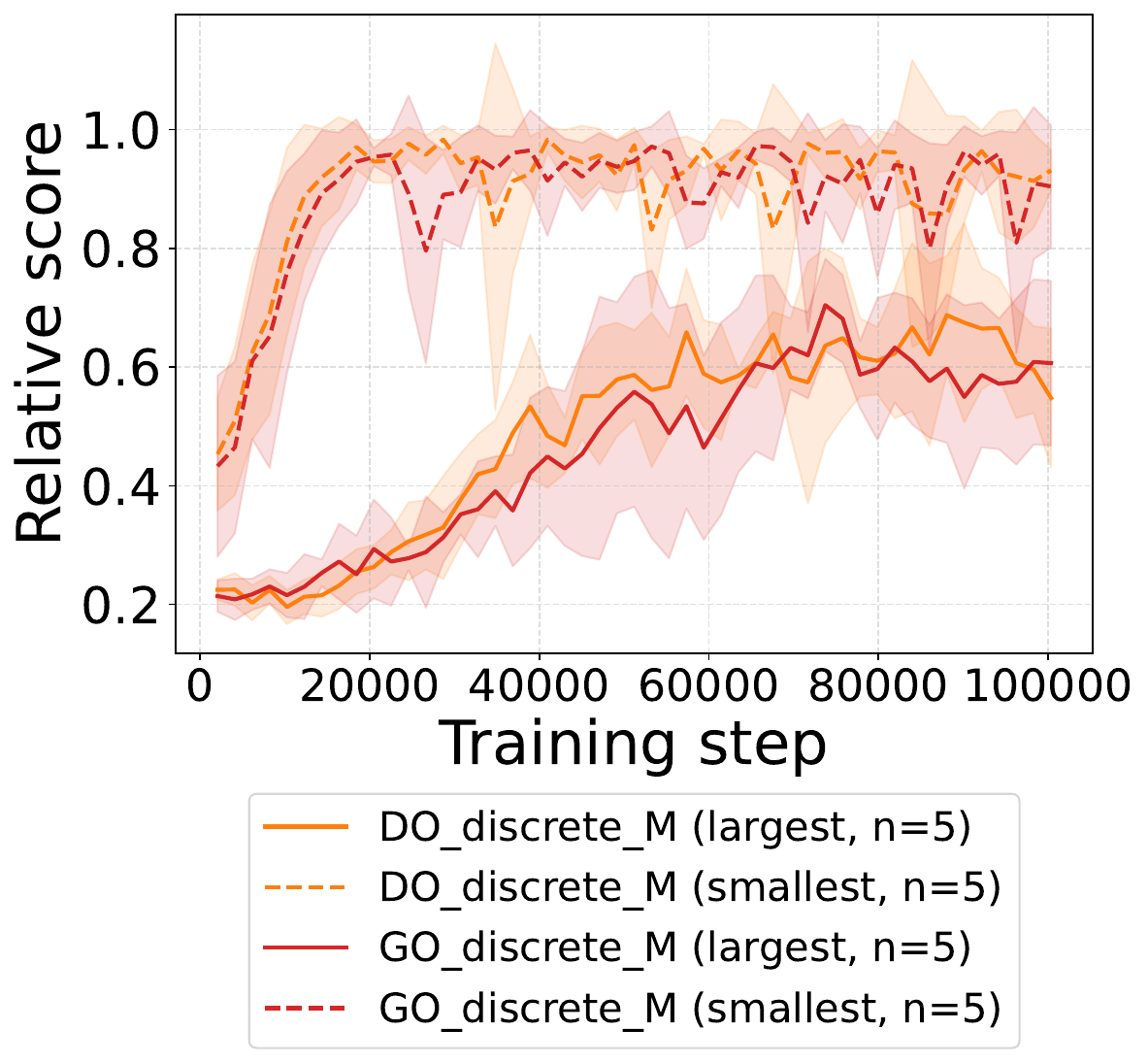}
        \caption{MinVertex Discrete-M S/L}
    \end{subfigure}
    \hfill
    \begin{subfigure}{0.32\textwidth}
        \centering
        \includegraphics[width=\linewidth]{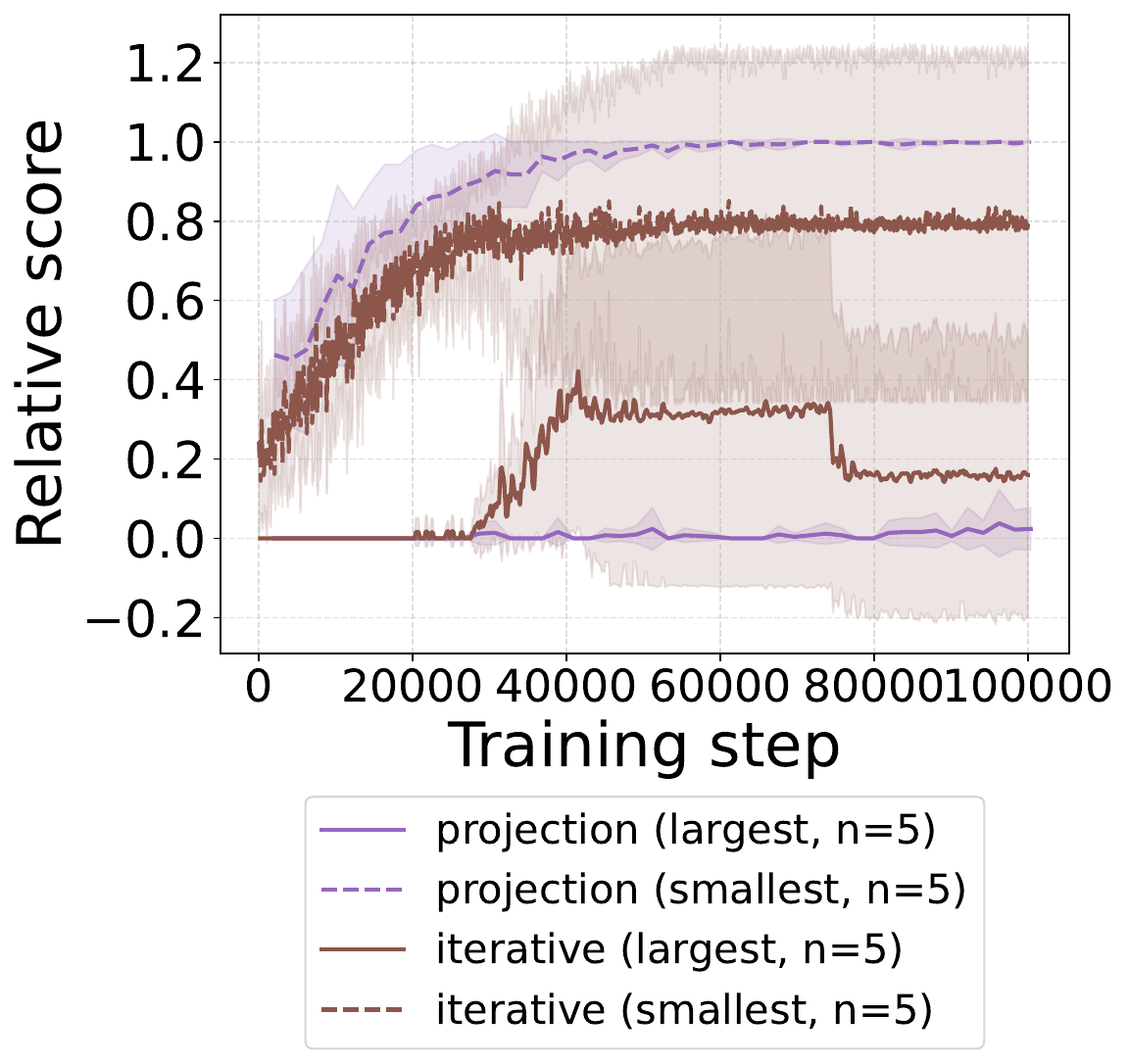}
        \caption{MinVertex Latent S/L}
    \end{subfigure}

    \begin{subfigure}{0.32\textwidth}
        \centering
        \includegraphics[width=\linewidth]{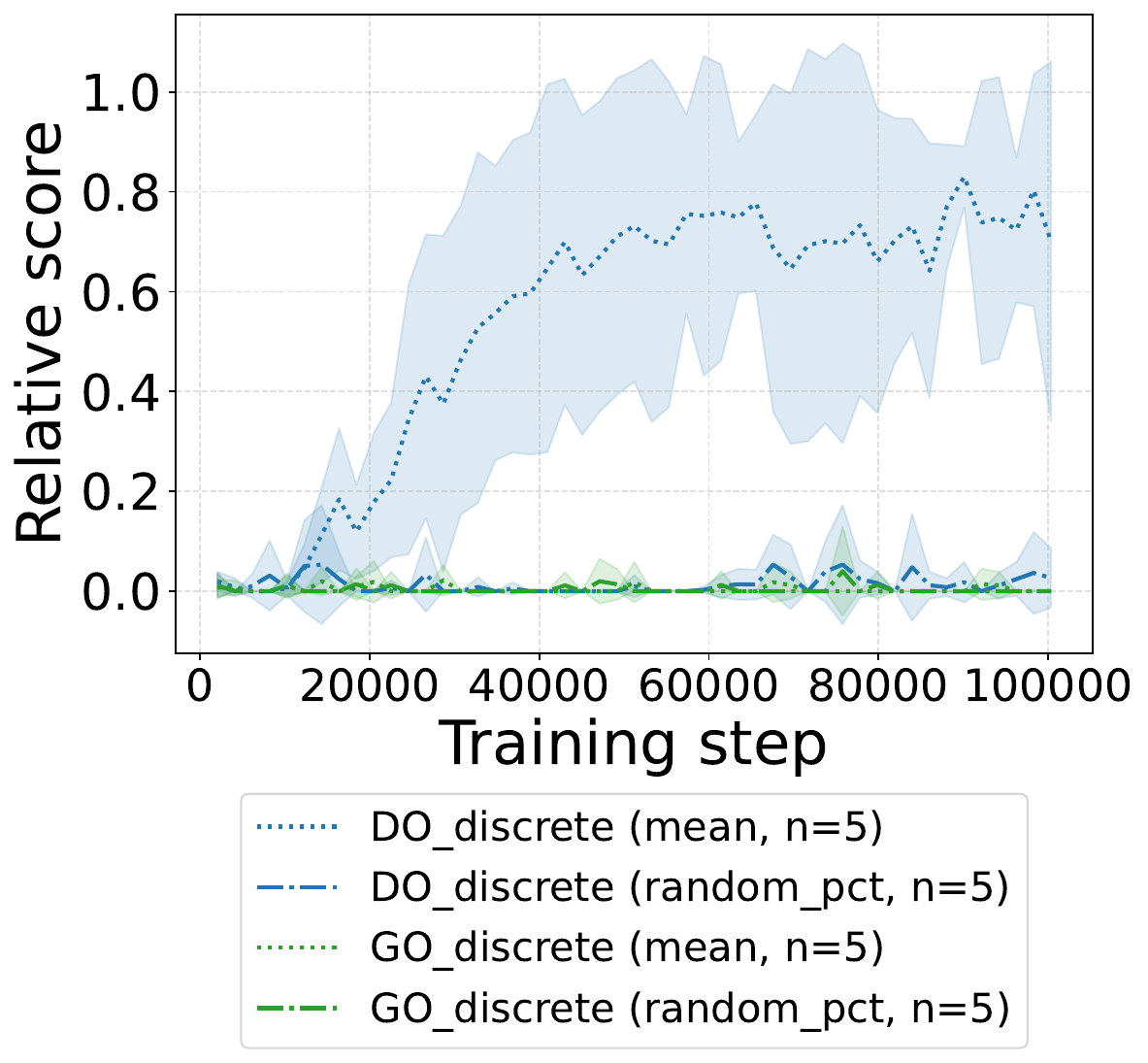}
        \caption{MinVertex Discrete M/V}
    \end{subfigure}
    \hfill
    \begin{subfigure}{0.32\textwidth}
        \centering
        \includegraphics[width=\linewidth]{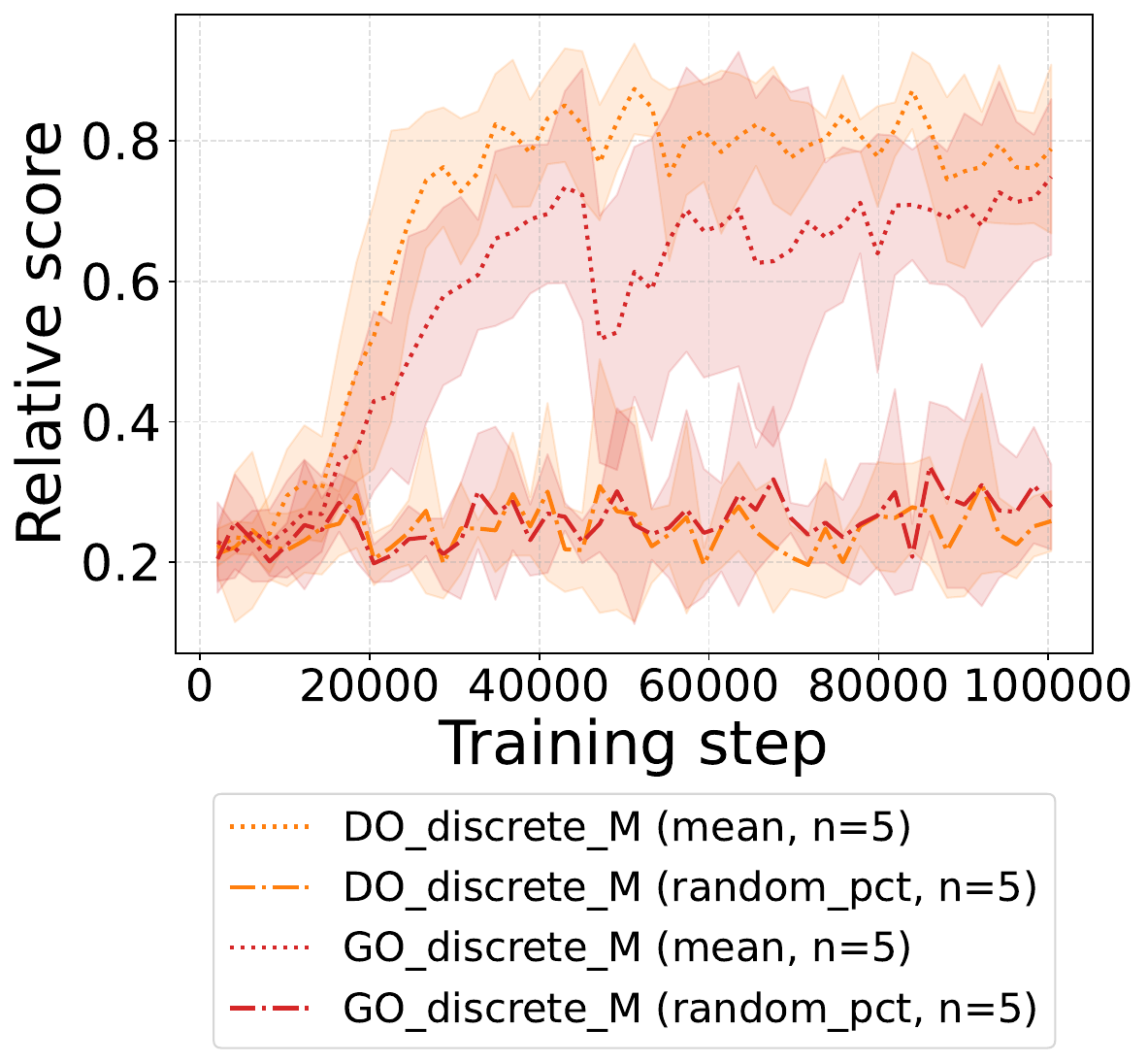}
        \caption{MinVertex Discrete-M M/V}
    \end{subfigure}
    \hfill
    \begin{subfigure}{0.32\textwidth}
        \centering
        \includegraphics[width=\linewidth]{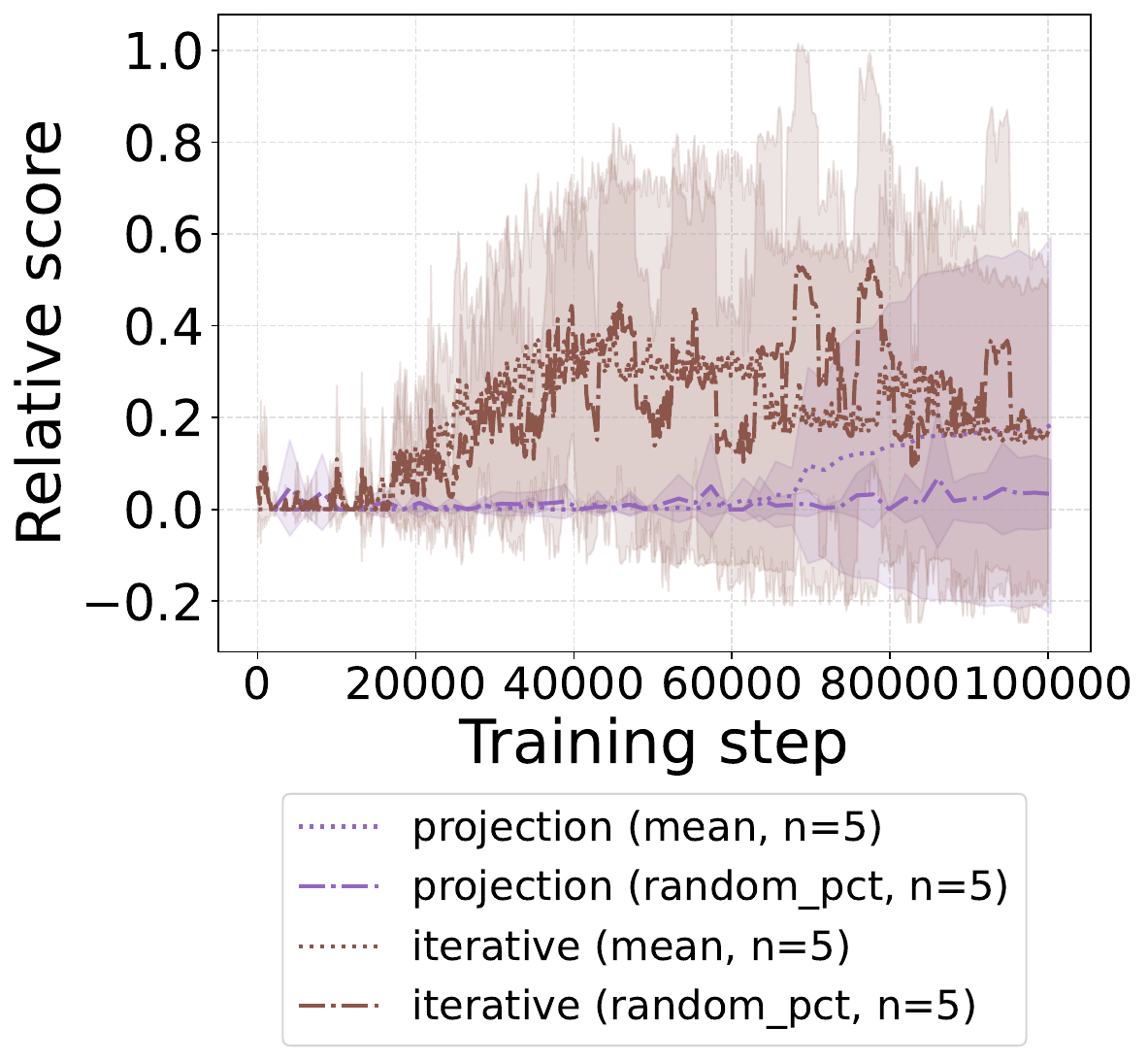}
        \caption{MinVertex Latent M/V}
    \end{subfigure}

  \caption{Average learning curves (mean ± standard deviation) across MinVertex benchmark environments under all training configurations.}
    \label{fig:all_learning_curves_minvertex}
\end{figure}

\begin{figure}[H]
    \centering

    \begin{subfigure}{0.32\textwidth}
        \centering
        \includegraphics[width=\linewidth]{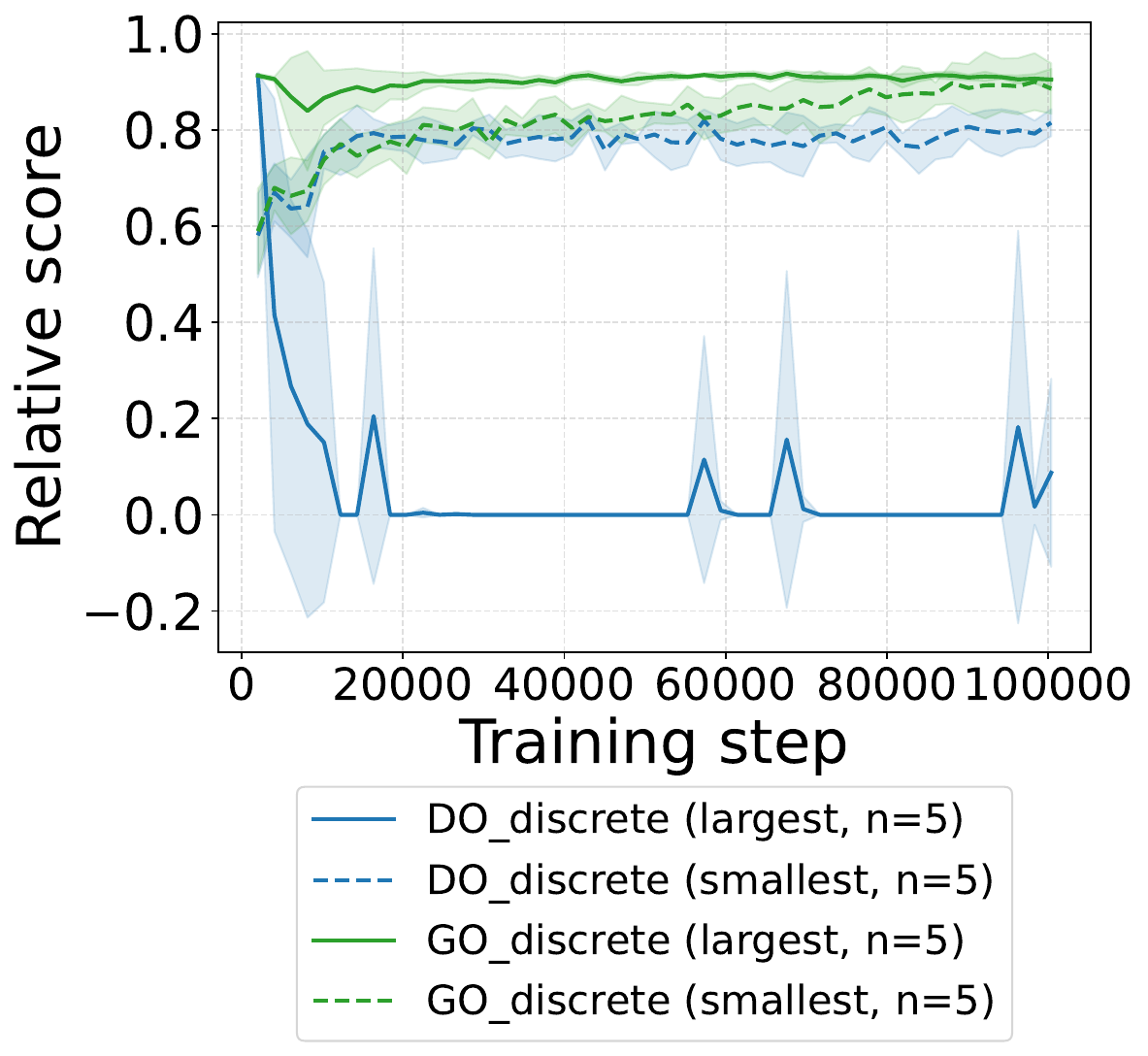}
        \caption{MaxCut Discrete S/L}
    \end{subfigure}
    \hfill
    \begin{subfigure}{0.32\textwidth}
        \centering
        \includegraphics[width=\linewidth]{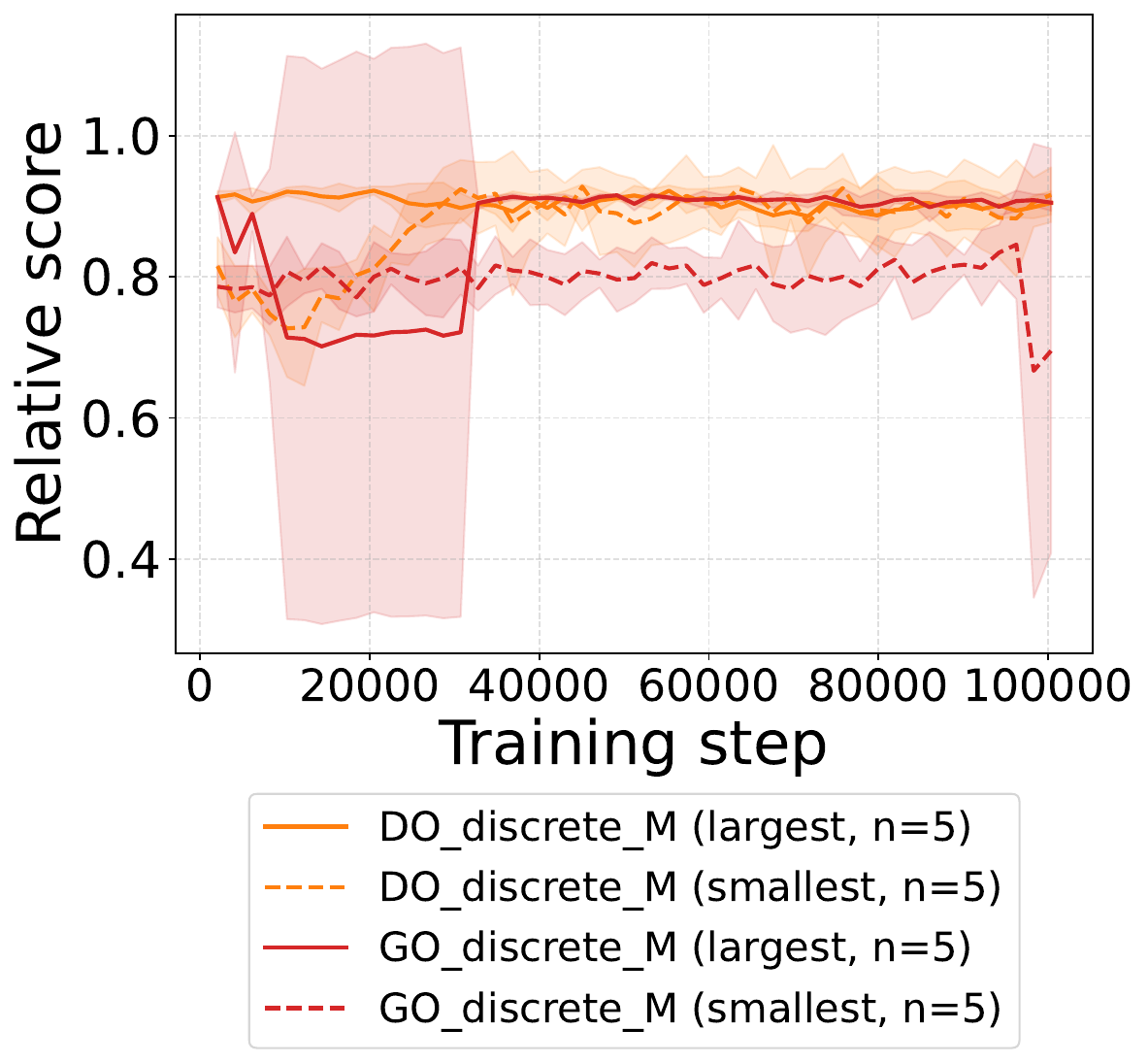}
        \caption{MaxCut Discrete-M S/L}
    \end{subfigure}
    \hfill
    \begin{subfigure}{0.32\textwidth}
        \centering
        \includegraphics[width=\linewidth]{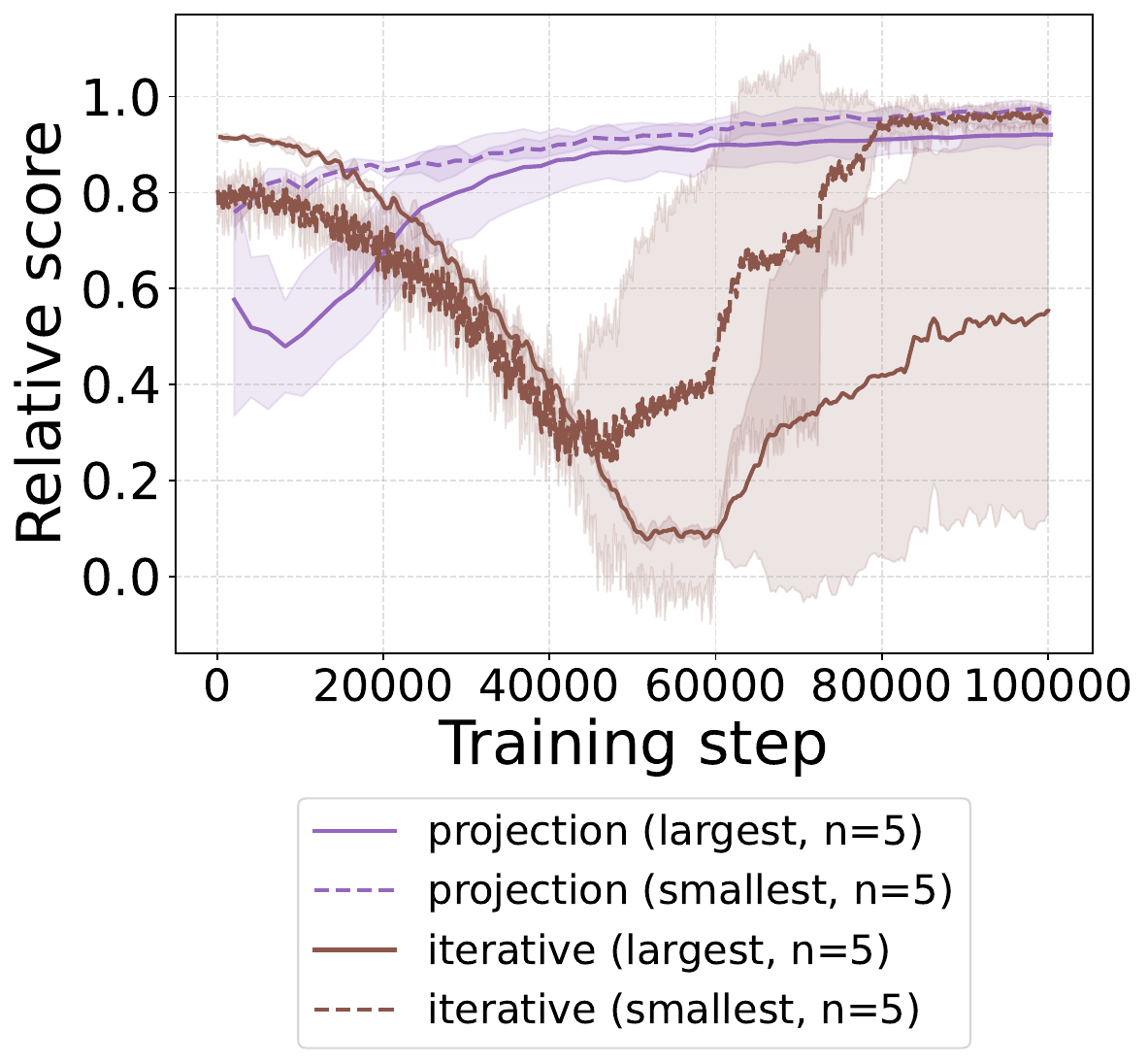}
        \caption{MaxCut Latent S/L}
    \end{subfigure}

    \begin{subfigure}{0.32\textwidth}
        \centering
        \includegraphics[width=\linewidth]{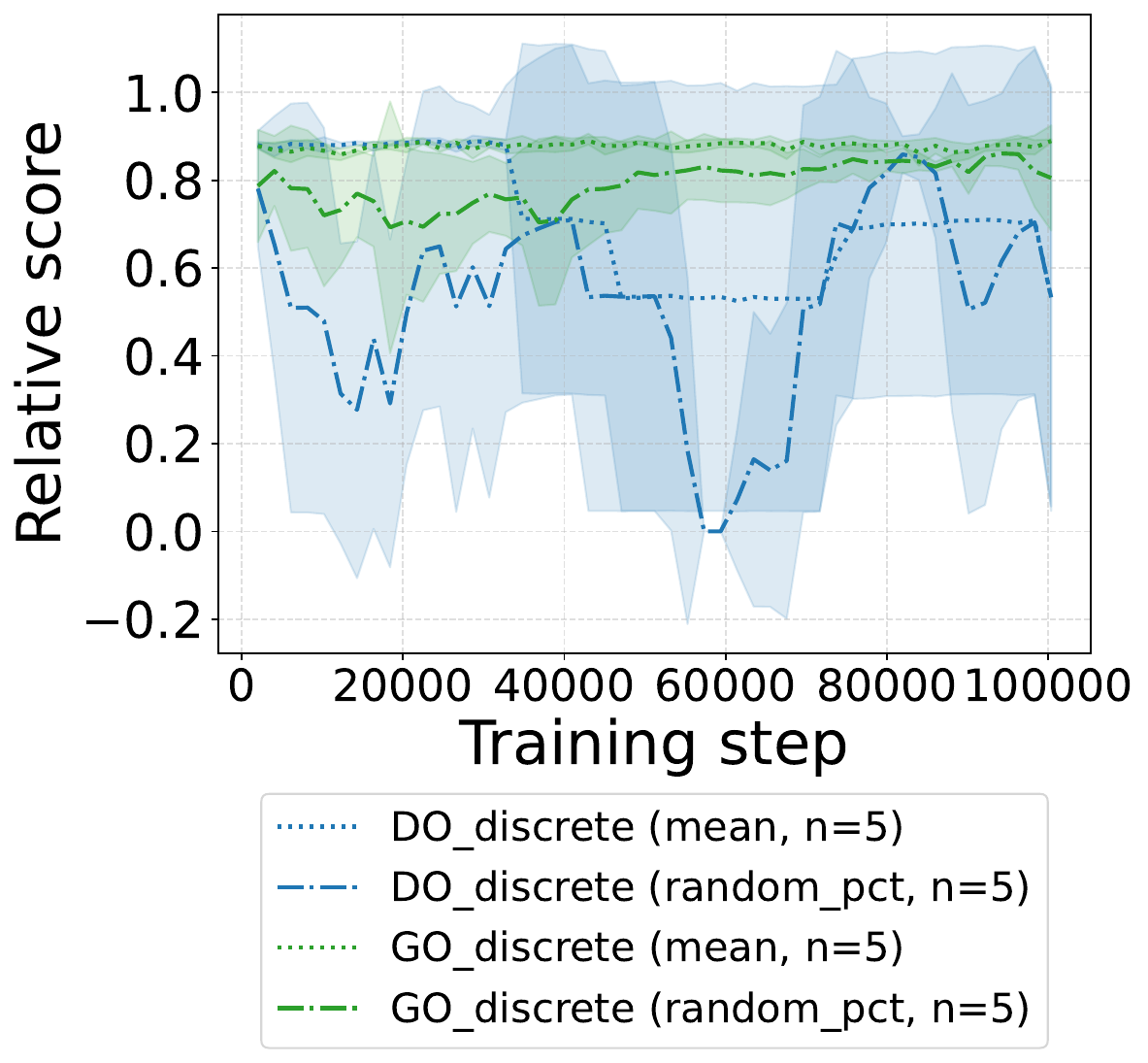}
        \caption{MaxCut Discrete M/V}
    \end{subfigure}
    \hfill
    \begin{subfigure}{0.32\textwidth}
        \centering
        \includegraphics[width=\linewidth]{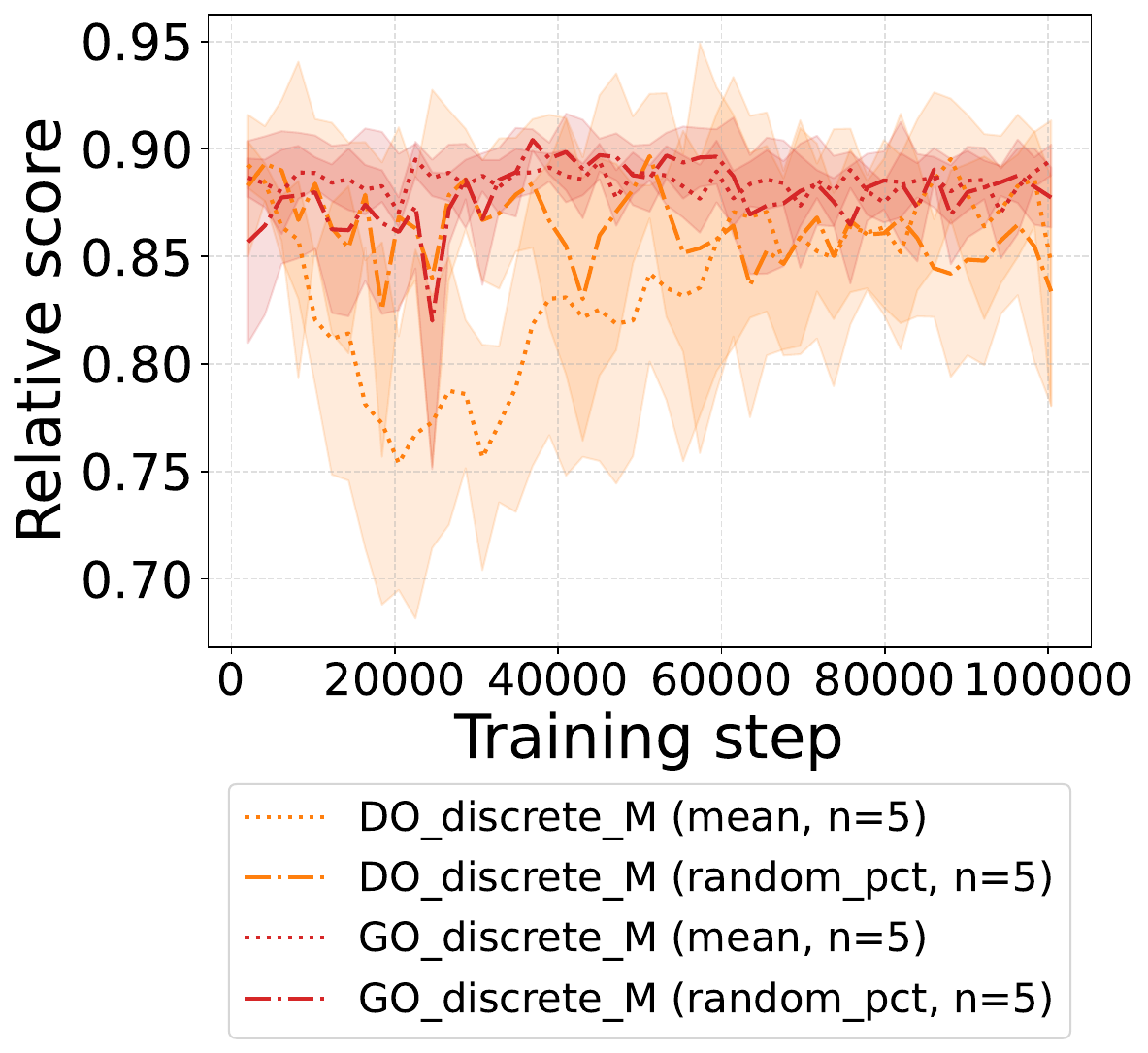}
        \caption{MaxCut Discrete-M M/V}
    \end{subfigure}
    \hfill
    \begin{subfigure}{0.32\textwidth}
        \centering
        \includegraphics[width=\linewidth]{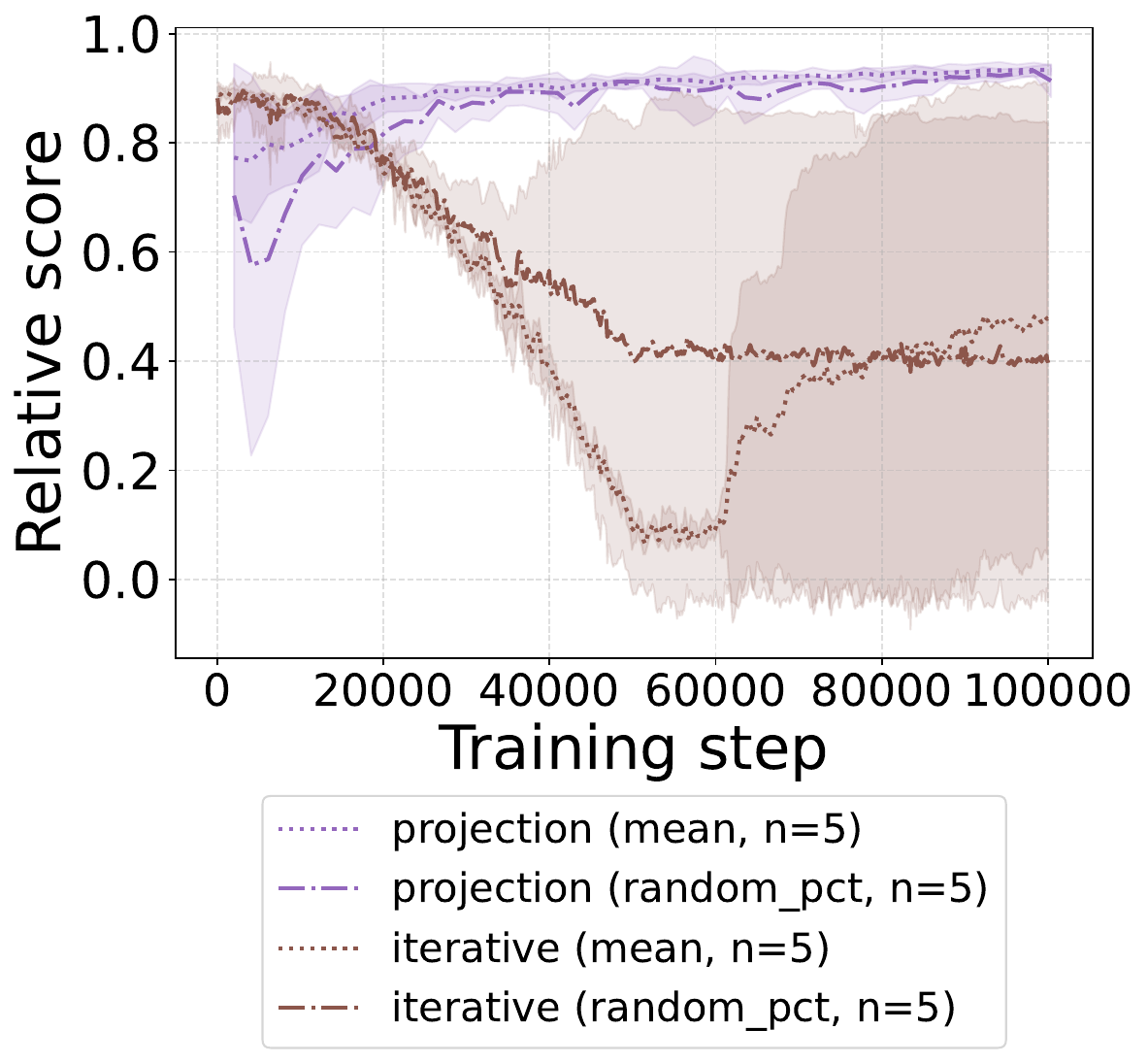}
        \caption{MaxCut Latent M/V}
    \end{subfigure}

  \caption{Average learning curves (mean ± standard deviation) across MaxCut benchmark environments under all training configurations.}
    \label{fig:all_learning_curves_MaxCut}
\end{figure}

\begin{figure}[H]
    \centering

    \begin{subfigure}{0.32\textwidth}
        \centering
        \includegraphics[width=\linewidth]{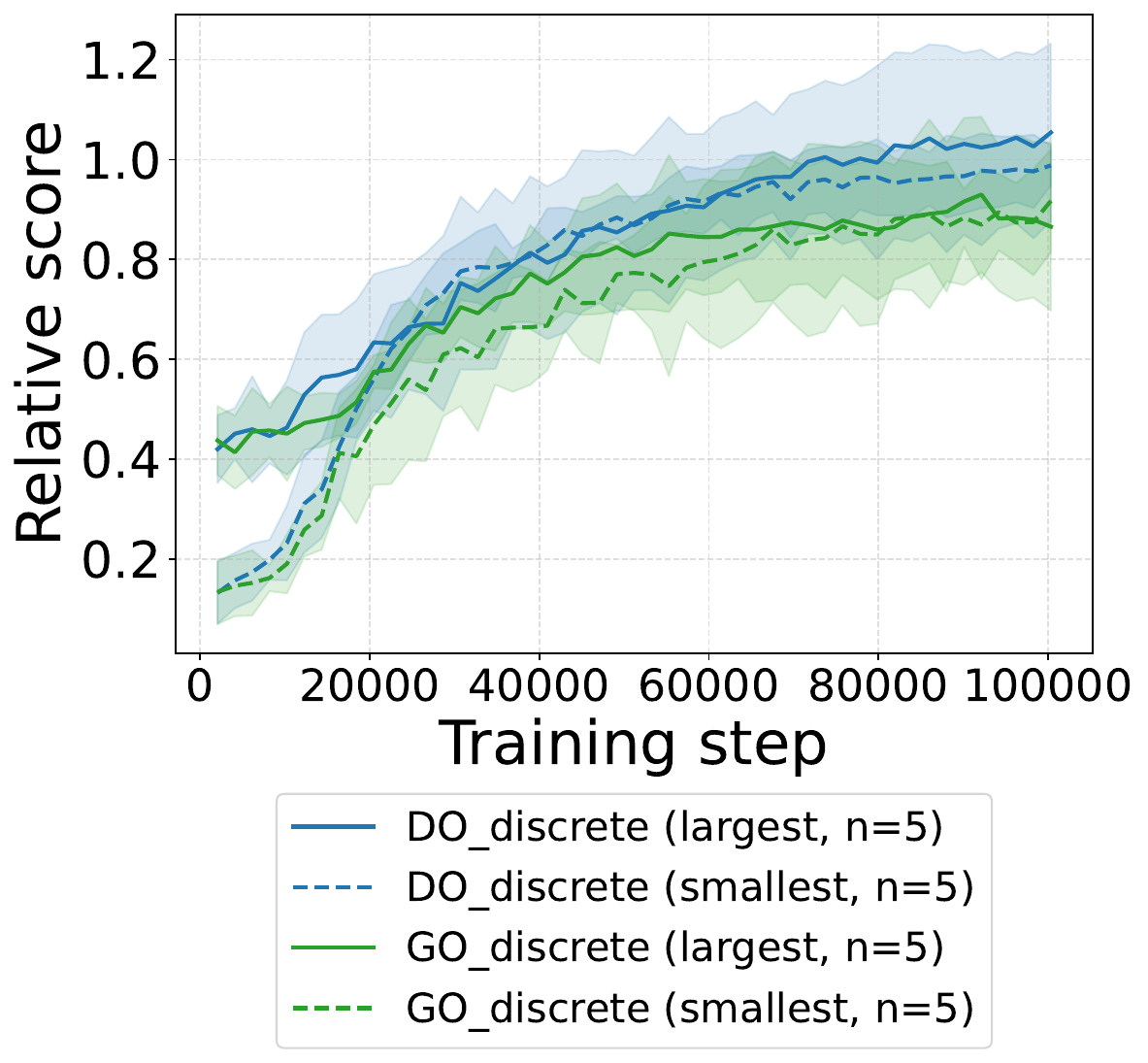}
        \caption{Placement Discrete S/L}
    \end{subfigure}
    \hfill
    \begin{subfigure}{0.32\textwidth}
        \centering
        \includegraphics[width=\linewidth]{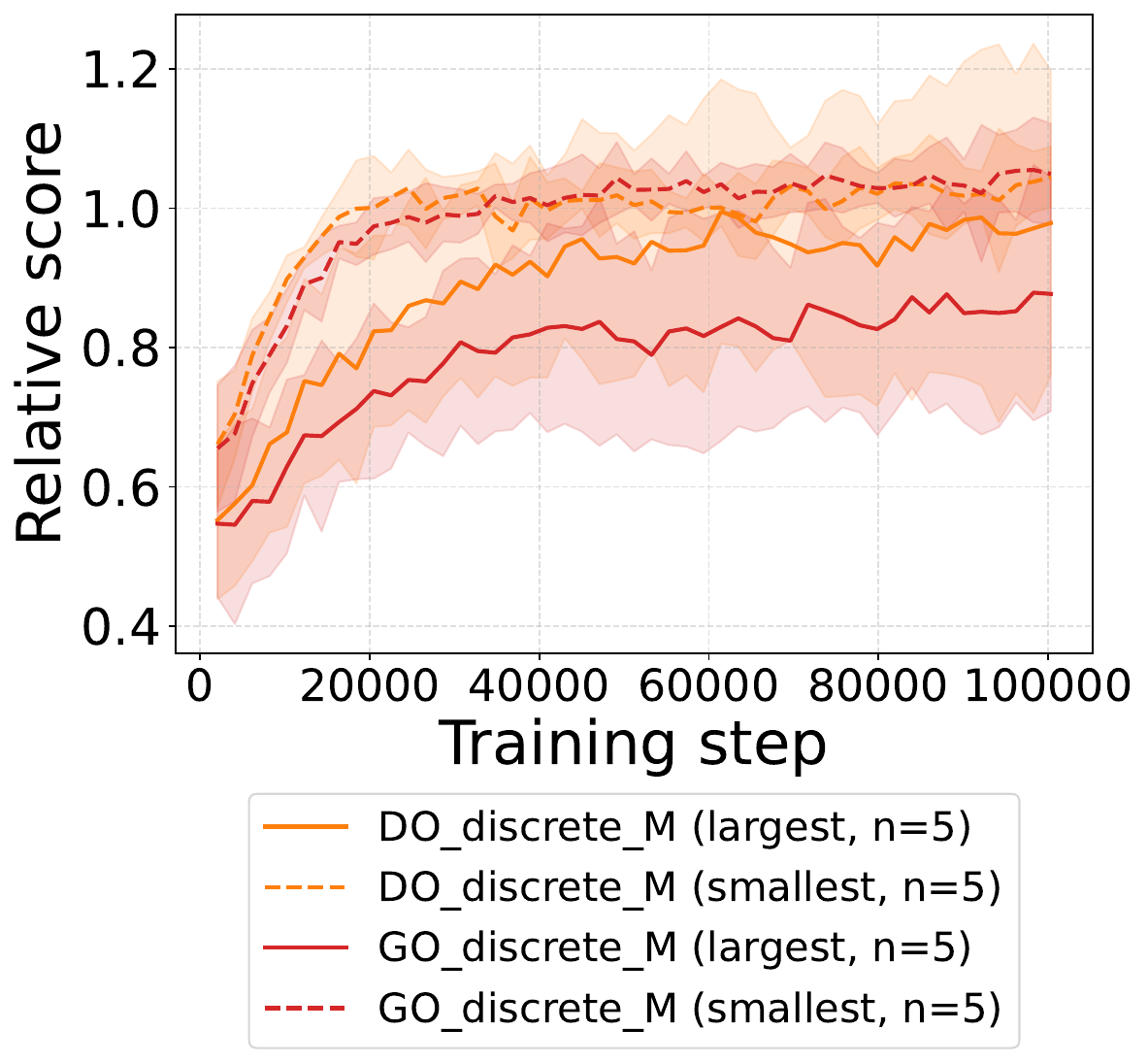}
        \caption{Placement Discrete-M S/L}
    \end{subfigure}
    \hfill
    \begin{subfigure}{0.32\textwidth}
        \centering
        \includegraphics[width=\linewidth]{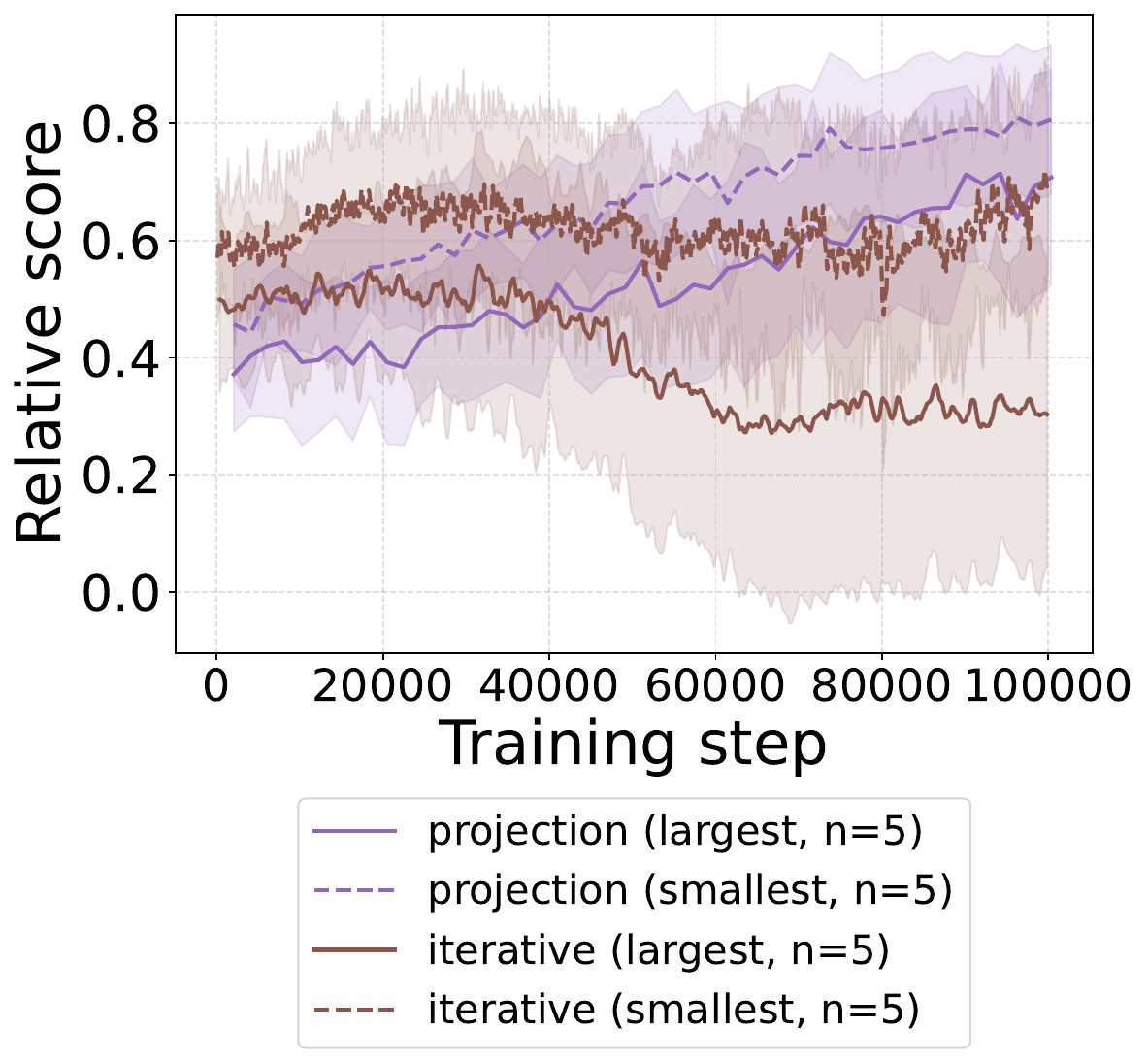}
        \caption{Placement Latent S/L}
    \end{subfigure}

    \begin{subfigure}{0.32\textwidth}
        \centering
        \includegraphics[width=\linewidth]{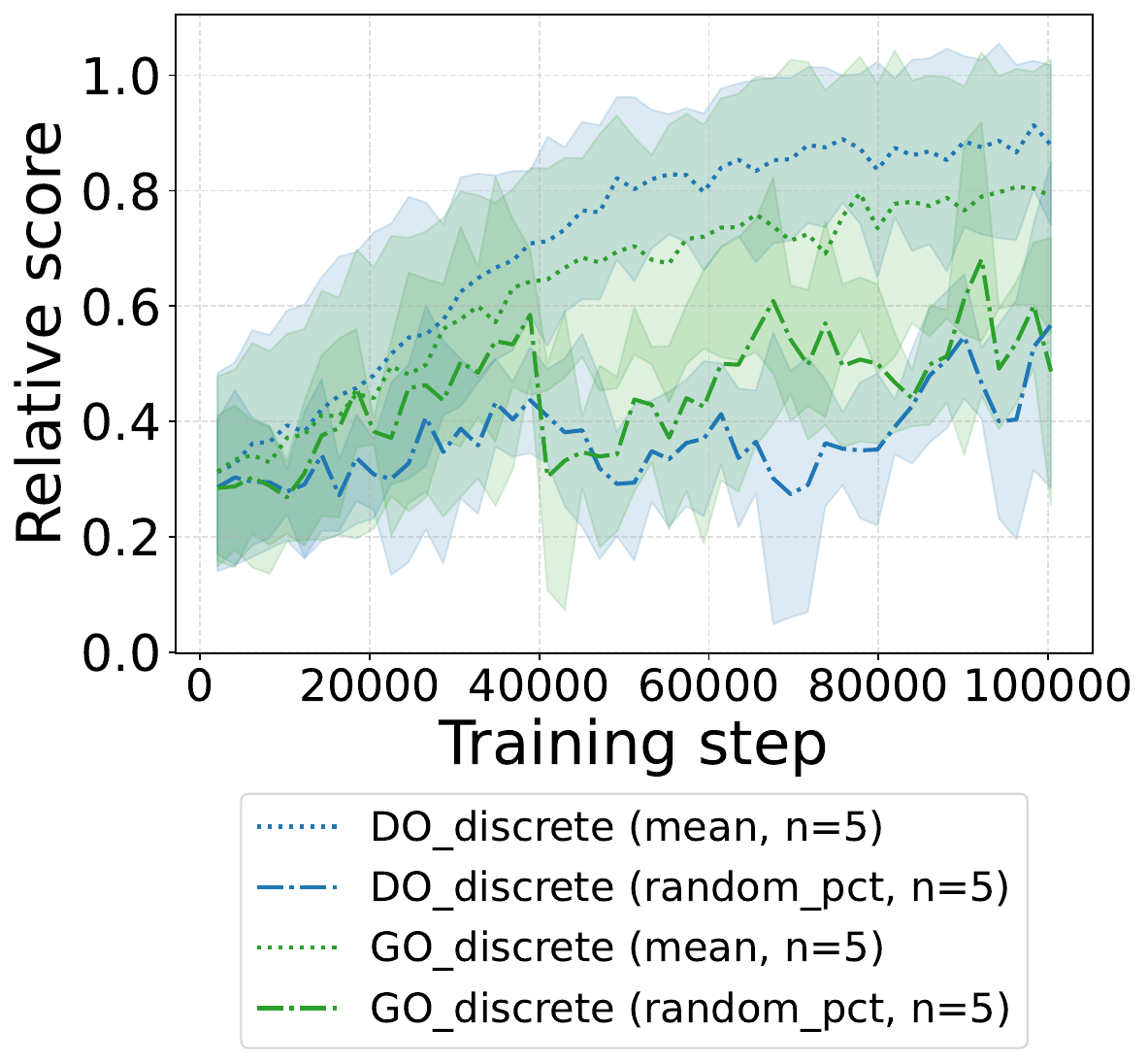}
        \caption{Placement Discrete M/V}
    \end{subfigure}
    \hfill
    \begin{subfigure}{0.32\textwidth}
        \centering
        \includegraphics[width=\linewidth]{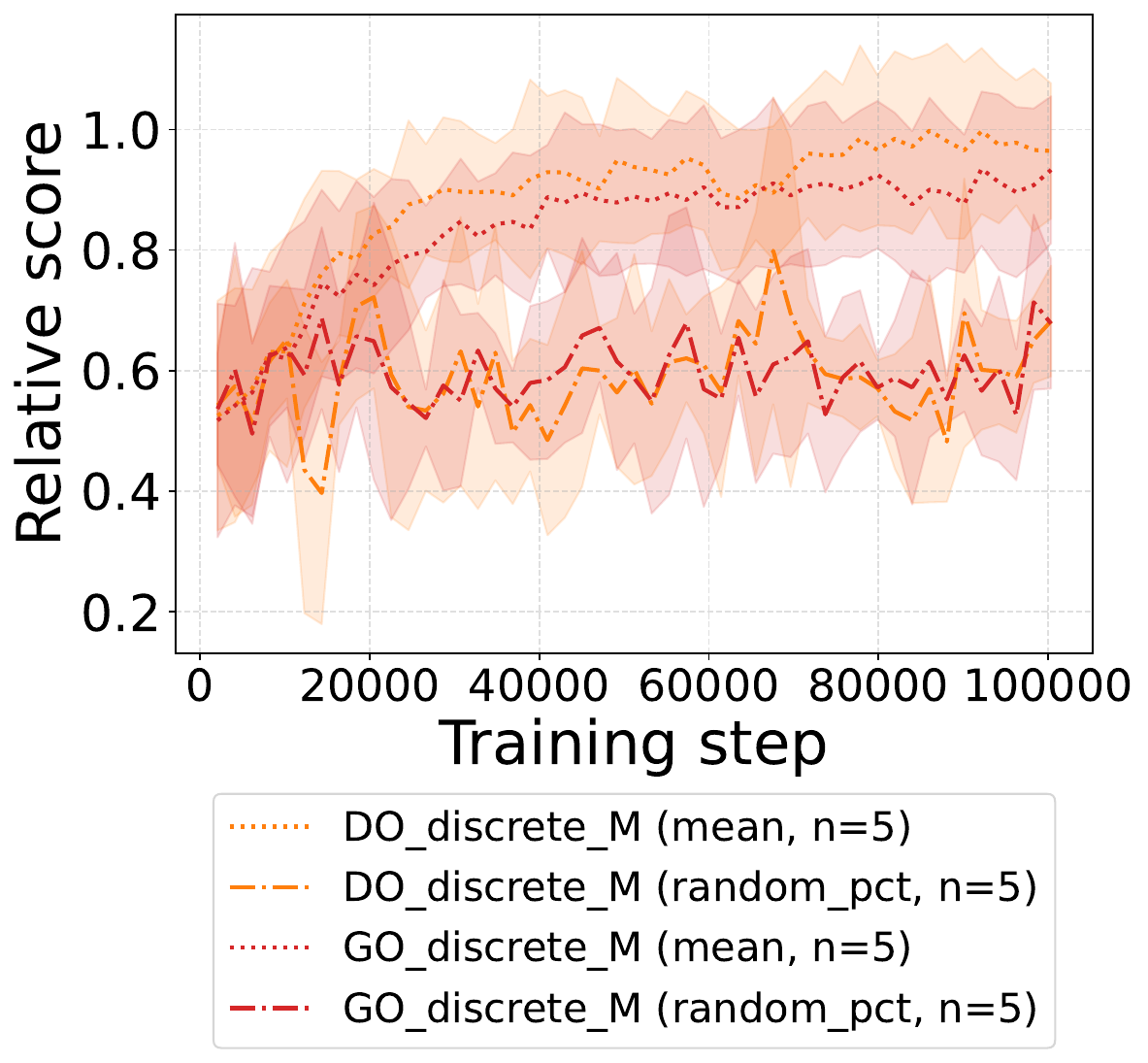}
        \caption{Placement Discrete-M M/V}
    \end{subfigure}
    \hfill
    \begin{subfigure}{0.32\textwidth}
        \centering
        \includegraphics[width=\linewidth]{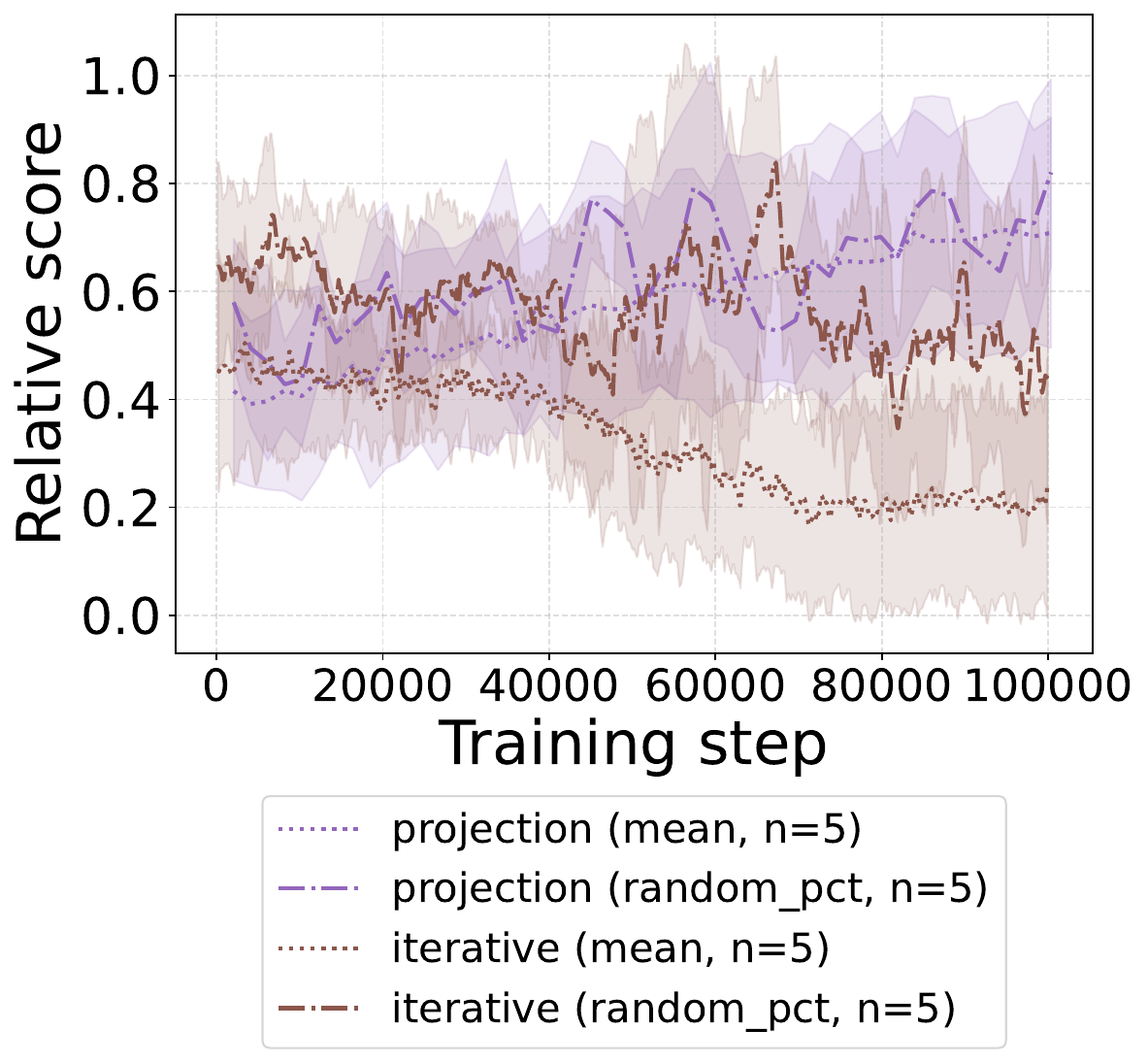}
        \caption{Placement Latent M/V}
    \end{subfigure}

  \caption{Average learning curves (mean ± standard deviation) across Placement benchmark environments under all training configurations.}
    \label{fig:all_learning_curves_Placement}
\end{figure}

\begin{figure}[H]
    \centering

    \begin{subfigure}{0.32\textwidth}
        \centering
        \includegraphics[width=\linewidth]{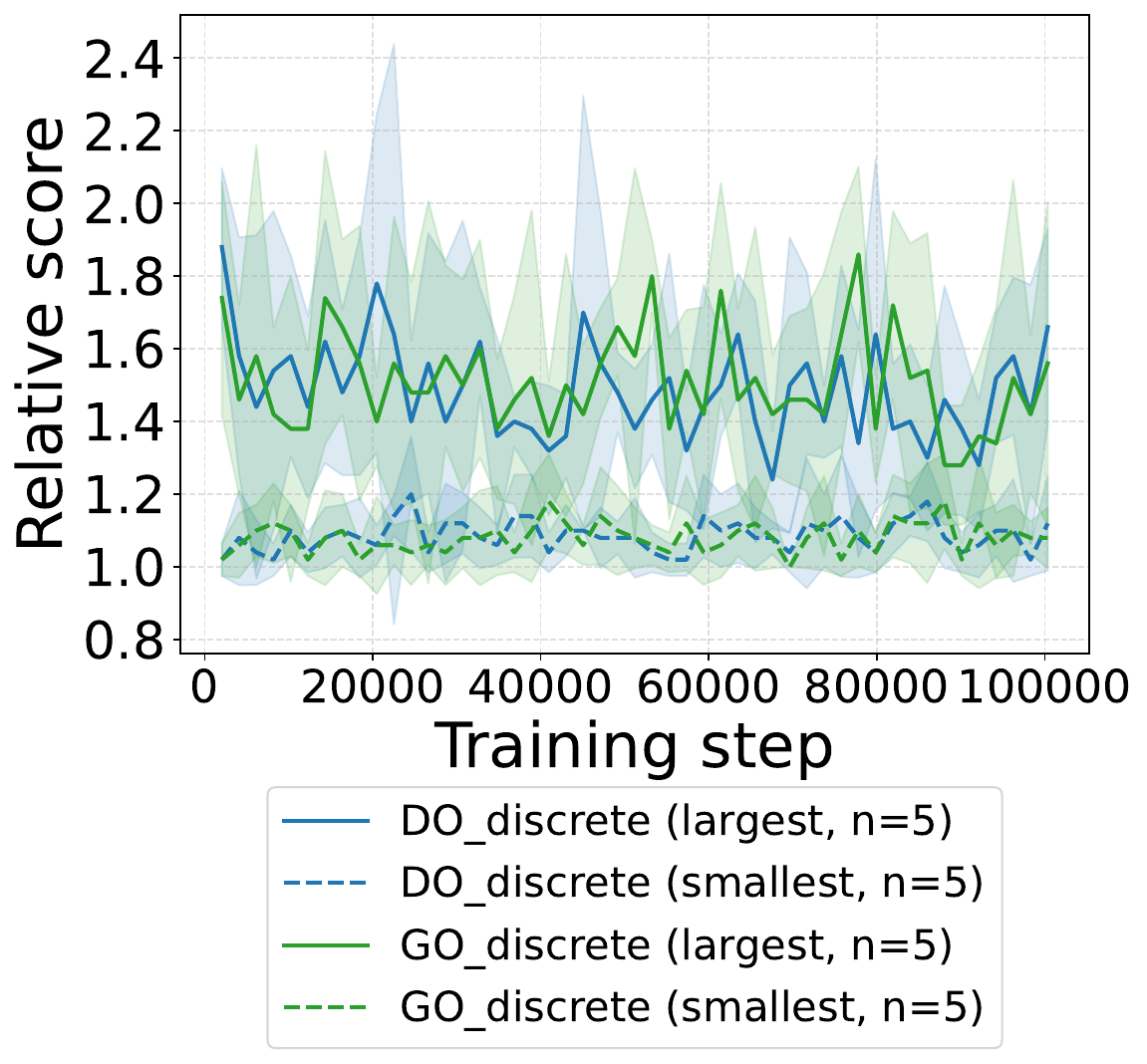}
        \caption{Cyber-Path Discrete S/L}
    \end{subfigure}
    \hfill
    \begin{subfigure}{0.32\textwidth}
        \centering
        \includegraphics[width=\linewidth]{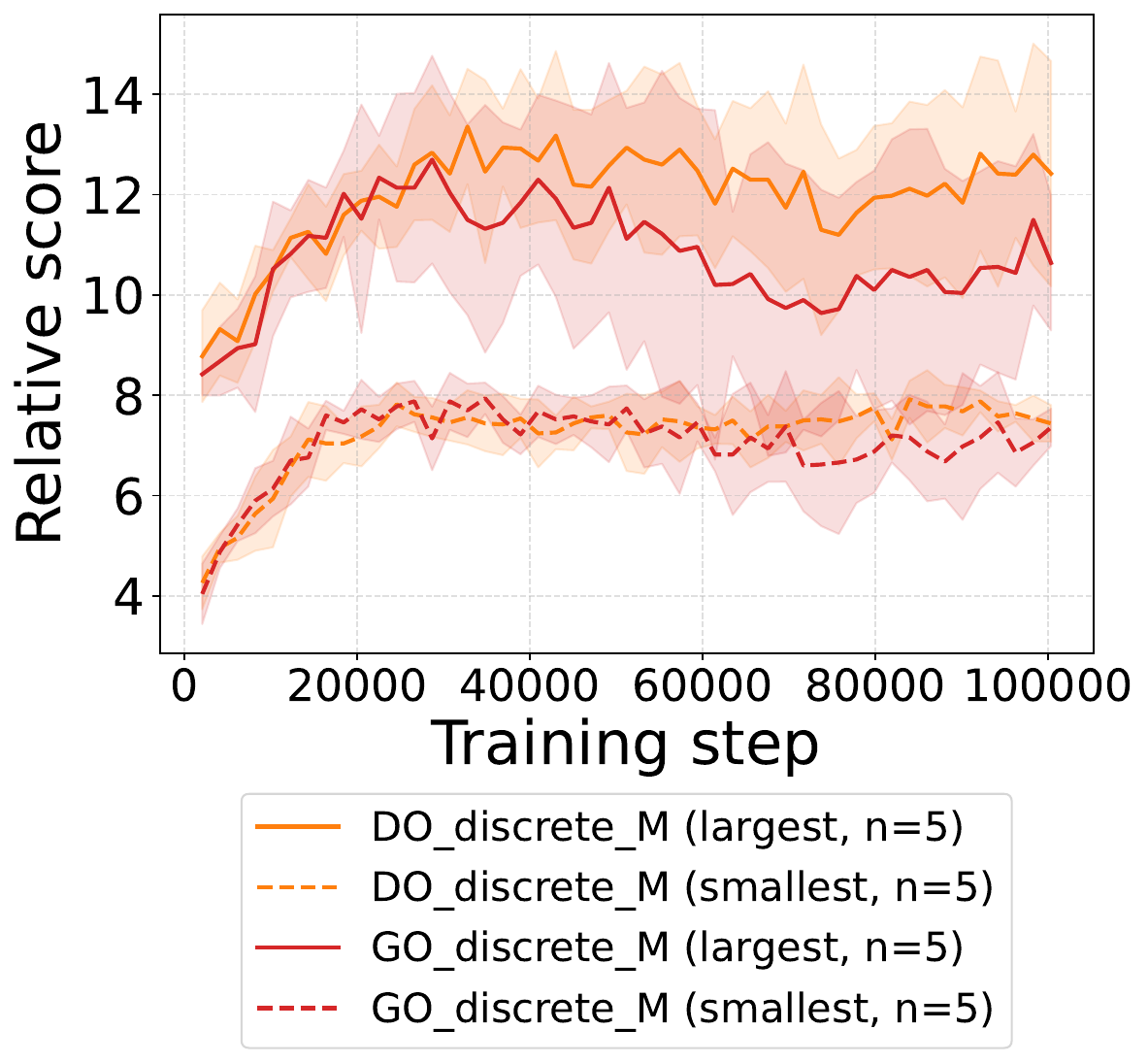}
        \caption{Cyber-Path Discrete-M S/L}
    \end{subfigure}
    \hfill
    \begin{subfigure}{0.32\textwidth}
        \centering
        \includegraphics[width=\linewidth]{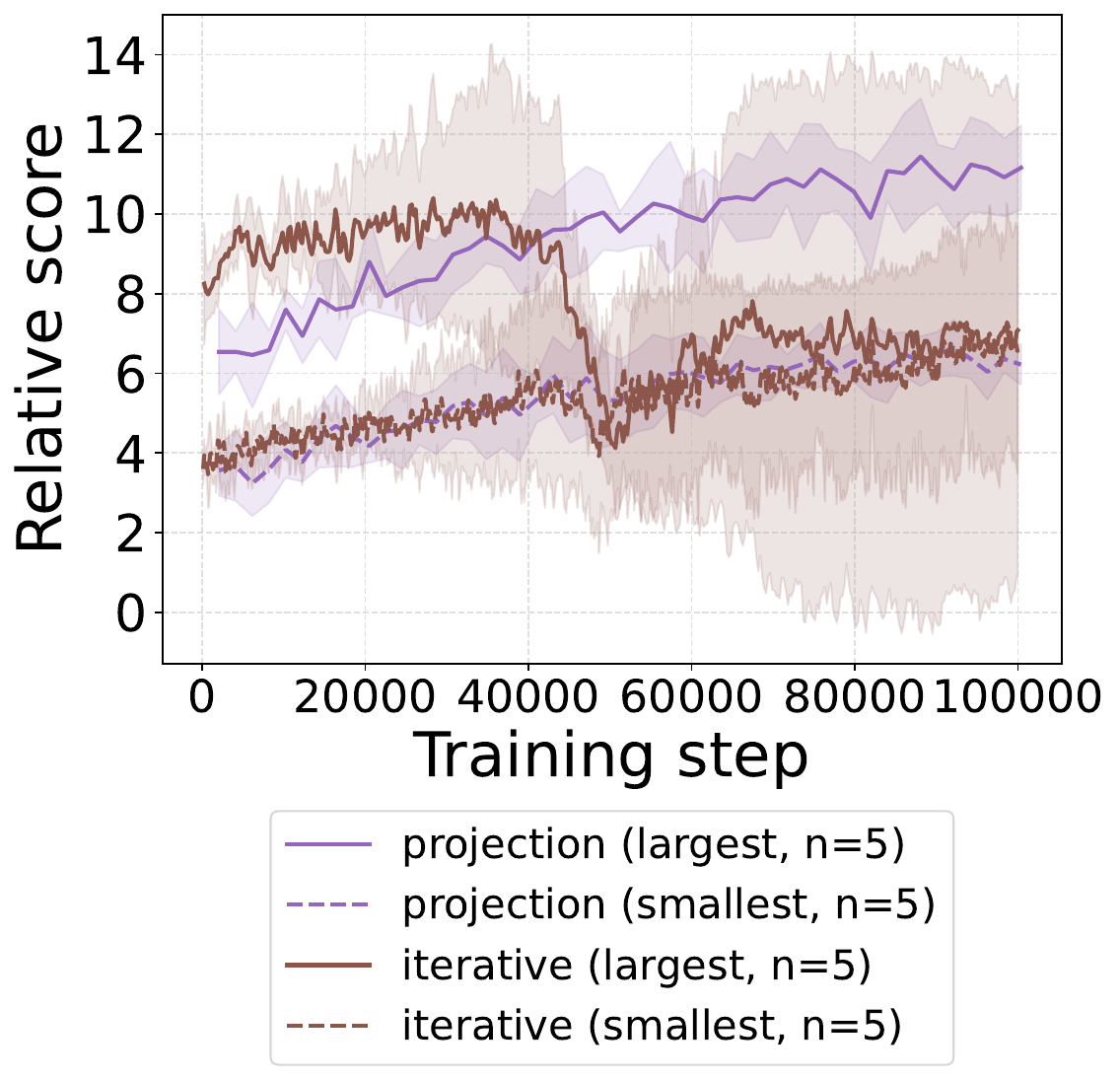}
        \caption{Cyber-Path Latent S/L}
    \end{subfigure}

    \begin{subfigure}{0.32\textwidth}
        \centering
        \includegraphics[width=\linewidth]{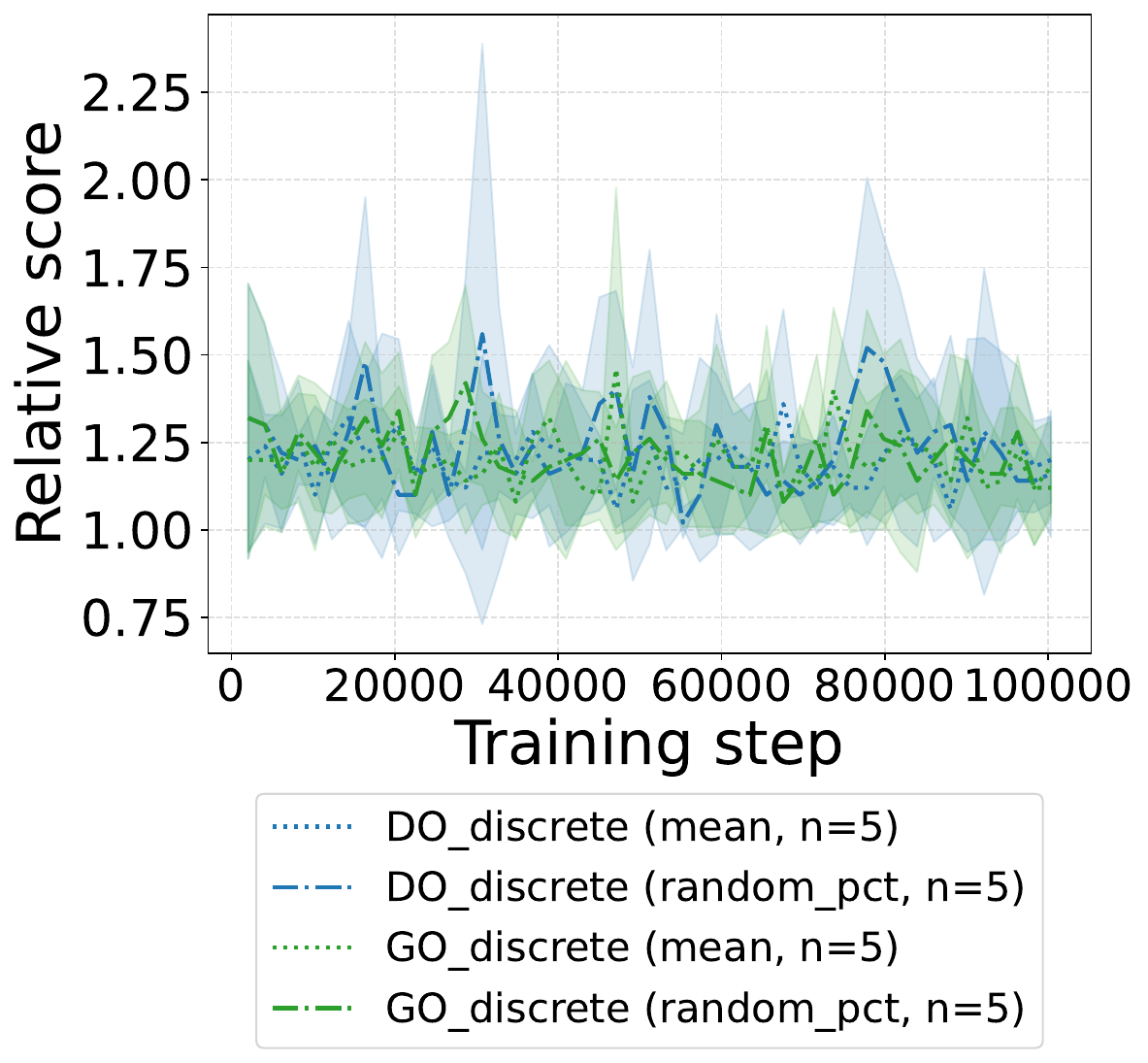}
        \caption{Cyber-Path Discrete M/V}
    \end{subfigure}
    \hfill
    \begin{subfigure}{0.32\textwidth}
        \centering
        \includegraphics[width=\linewidth]{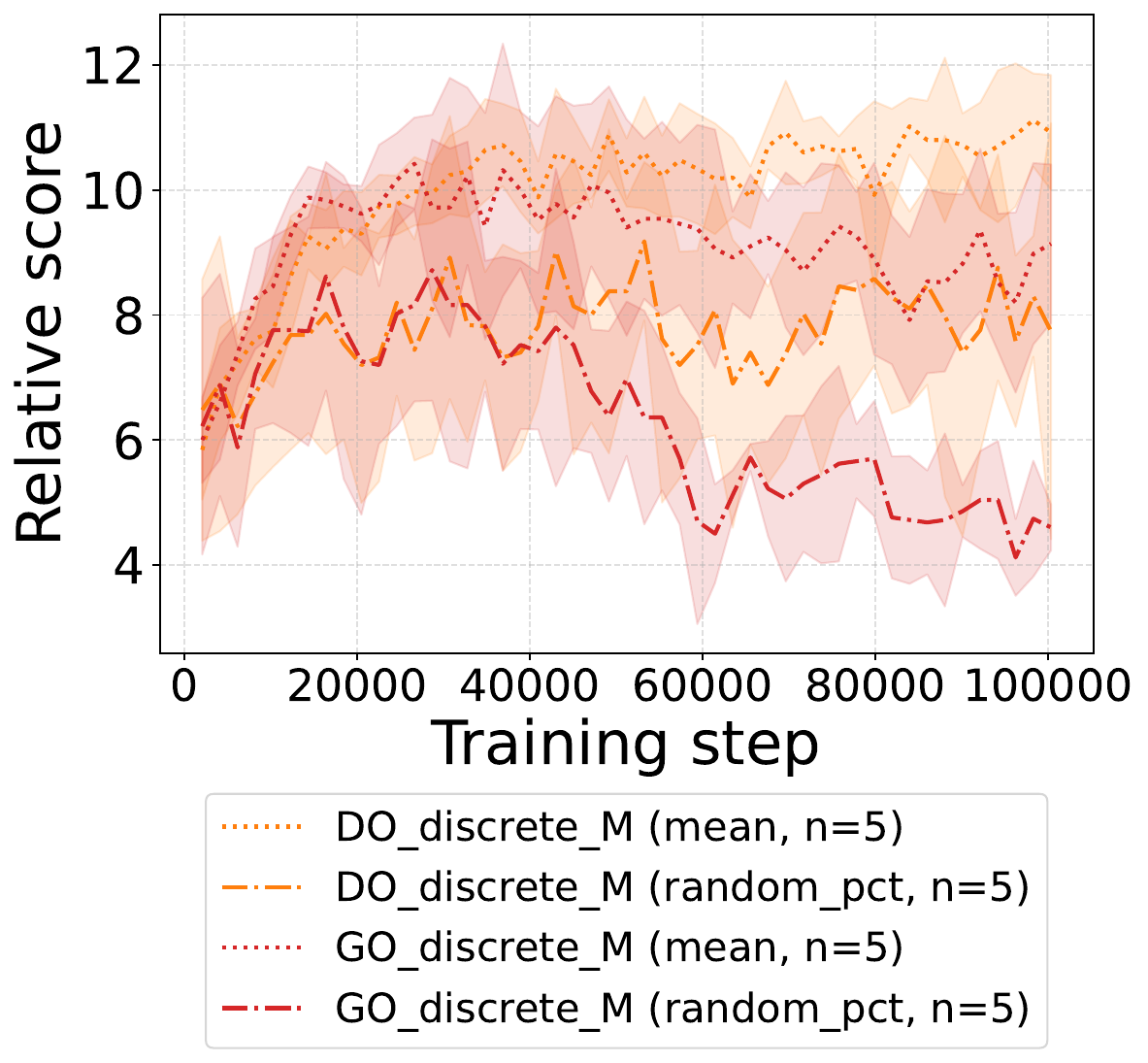}        \caption{Cyber-Path Discrete-M M/V}
    \end{subfigure}
    \hfill
    \begin{subfigure}{0.32\textwidth}
        \centering
        \includegraphics[width=\linewidth]{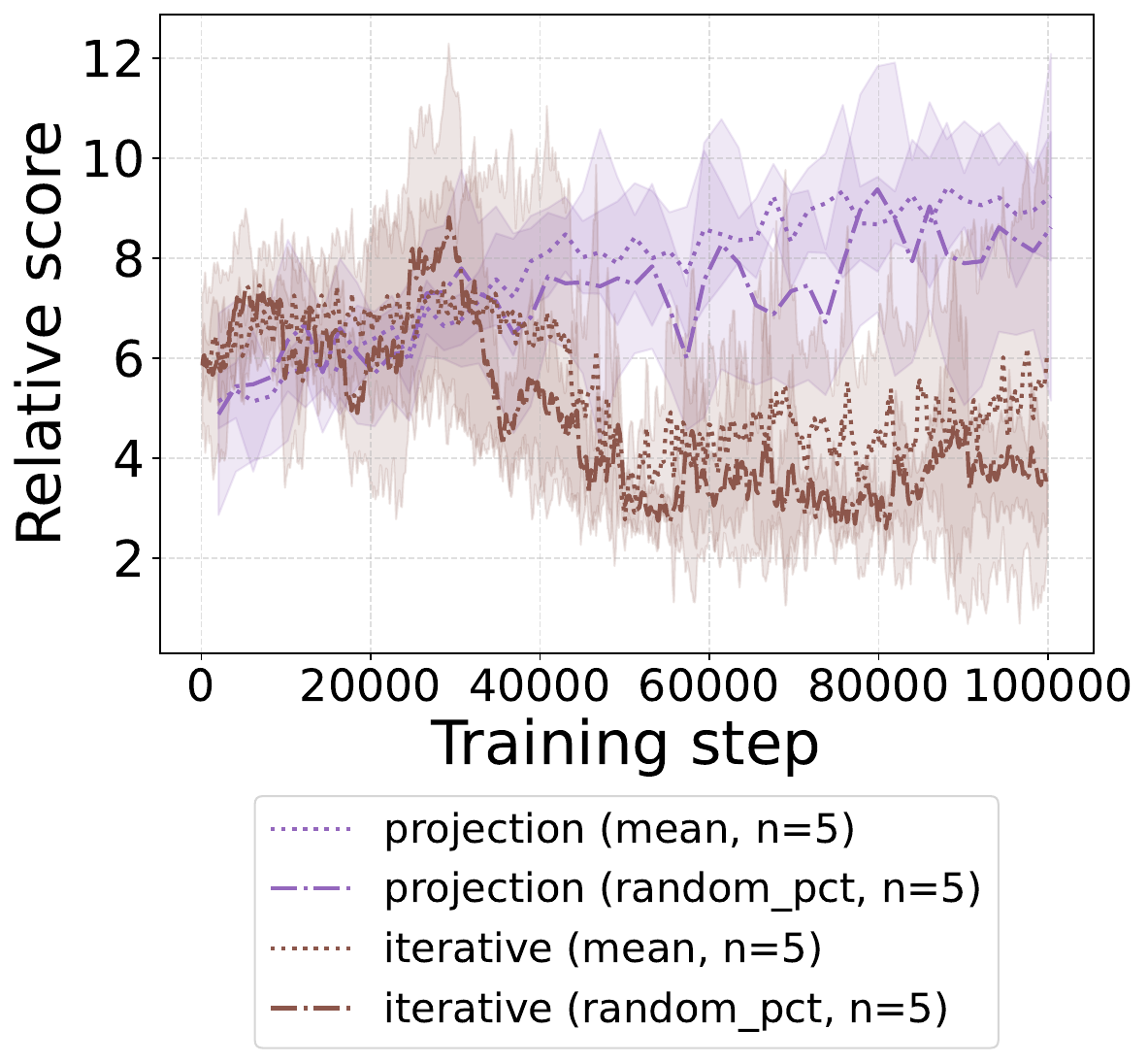}
        \caption{Cyber-Path Latent M/V}
    \end{subfigure}

  \caption{Average learning curves (mean ± standard deviation) across Cyber-Path benchmark environments under all training configurations.}
    \label{fig:all_learning_curves_cyber}
\end{figure}

\begin{figure}[H]
    \centering

    \begin{subfigure}{0.32\textwidth}
        \centering
        \includegraphics[width=\linewidth]{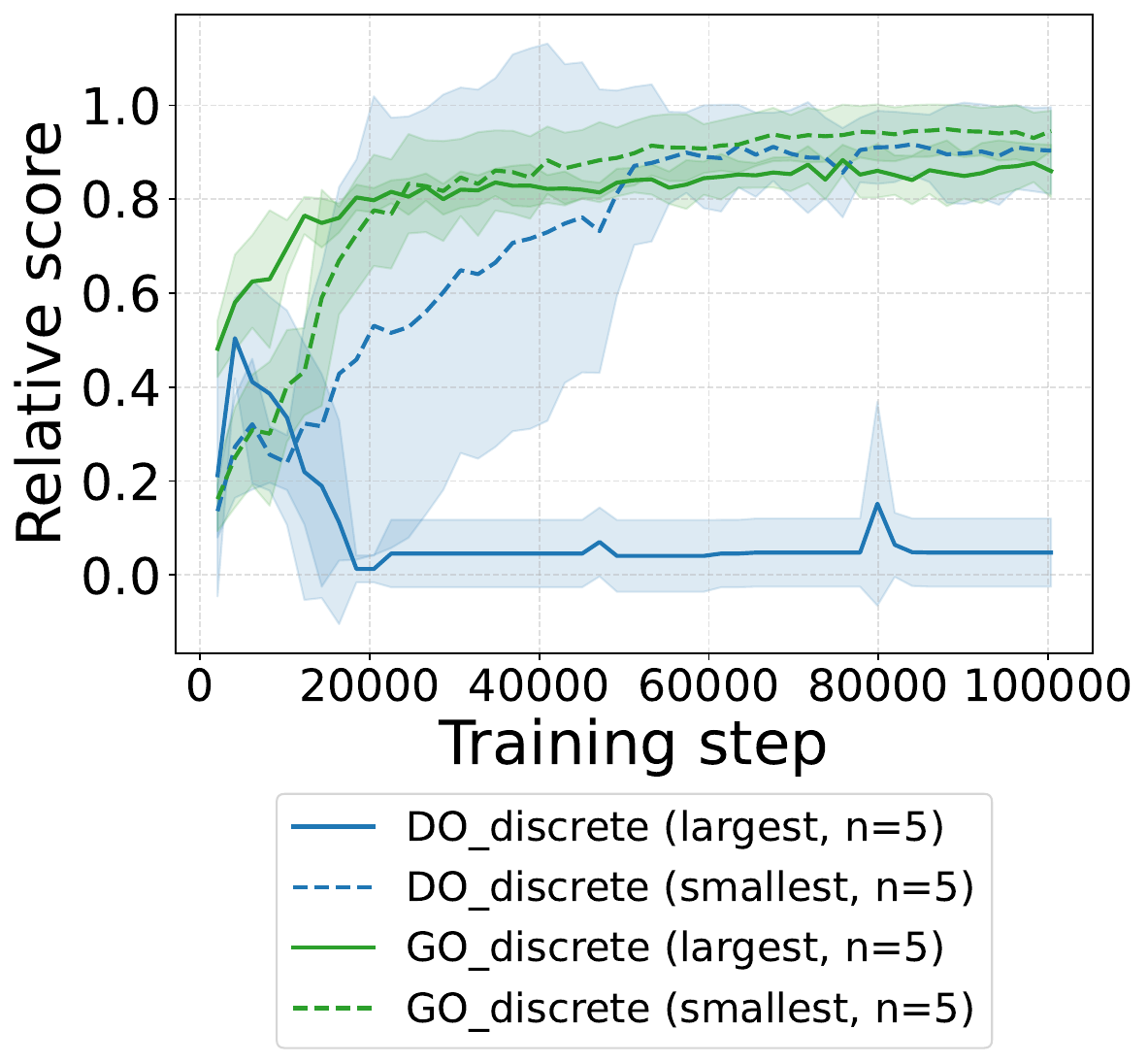}
        \caption{OSPF Discrete S/L}
    \end{subfigure}
    \hfill
    \begin{subfigure}{0.32\textwidth}
        \centering
        \includegraphics[width=\linewidth]{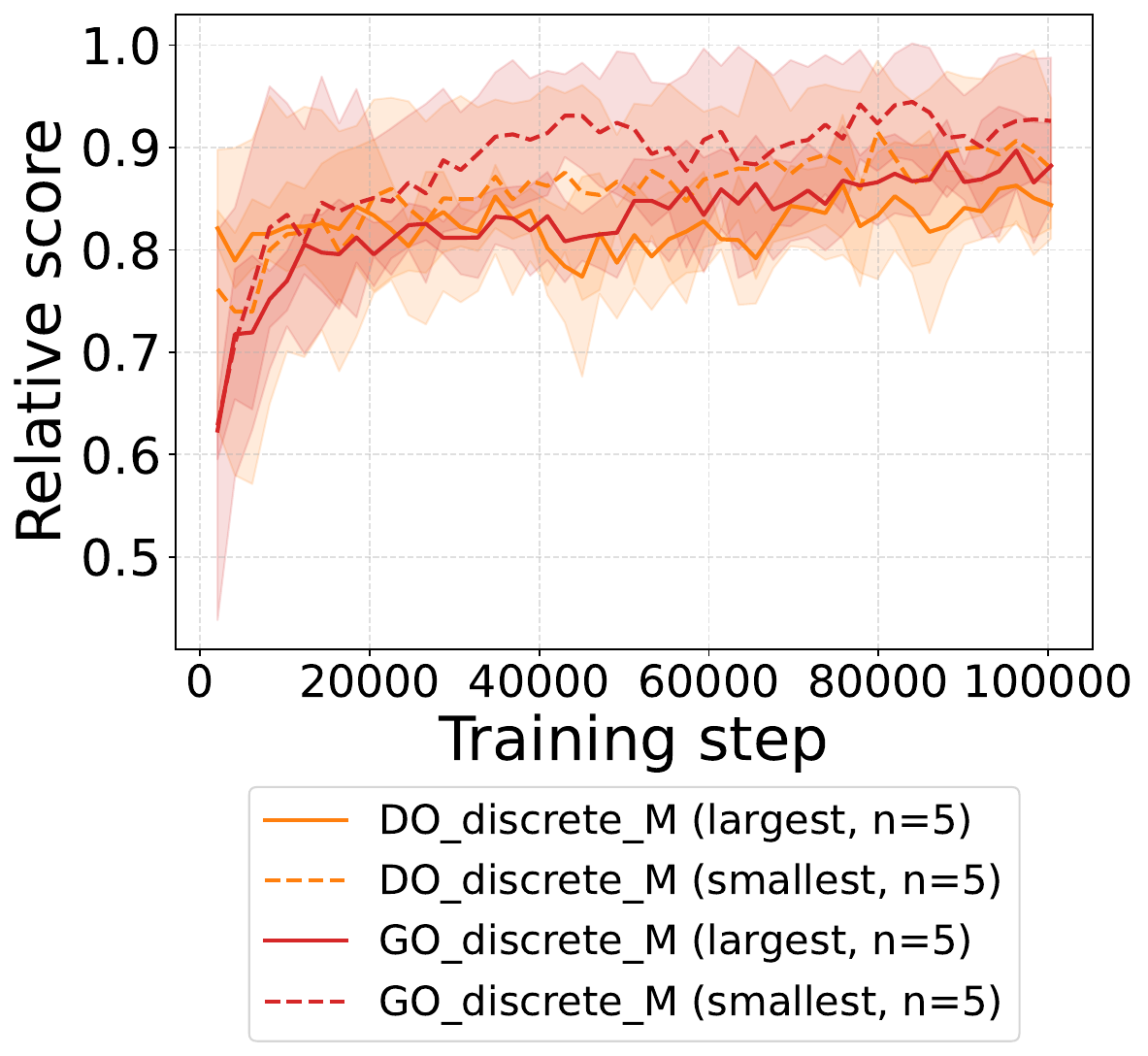}
        \caption{OSPF Discrete-M S/L}
    \end{subfigure}
    \hfill
    \begin{subfigure}{0.32\textwidth}
        \centering
        \includegraphics[width=\linewidth]{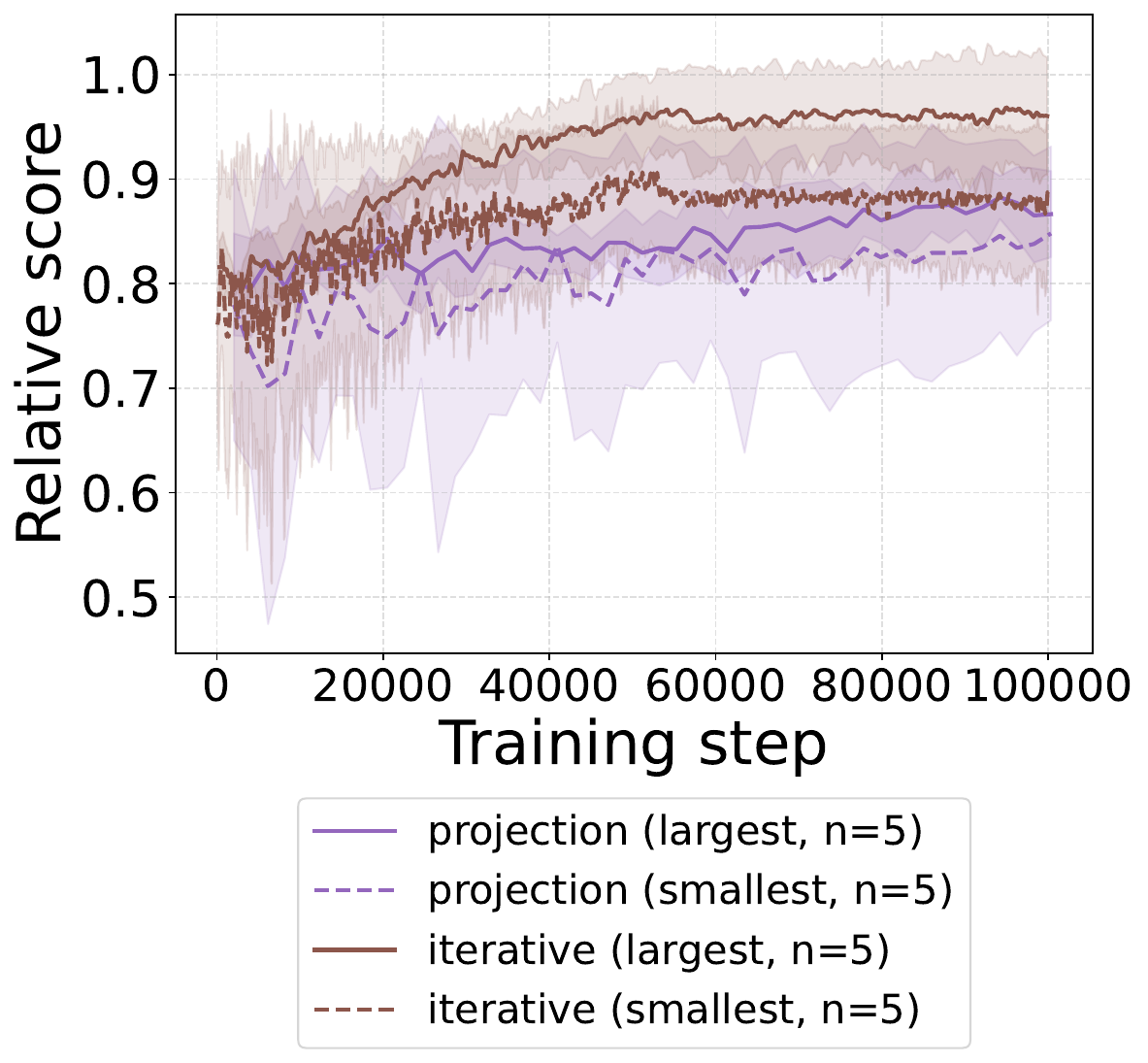}
        \caption{OSPF Latent S/L}
    \end{subfigure}

    \begin{subfigure}{0.32\textwidth}
        \centering
        \includegraphics[width=\linewidth]{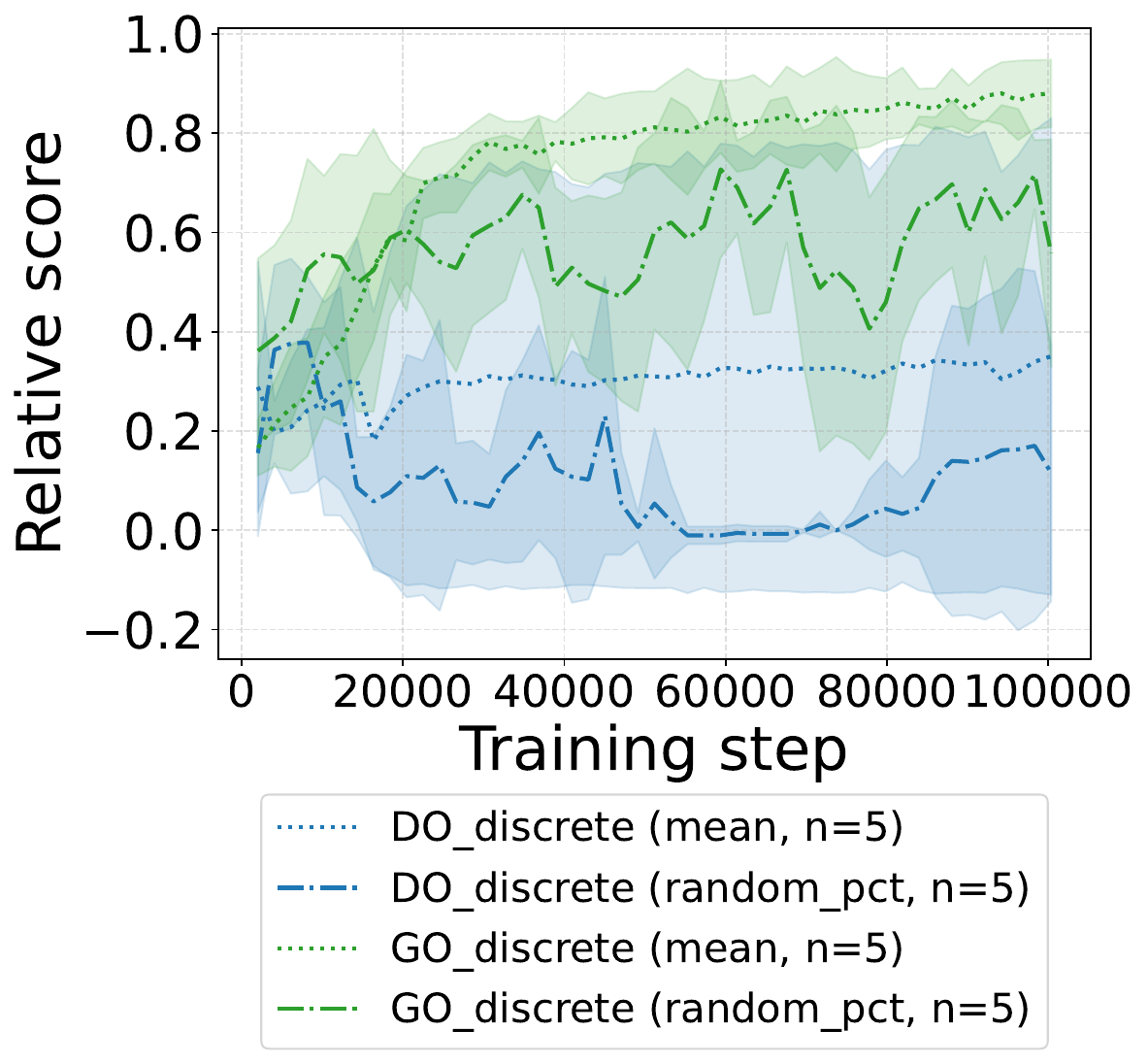}
        \caption{OSPF Discrete M/V}
    \end{subfigure}
    \hfill
    \begin{subfigure}{0.32\textwidth}
        \centering
        \includegraphics[width=\linewidth]{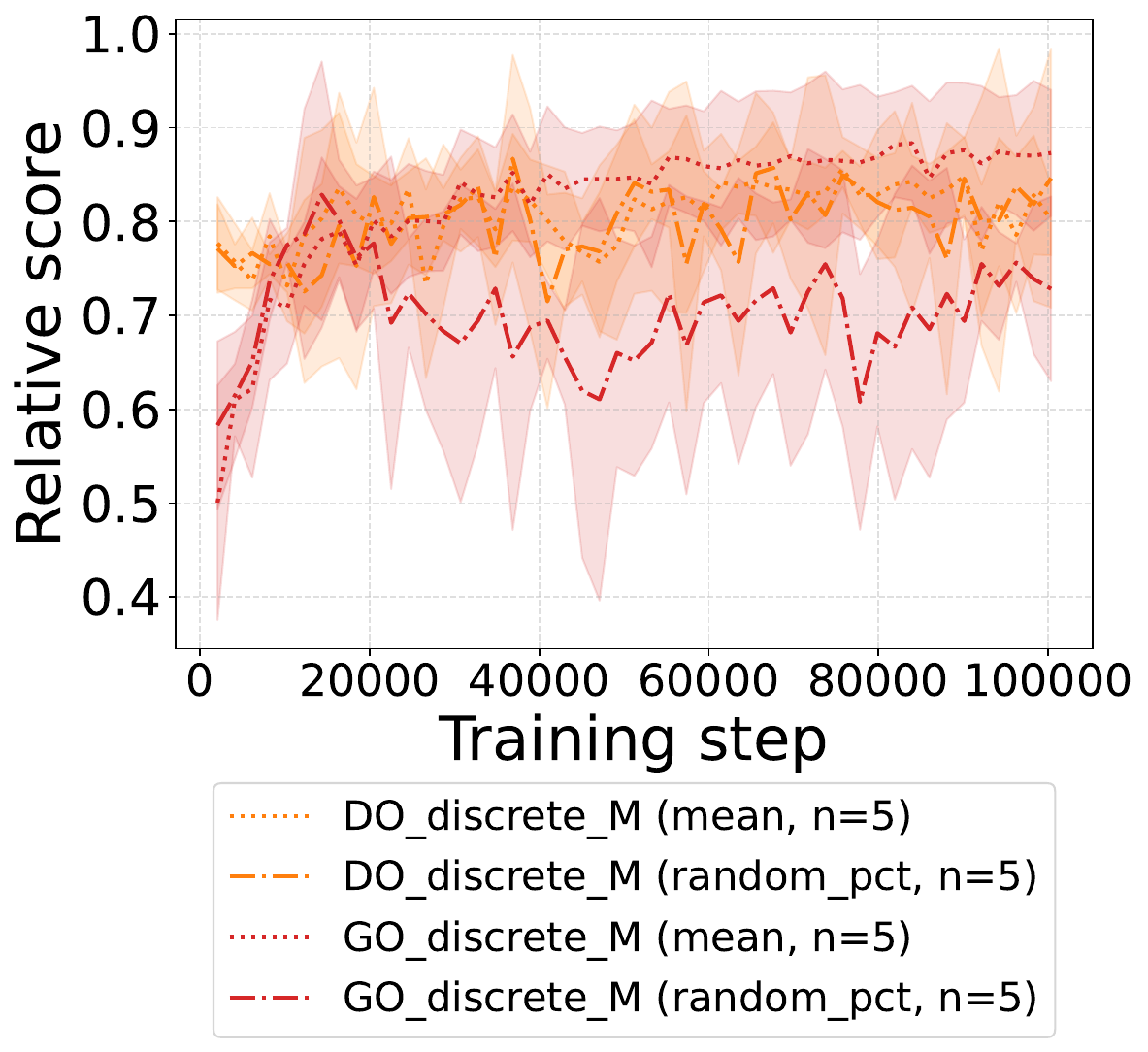}
        \caption{OSPF Discrete-M M/V}
    \end{subfigure}
    \hfill
    \begin{subfigure}{0.32\textwidth}
        \centering
        \includegraphics[width=\linewidth]{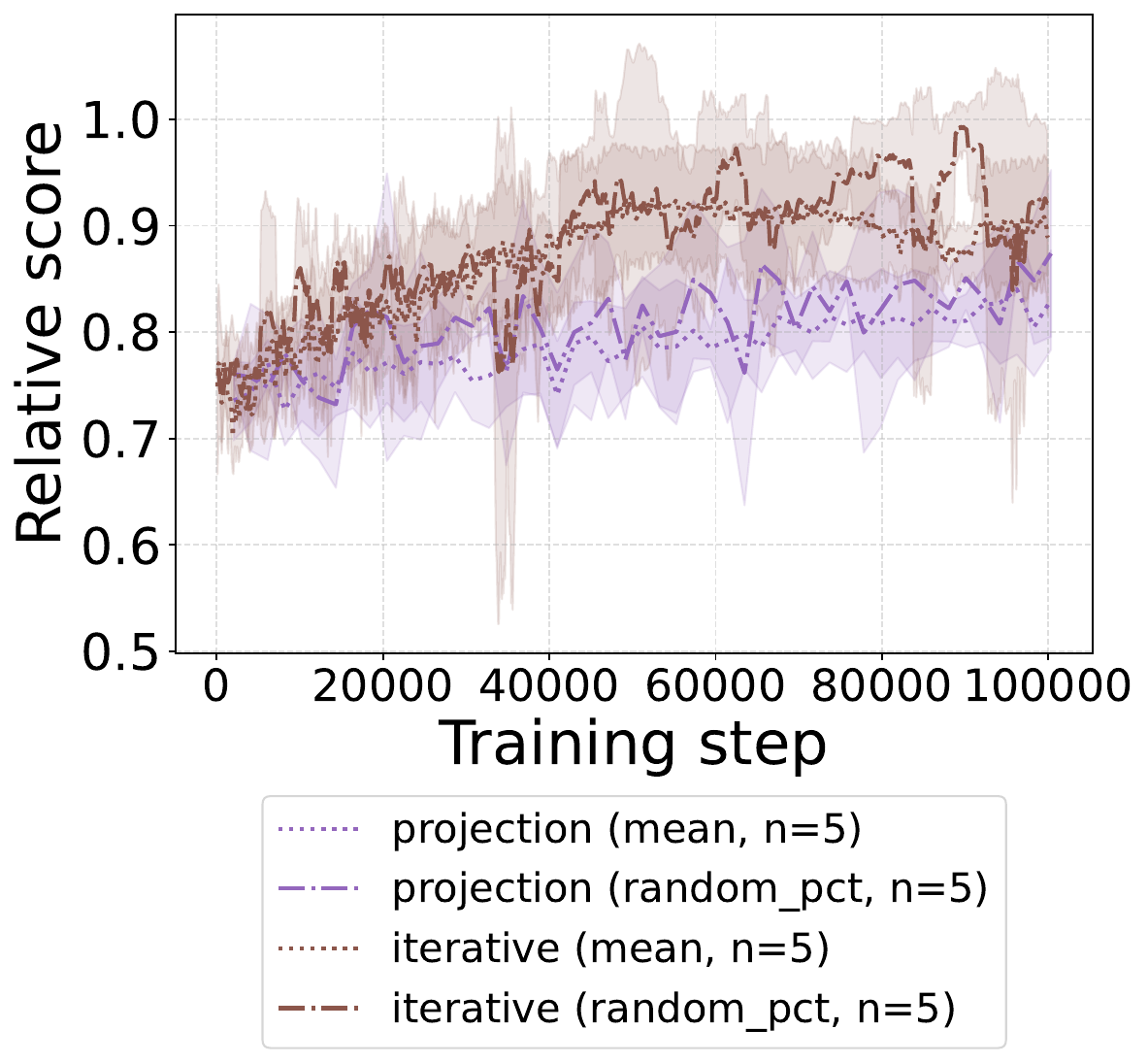}
        \caption{OSPF Latent M/V}
    \end{subfigure}

  \caption{Average learning curves (mean ± standard deviation) across OSPF benchmark environments under all training configurations.}
    \label{fig:all_learning_curves_OSPF}
\end{figure}

\begin{figure}[H]
    \centering

    \begin{subfigure}{0.32\textwidth}
        \centering
        \includegraphics[width=\linewidth]{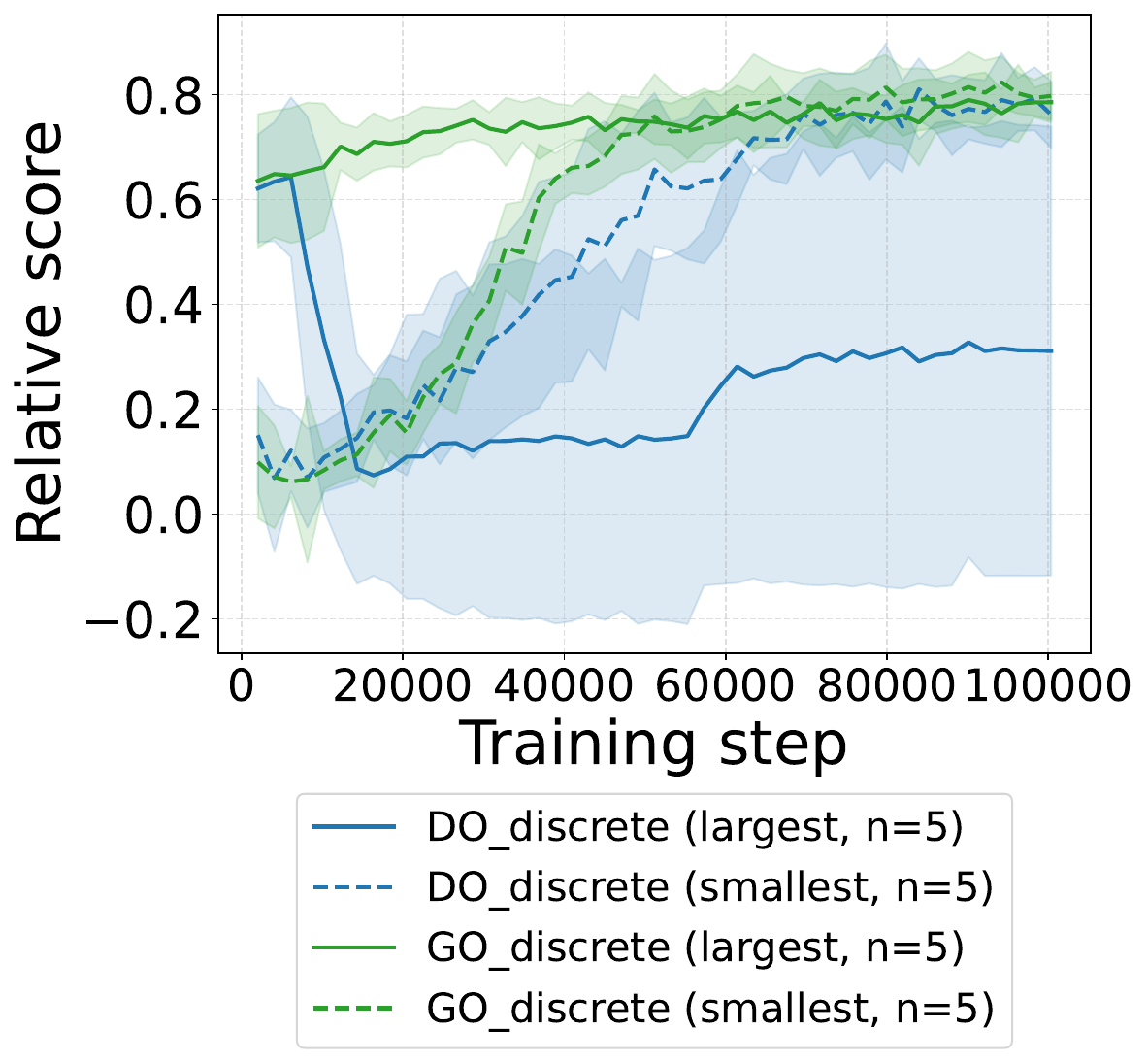}
        \caption{Traffic Discrete S/L}
    \end{subfigure}
    \hfill
    \begin{subfigure}{0.32\textwidth}
        \centering
        \includegraphics[width=\linewidth]{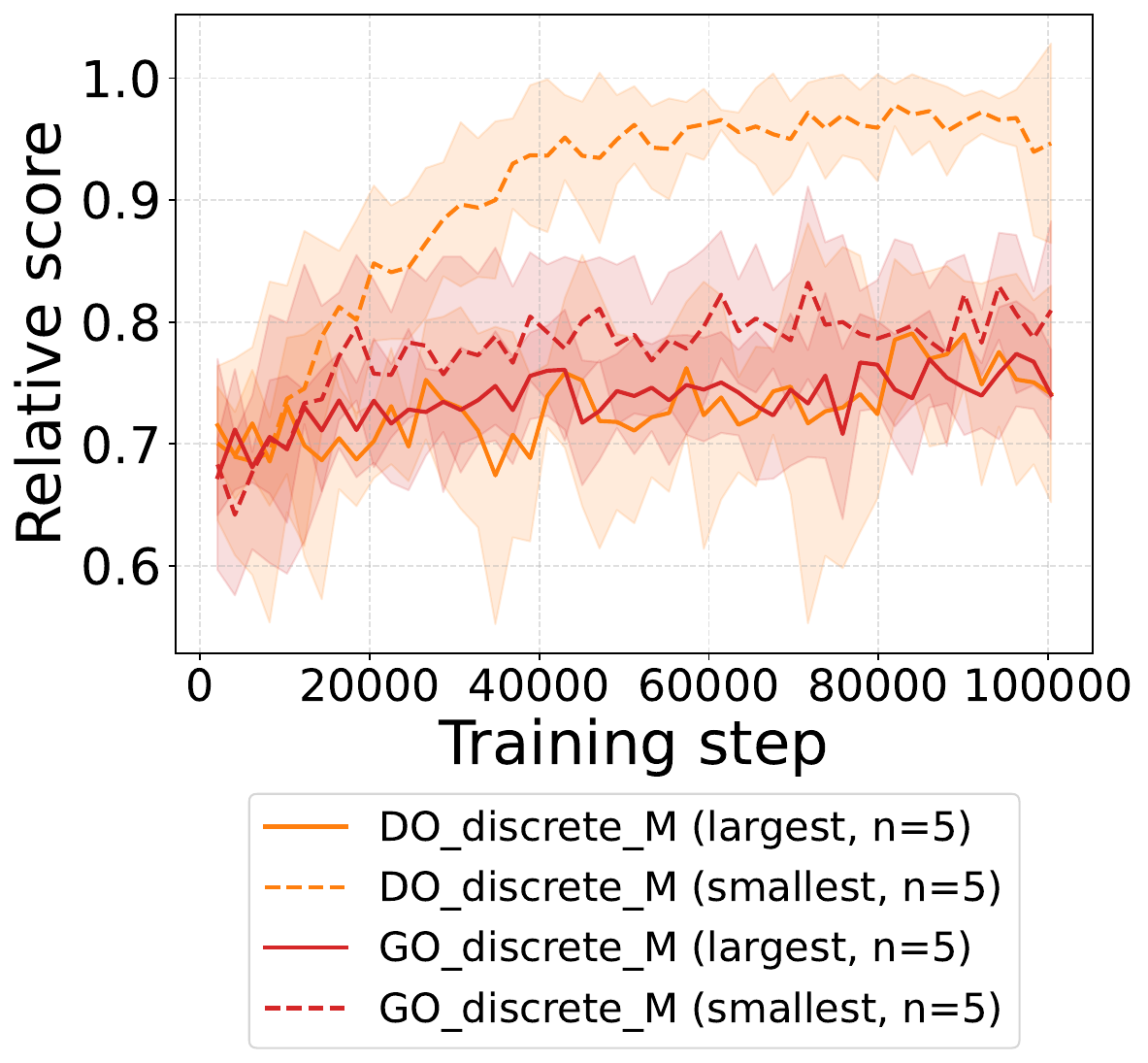}
        \caption{Traffic Discrete-M S/L}
    \end{subfigure}
    \hfill
    \begin{subfigure}{0.32\textwidth}
        \centering
        \includegraphics[width=\linewidth]{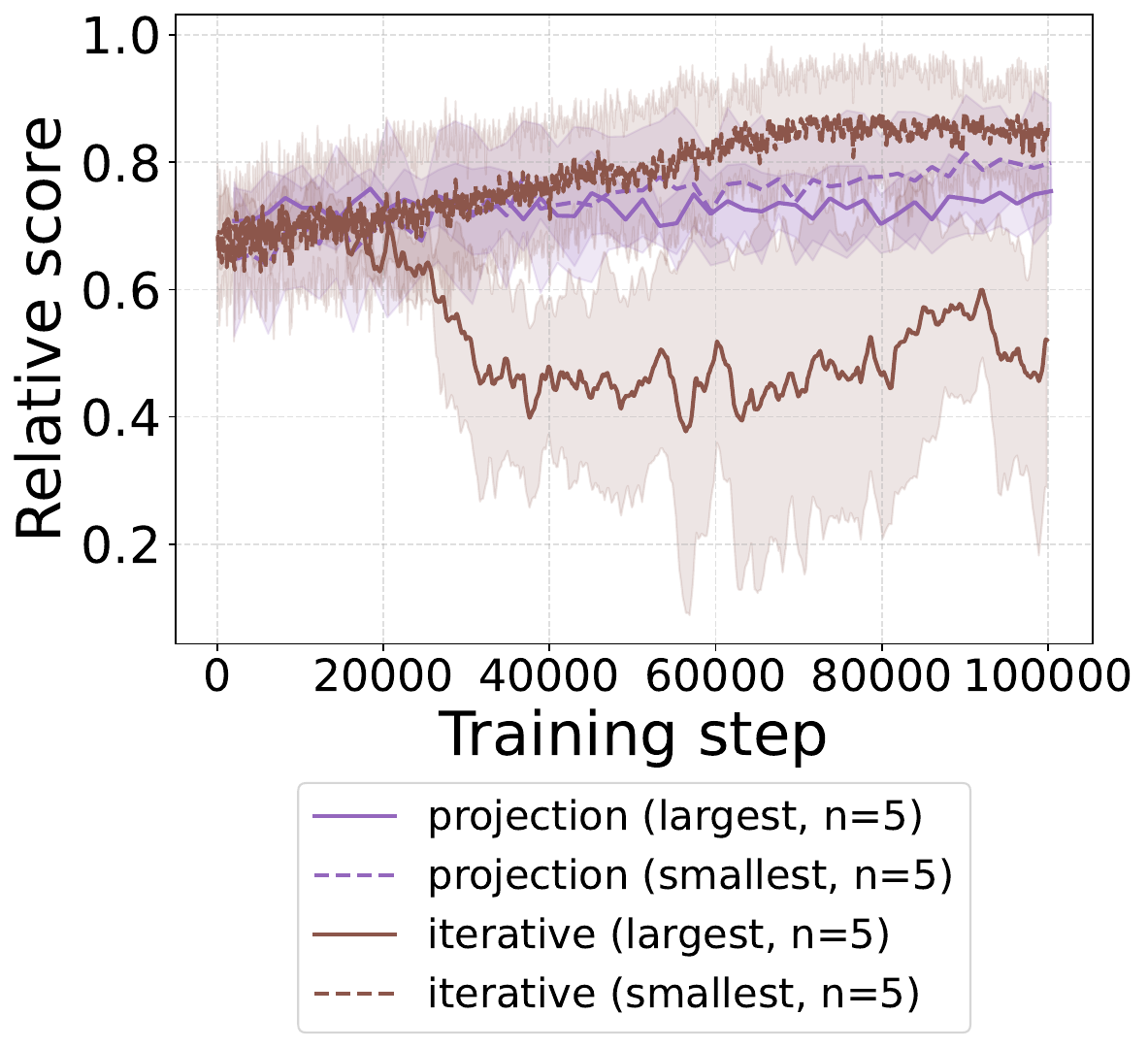}
        \caption{Traffic Latent S/L}
    \end{subfigure}

    \begin{subfigure}{0.32\textwidth}
        \centering
        \includegraphics[width=\linewidth]{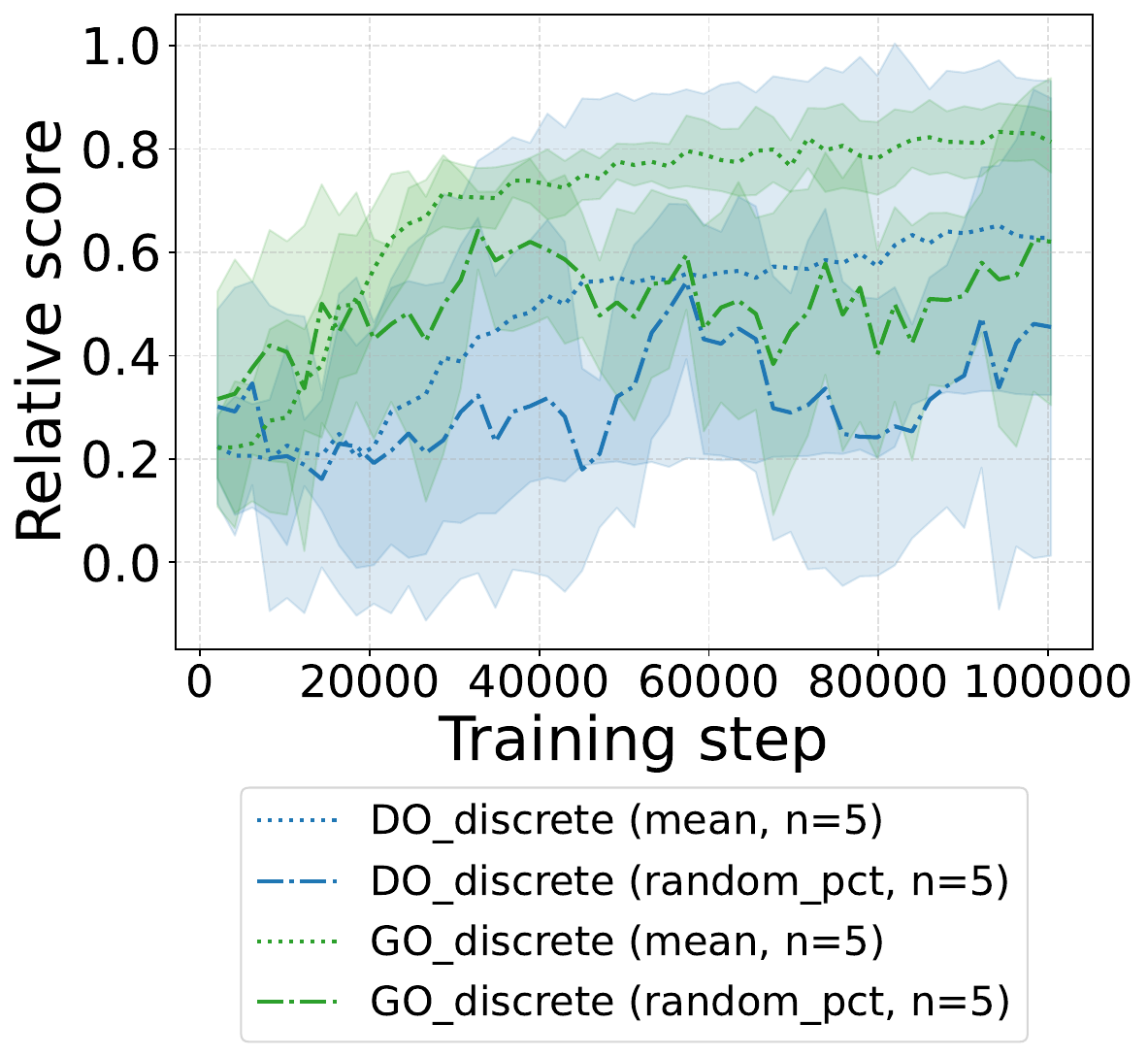}
        \caption{Traffic Discrete M/V}
    \end{subfigure}
    \hfill
    \begin{subfigure}{0.32\textwidth}
        \centering
        \includegraphics[width=\linewidth]{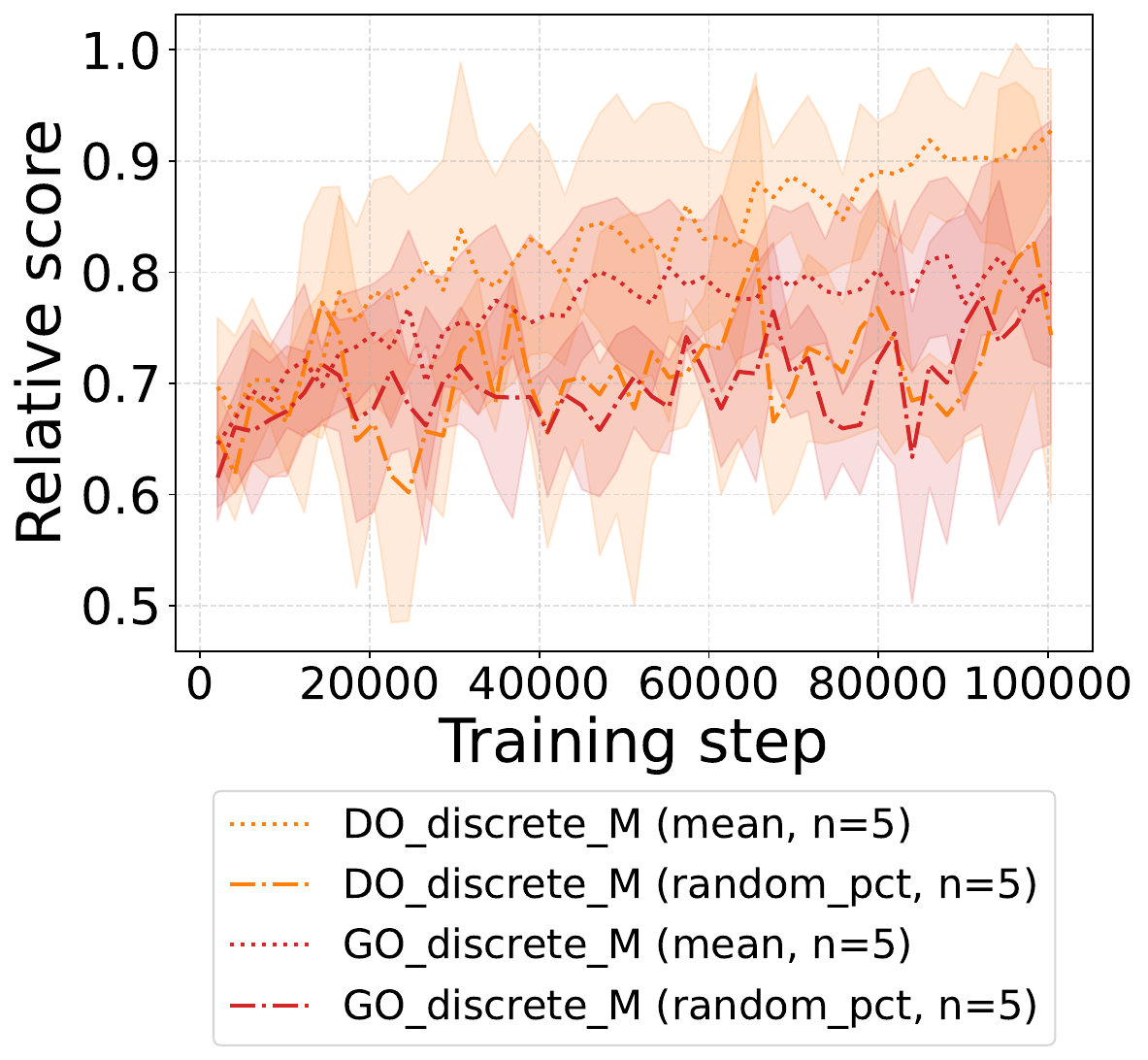}
        \caption{Traffic Discrete-M M/V}
    \end{subfigure}
    \hfill
    \begin{subfigure}{0.32\textwidth}
        \centering
        \includegraphics[width=\linewidth]{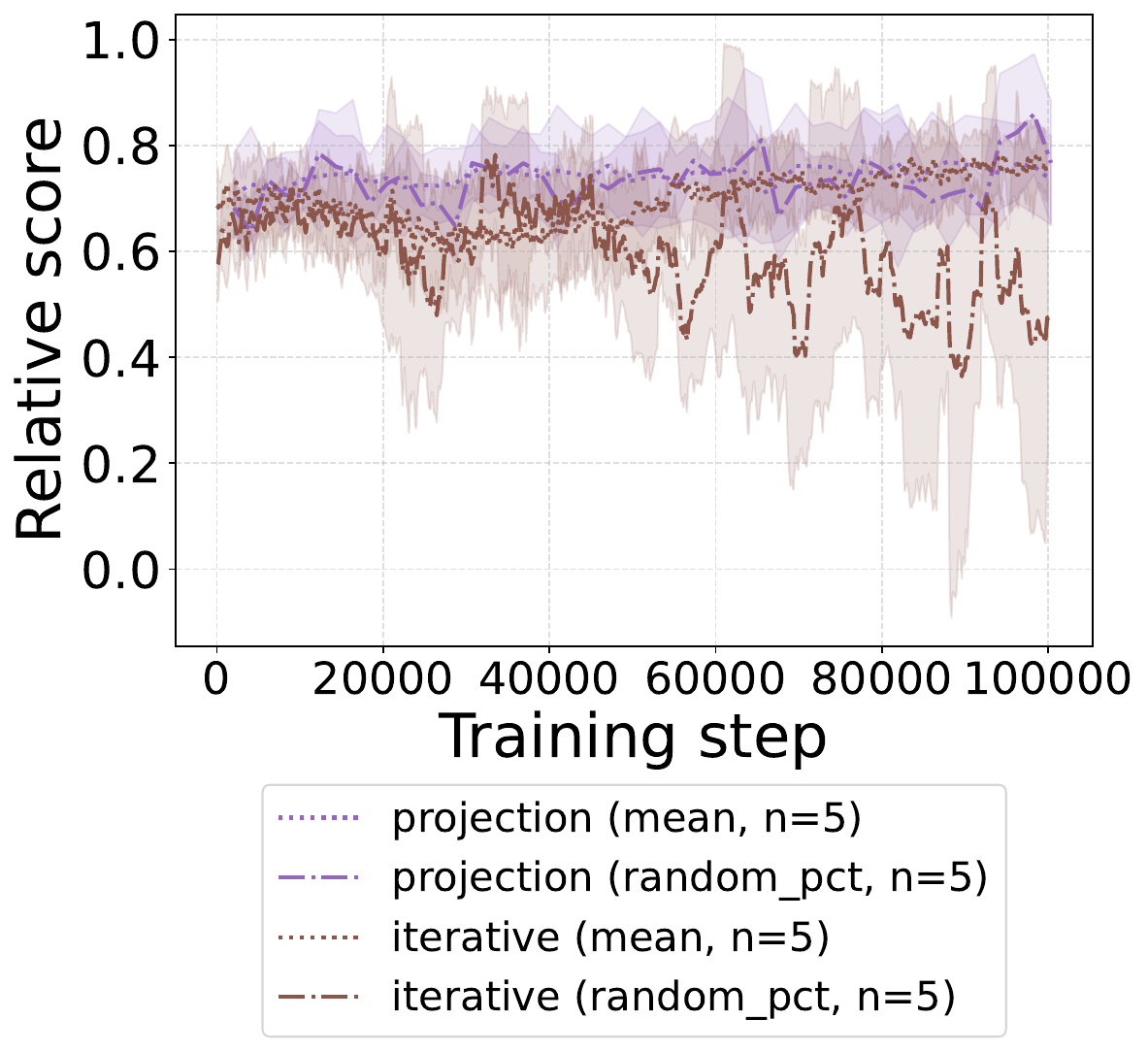}
        \caption{Traffic Latent M/V}
    \end{subfigure}

  \caption{Average learning curves (mean ± standard deviation) across Traffic benchmark environments under all training configurations.}
    \label{fig:all_learning_curves_Traffic}
\end{figure}

Figures \ref{fig:all_learning_curves_travel}, \ref{fig:all_learning_curves_minvertex}, \ref{fig:all_learning_curves_MaxCut}, \ref{fig:all_learning_curves_Placement}, \ref{fig:all_learning_curves_cyber}, \ref{fig:all_learning_curves_OSPF} and \ref{fig:all_learning_curves_Traffic} show the training learning curves for the different approaches, reporting the mean and standard deviation computed over the five independent runs per training strategy. These curves also highlight that several runs fail to converge, particularly for iterative methods in real-world settings and for discrete approaches in some benchmark-strategy pairs when masking is not applied.

\begin{figure}[H]
    \centering
    
    \begin{subfigure}{0.32\textwidth}
        \centering
        \includegraphics[width=\linewidth]{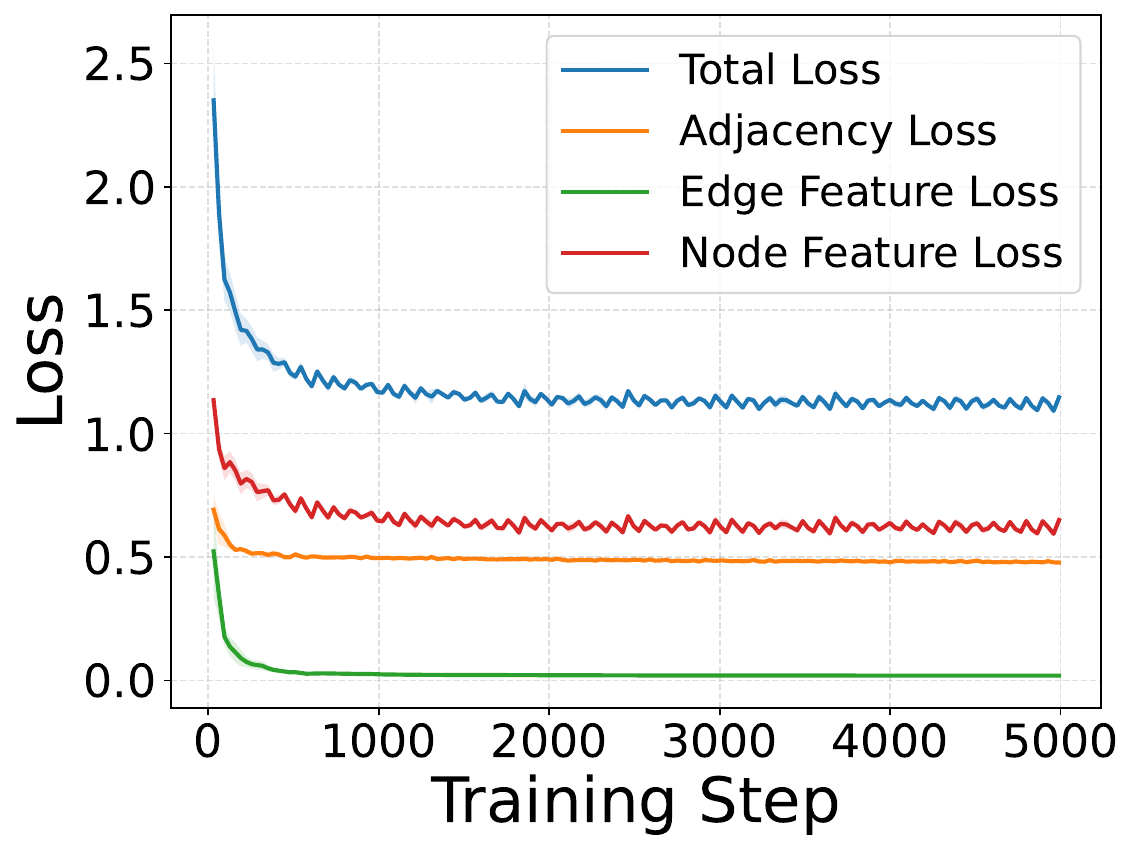}
        \caption{TSP}
    \end{subfigure}
    \hfill
    \begin{subfigure}{0.32\textwidth}
        \centering
        \includegraphics[width=\linewidth]{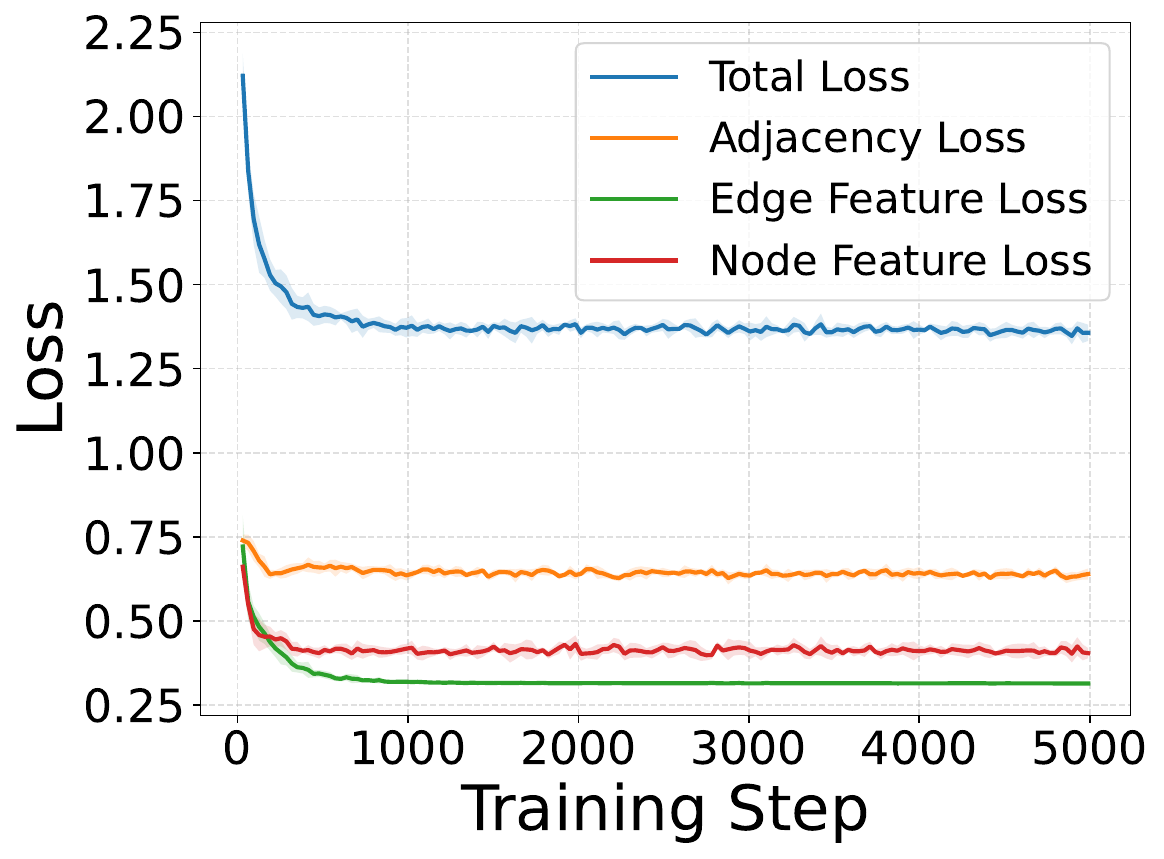}
        \caption{MinVertex}
    \end{subfigure}
    \hfill
    \begin{subfigure}{0.32\textwidth}
        \centering
        \includegraphics[width=\linewidth]{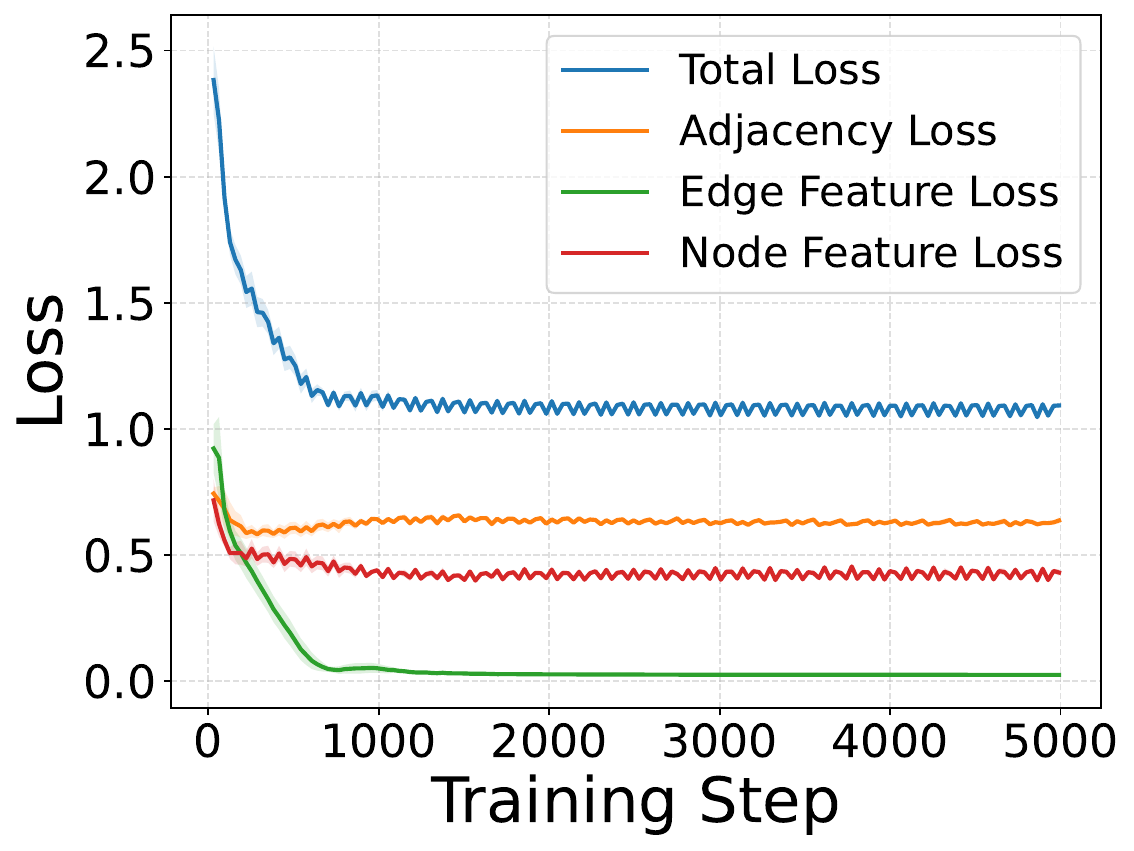}
        \caption{MaxCut}
    \end{subfigure}

    \begin{subfigure}{0.24\textwidth}
        \centering
        \includegraphics[width=\linewidth]{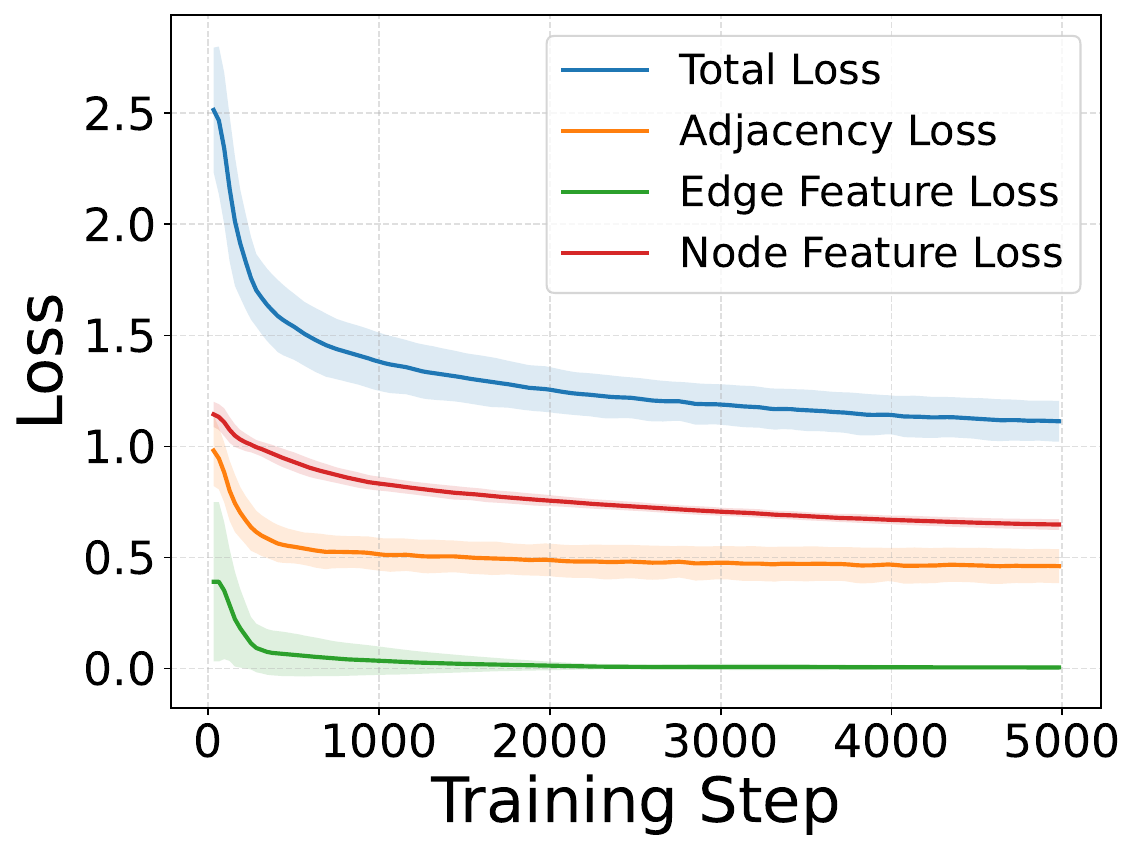}
        \caption{Placement}
    \end{subfigure}
    \hfill
    \begin{subfigure}{0.24\textwidth}
        \centering
        \includegraphics[width=\linewidth]{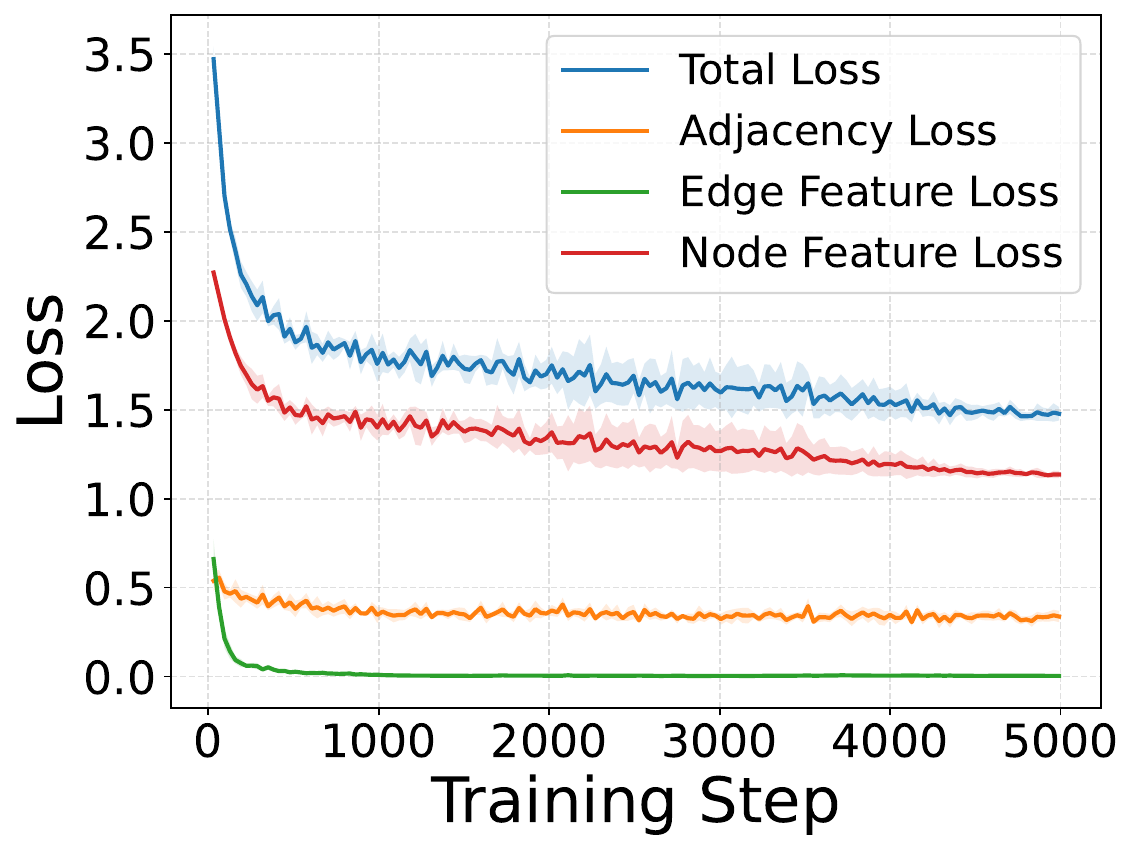}
        \caption{Cyber-Path}
    \end{subfigure}
    \hfill
    \begin{subfigure}{0.24\textwidth}
        \centering
        \includegraphics[width=\linewidth]{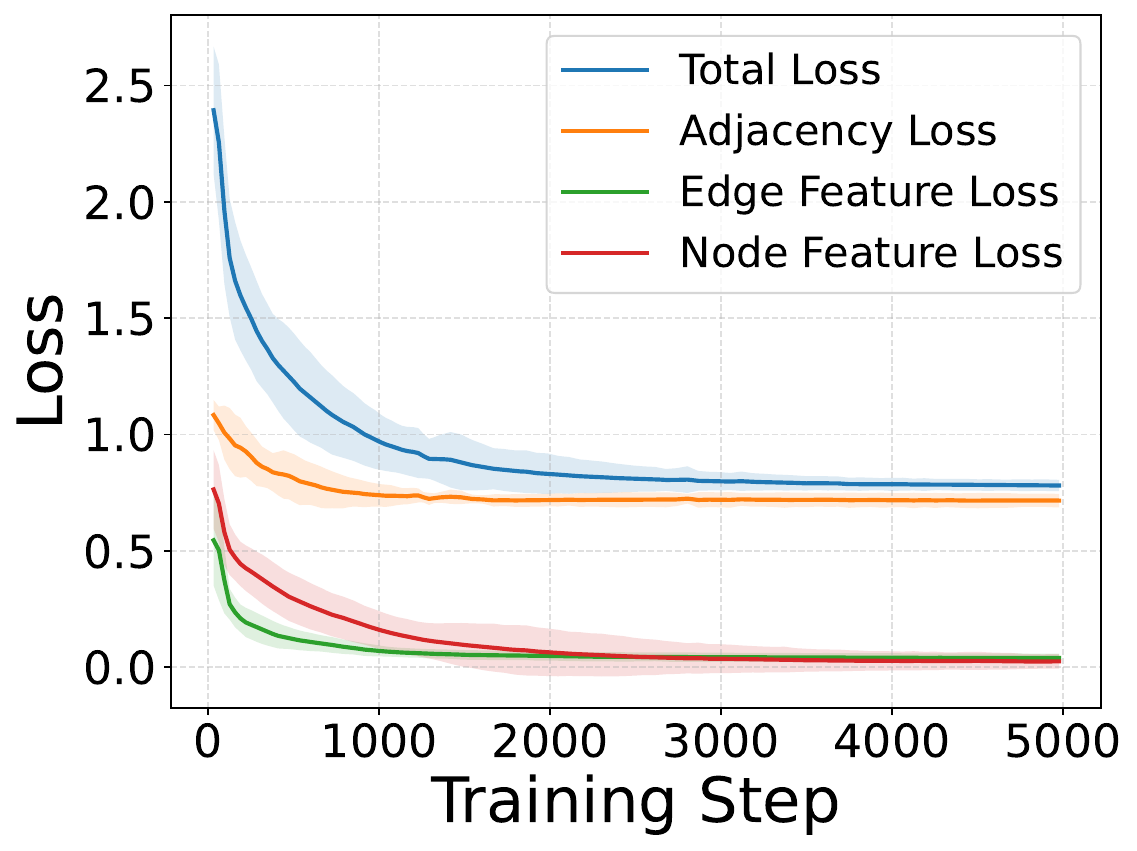}
        \caption{OSPF Engineering}
    \end{subfigure}
    \hfill
    \begin{subfigure}{0.24\textwidth}
        \centering
        \includegraphics[width=\linewidth]{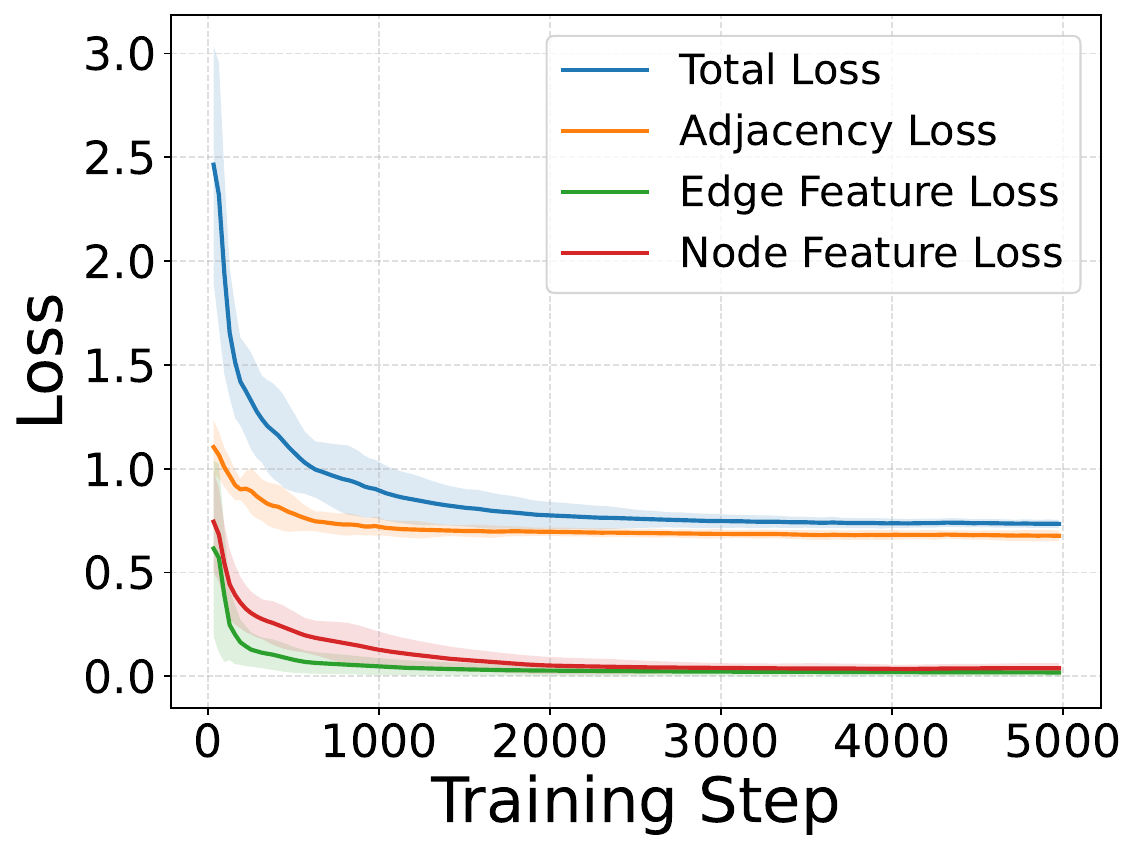}
        \caption{Traffic Engineering}
    \end{subfigure}

    \caption{Graph autoencoder learning curves for five samples of 20 nodes across the seven benchmarks.}
    \label{fig:GAE_curves}
\end{figure}

Figure~\ref{fig:GAE_curves} illustrates the evolution of the total loss during GAE training across benchmark environments, based on five representative 20-node samples and five training runs, with curves aggregated using mean and standard deviation. The results show higher variability on real-world benchmarks.

\section{Training time}
\label{chap:training_time}
We report the computational training cost of the considered approaches on the most demanding benchmark, \textit{Traffic}. 

We report the comparison using the most demanding training strategy, the \textit{varied} (V) training strategy, in which agents are exposed to multiple training environments, resulting in the most costly strategy in terms of training time.

Results computed on the benchmark machine used for the generalization study\footnote{\textit{Benchmark Machine (generalization study):} Dual Intel Xeon Gold 6258R (2 $\times$ 28 cores, 112 threads total) @ 2.70\,GHz, 503\,GiB RAM, NVIDIA RTX A6000 GPU, Ubuntu 22.04.3 LTS.} are reported in Table \ref{tab:full_time} over $5$ independent runs, each using $20$ training environments and $81$ test environments. 

\begin{table*}[t]
\centering
\caption{Training time (seconds) across 5 runs on the traffic engineering benchmark.}
\begin{tabular}{l c c c c c c}
\toprule
Method & Run 1 & Run 2 & Run 3 & Run 4 & Run 5 & Mean $\pm$ Std \\
\midrule

\midrule
P-discrete 
& 671 & 678 & 675 & 675 & 684 & 676 $\pm$ 5 \\
P-discrete-M
& 779 & 778 & 805 & 783 & 792 & 787 $\pm$ 10 \\
G-discrete 
& 1198 & 1084 & 3998 & 1300 & 1239 & 1764 $\pm$ 1100 \\
G-discrete-M
& 1738 & 1614 & 5393 & 1759 & 1739 & 2449 $\pm$ 1600 \\
Iterative 
& 3776 & 2717 & 6718 & 2570 & 5095 & 4175 $\pm$ 1500 \\
Projection 
& 5844 & 3384 & 13417 & 5788 & 4869 & 7460 $\pm$ 3800 \\

\bottomrule
\end{tabular}
\label{tab:full_time}
\end{table*}

In general, \textit{projection} methods consistently exhibit the highest computational cost at training time, followed by the \textit{iterative} approach, while \textit{discrete} baselines remain significantly more efficient in the training stage, as expected.

\section{Scalability curves}
\label{chap:scalability_curves}
\begin{figure}[H]
    \centering

     \begin{subfigure}{0.32\textwidth}
        \centering
        \includegraphics[width=\linewidth]{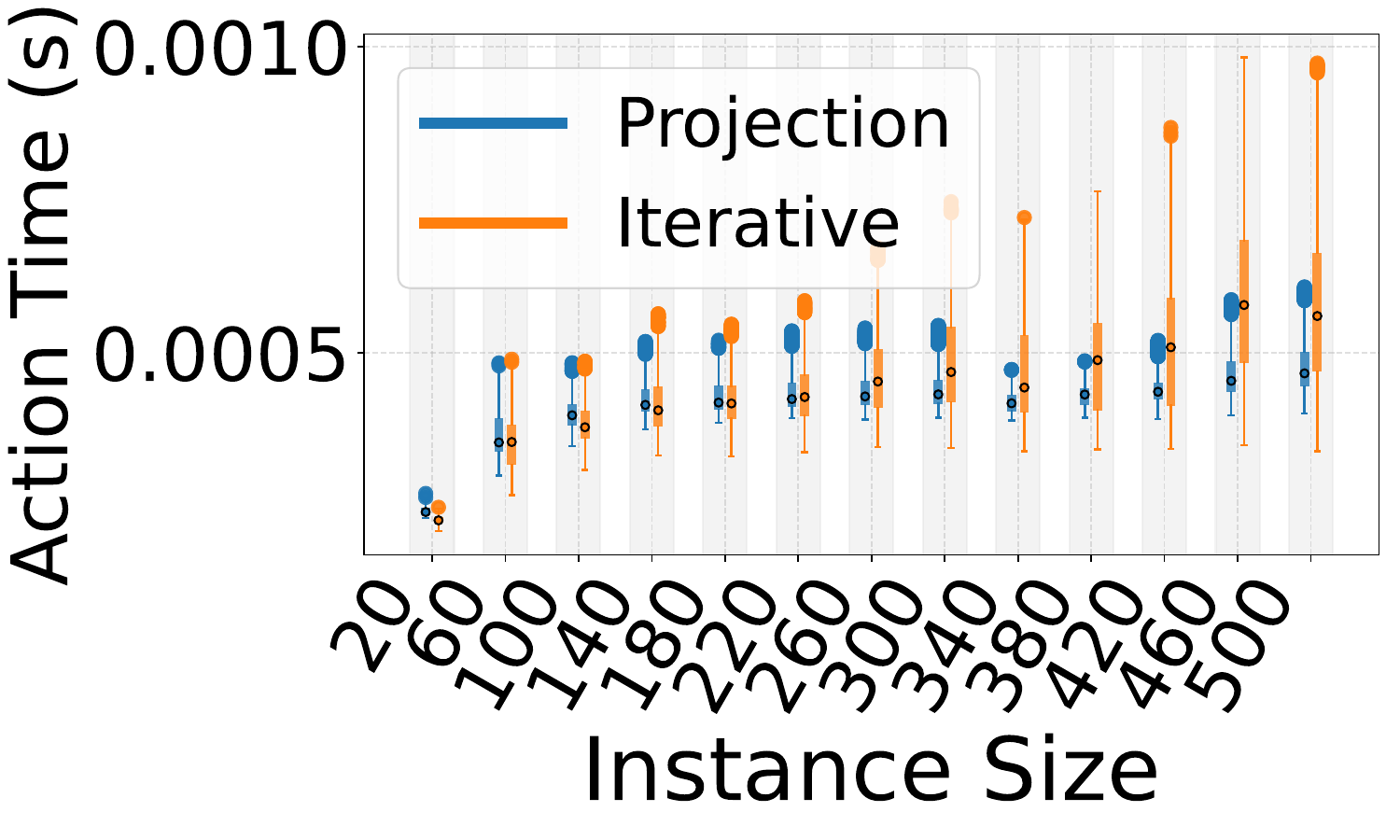}
        \caption{TSP}
    \end{subfigure}
    \hfill
    \begin{subfigure}{0.32\textwidth}
        \centering
        \includegraphics[width=\linewidth]{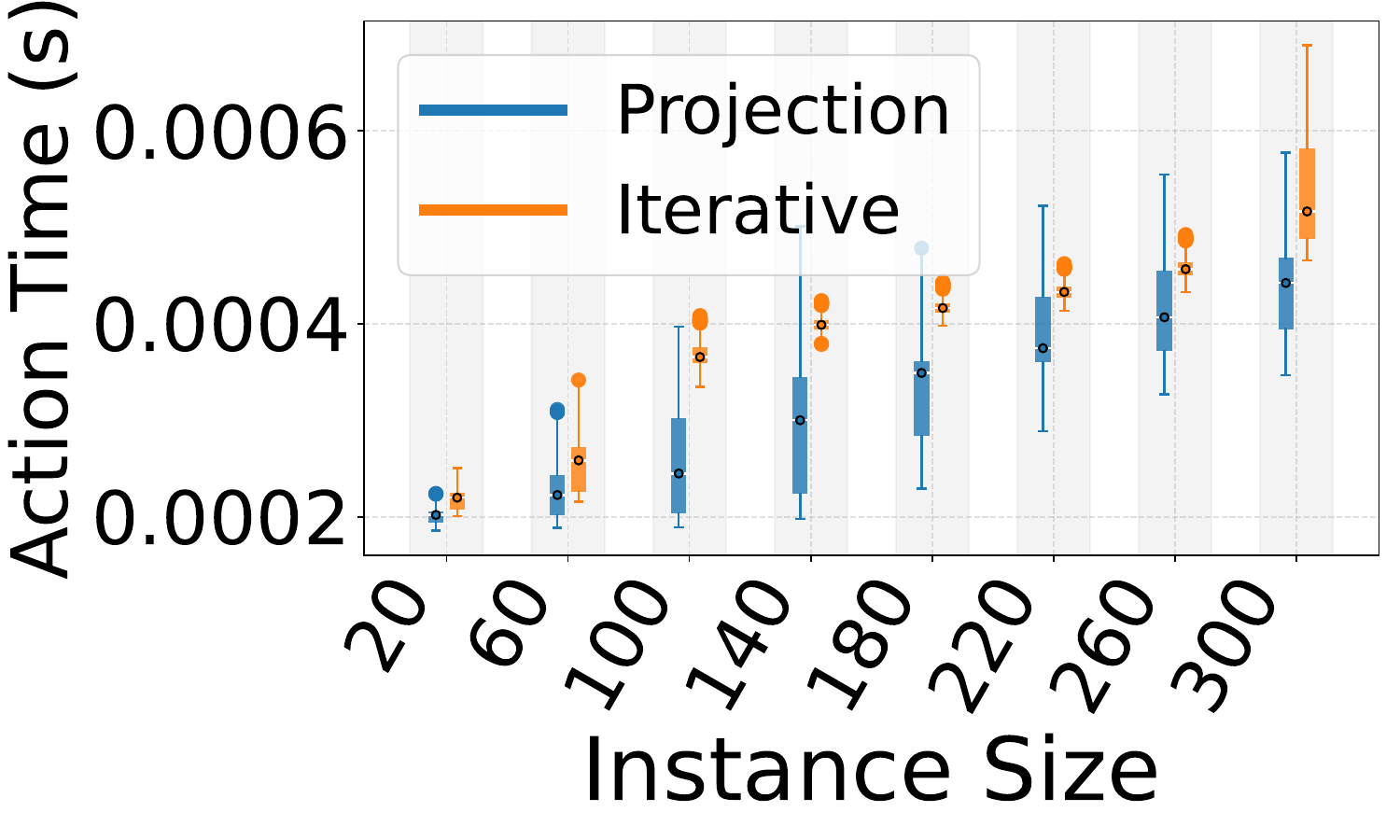}
        \caption{MinVertex}
    \end{subfigure}
    \hfill
    \begin{subfigure}{0.32\textwidth}
        \centering
        \includegraphics[width=\linewidth]{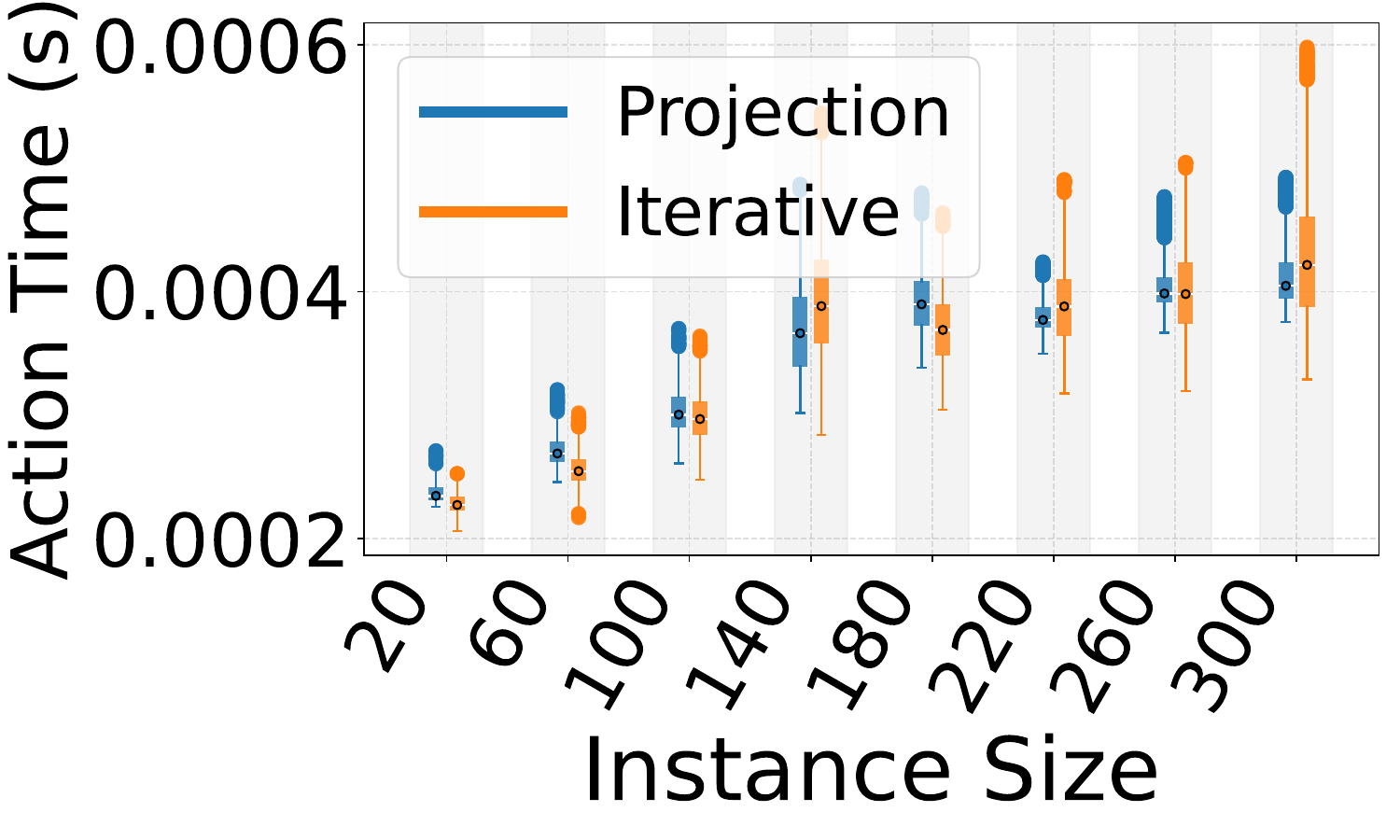}
        \caption{MaxCut}
    \end{subfigure}
    
    \begin{subfigure}{0.24\textwidth}
        \centering
        \includegraphics[width=\linewidth]{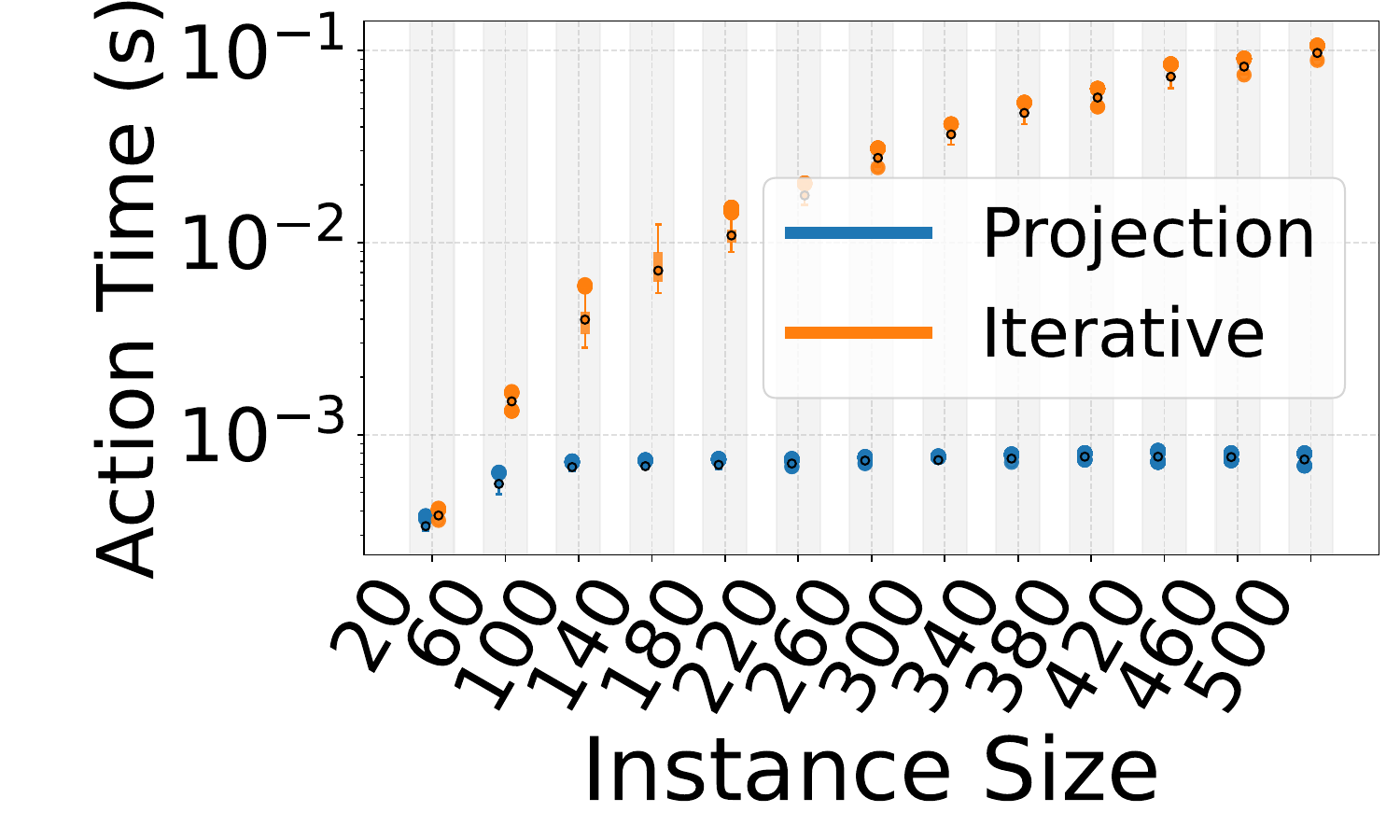}
        \caption{Placement}
    \end{subfigure}
    \hfill
    \begin{subfigure}{0.24\textwidth}
        \centering
        \includegraphics[width=\linewidth]{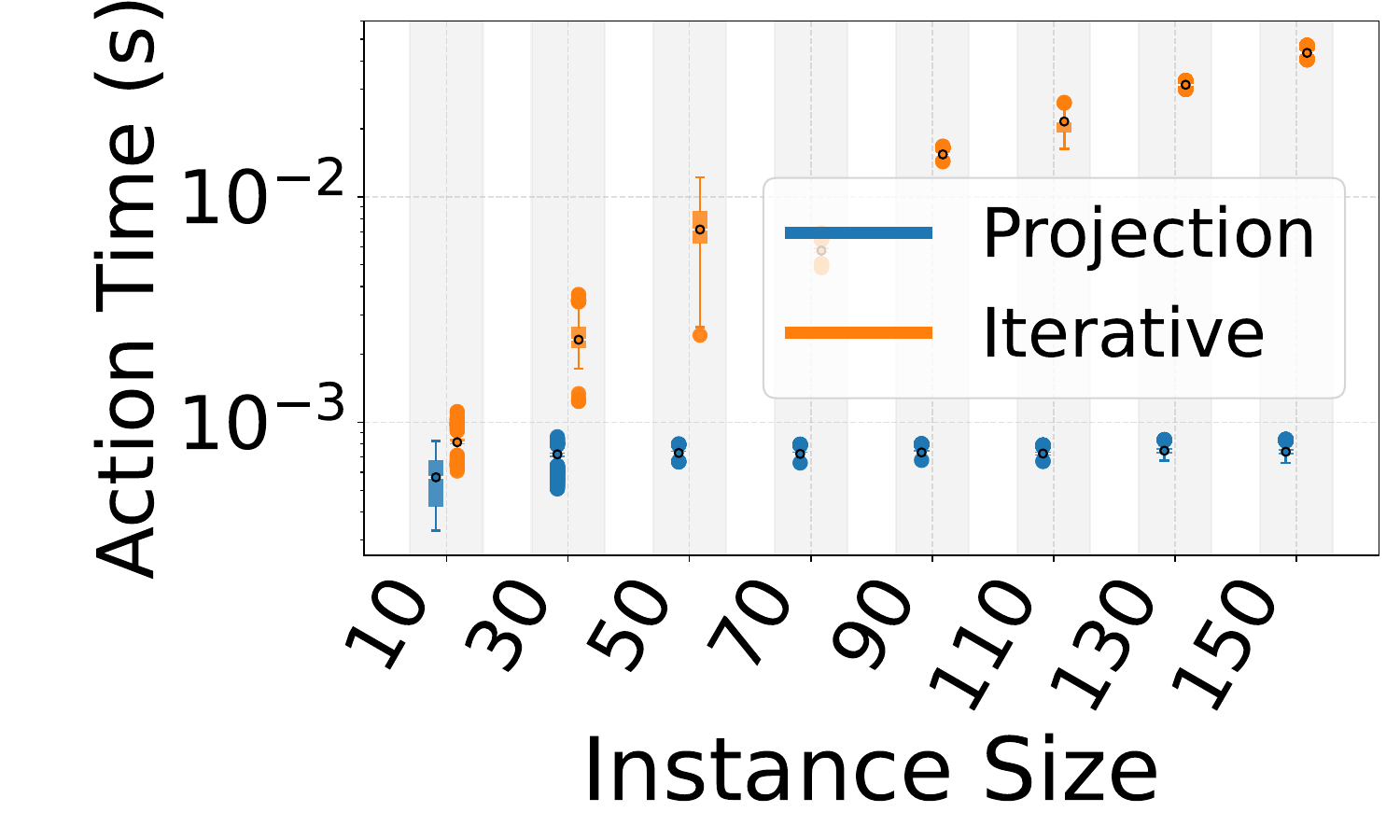}
        \caption{Cyber-Path}
    \end{subfigure}
    \hfill
    \begin{subfigure}{0.24\textwidth}
        \centering
        \includegraphics[width=\linewidth]{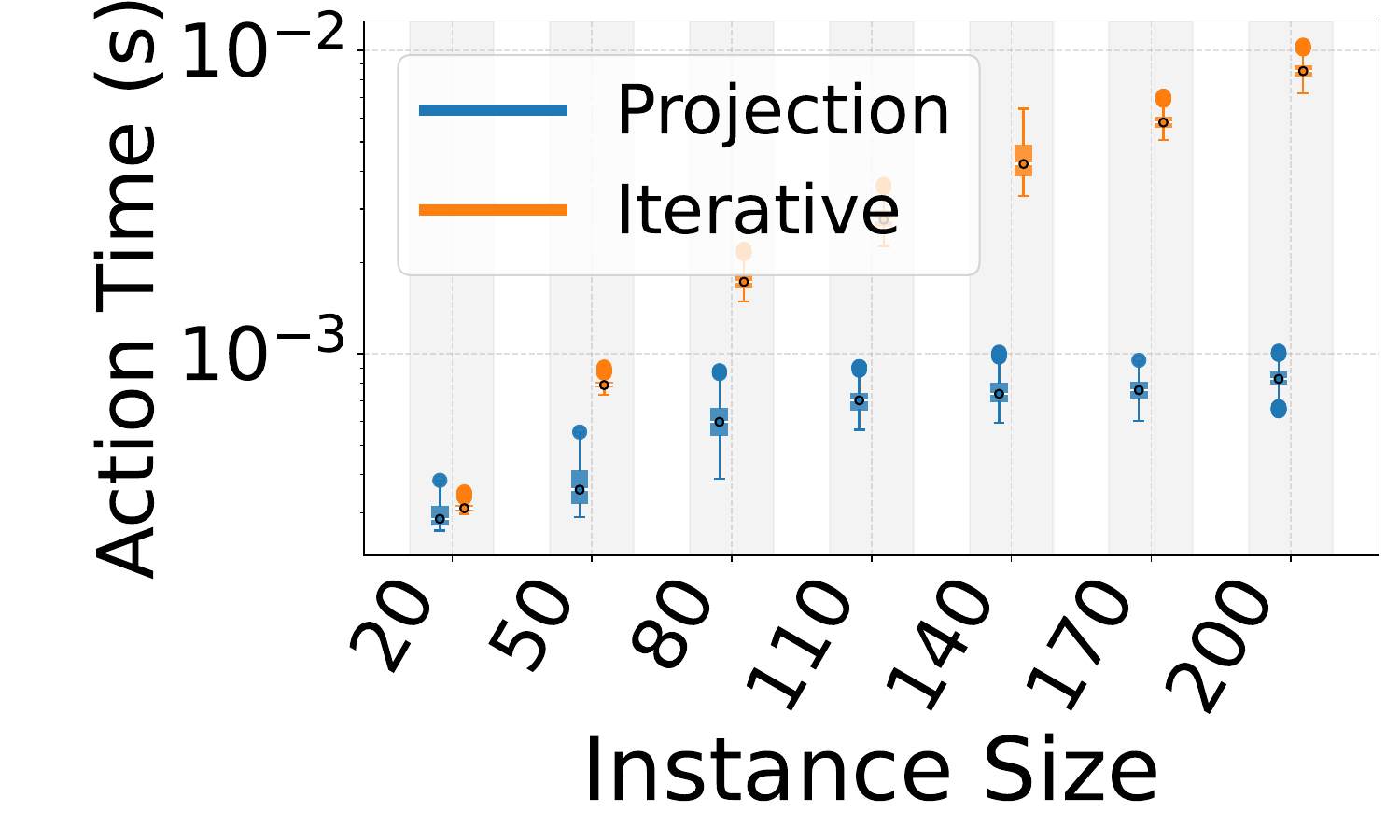}
        \caption{OSPF}
    \end{subfigure}
    \hfill
    \begin{subfigure}{0.24\textwidth}
        \centering
        \includegraphics[width=\linewidth]{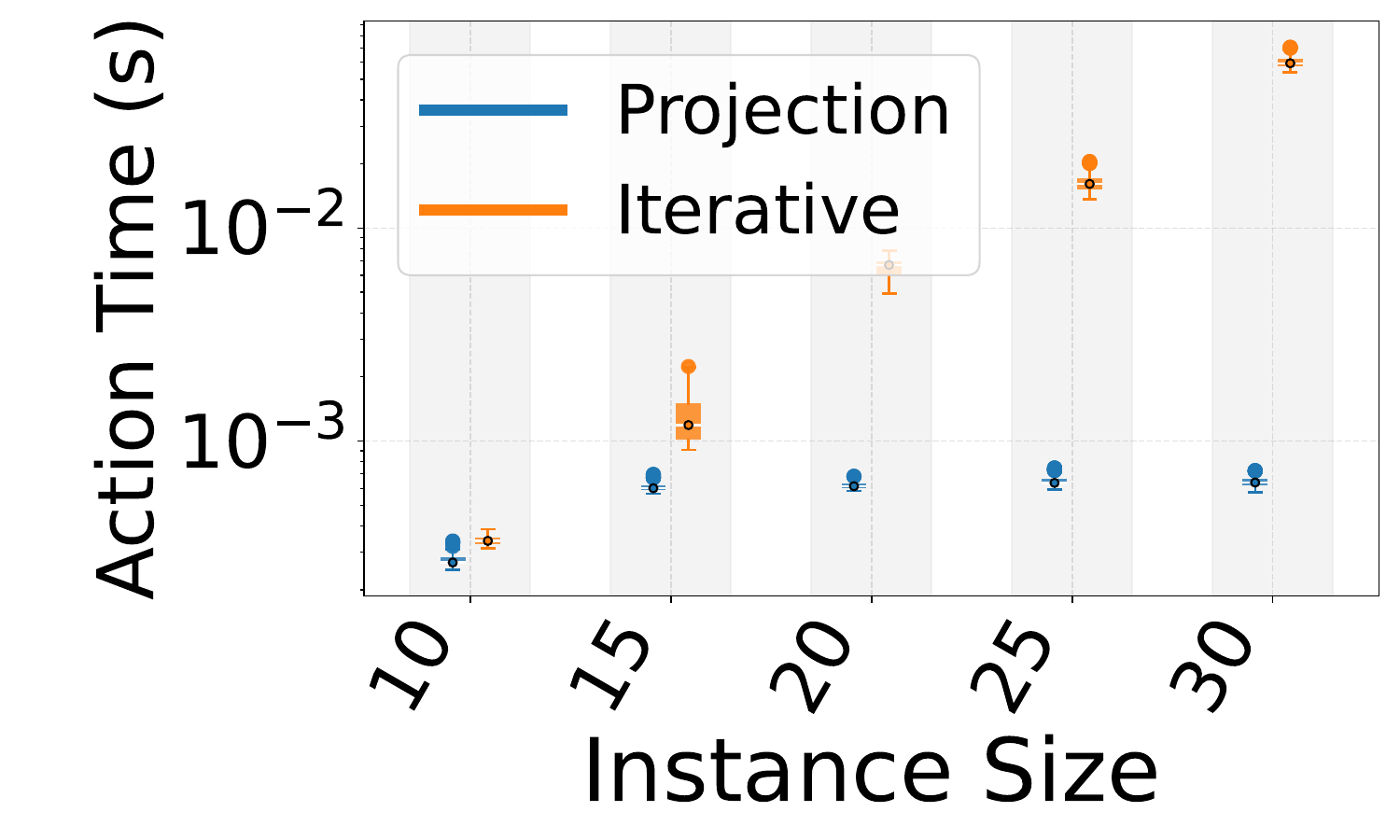}
        \caption{Traffic}
    \end{subfigure}


    \caption{Scalability curve for seven experimental settings.}
    \label{fig:scalability_all}
\end{figure}

Figures \ref{fig:scalability_all} present the full evolution of the action-selection time experiments in Section \ref{chap:efficiency} (evaluation time), including boxplots of the action time distribution across different graph sizes. The figures also indicate the sampled graph sizes for each benchmark, including the maximum size considered to perform the power-law estimation.

\subsection{Approximate Nearest-Neighbor Lookup}
\label{chap:approximateNN}
To further assess the scalability of the \textit{projection} approach, we evaluate FAISS IVF (Inverted File Index), an approximate NN method that partitions the embedding space into clusters and searches only a subset of relevant clusters, reducing retrieval cost compared with exhaustive search. IVF is primarily designed for static embedding sets; however, in our setting, action embeddings evolve throughout an episode because the current partial solution is encoded directly in the graph state (as discussed in Section \ref{chap:conclusion}). Consequently, graph updates modify the action embeddings, requiring the IVF index to be rebuilt or updated after each modification step. Thus, our evaluation does not represent the ideal use case for IVF, but nevertheless demonstrates a measurable speedup.

Table~\ref{tab:ivf_scaling} reports the mean action-selection time for \textit{TSP} across increasing graph sizes. Despite the additional index reconstruction overhead, IVF consistently reduces action time compared with the FAISS Flat Index. The advantage becomes increasingly pronounced at larger graph sizes, with IVF reducing the action time from \(0.49\)~ms/action to \(0.40\)~ms/action at 500 nodes.

\begin{table}[t]
\centering
\caption{Mean action-selection time (ms/action) for \textit{TSP} using FAISS Flat and IVF indexes across graph sizes.}
\label{tab:ivf_scaling}
\begin{tabular}{p{0.5cm}|p{0.5cm}p{0.5cm}p{0.5cm}p{0.5cm}p{0.5cm}p{0.5cm}p{0.5cm}p{0.5cm}p{0.5cm}p{0.5cm}p{0.5cm}p{0.5cm}p{0.5cm}}
\toprule
Size
& 20 & 60 & 100 & 140 & 180 & 220 & 260 & 300 & 340 & 380 & 420 & 460 & 500 \\
\midrule
Flat
& 0.25 & 0.37 & 0.41 & 0.44 & 0.44 & 0.45 & 0.45 & 0.46 & 0.42 & 0.43 & 0.46 & 0.48 & 0.49 \\
IVF
& 0.23 & 0.26 & 0.30 & 0.33 & 0.35 & 0.35 & 0.36 & 0.37 & 0.37 & 0.38 & 0.38 & 0.38 & 0.40 \\
\bottomrule
\end{tabular}
\end{table}

These results suggest that approximate NN search can further improve the scalability of \textit{Projection}, even when accounting for the additional cost of maintaining a dynamic index. A more effective use of sub-linear NN indexes could be enabled by decoupling the static graph representation from the evolving partial solution, such that action embeddings remain fixed throughout an episode. This would eliminate frequent index reconstruction and is a promising direction for future work, as discussed in the revised Limitations section.

\section{Extended generalization study}
\label{chap:extended_generalization}
This section extends the generalization study by reporting additional statistical metrics in Tables~\ref{tab:generalization_extended1}–\ref{tab:generalization_extended4}. These include the mean performance, asymmetric spread estimates based on percentiles (lower spread: mean − 16th percentile; upper spread: 84th percentile − mean), Shapiro–Wilk normality test results, and bootstrapped confidence intervals (BCIs) of the normalized scores. These metrics provide a more reliable and robust estimate of performance across benchmarks and training strategies.
The observed variability reflects differences in policy initialization, scenario splits, and the retraining of the GAE, which is performed independently for each training strategy. 

In addition, Tables ~\ref{tab:generalization_extended1_nomax}–\ref{tab:generalization_extended4_nomax} provide further analysis logs with the complete evaluation setting without the best selection per environment, aggregating the results for all five runs to highlight the differences induced by the evaluation protocol when this selection step is removed, showcasing that the ranking and the conclusions among parameterization strategies remain consistent.

\begin{table}[ht]
\centering
\footnotesize
\setlength{\tabcolsep}{4pt}
\renewcommand{\arraystretch}{1.1}
\begin{tabular}{llcccccc}
\toprule
Regime & Metric & P-Discrete & P-Discrete-M & G-Discrete & G-Discrete-M & Iterative & Projection (ours) \\
\midrule
\multicolumn{8}{c}{\textbf{TSP}} \\
\midrule
\multirow{5}{*}{S} & \textit{Mean} & \textbf{0.02} & \textbf{0.51} & \textbf{0.00} & \textbf{0.51} & \textbf{0.92} & \textbf{0.79} \\
 & \textit{Std} & -0.02/+-0.02 & -0.06/+0.06 & -0.00/+0.00 & -0.06/+0.05 & -0.26/+0.28 & -0.18/+0.18 \\
 & \textit{IQM} & 0.00 & 0.51 & 0.00 & 0.51 & 0.91 & 0.79 \\
 & \textit{BCI} & [0.01, 0.03] & [0.50, 0.51] & [0.00, 0.00] & [0.51, 0.52] & [0.90, 0.94] & [0.78, 0.81] \\
 & \textit{Normal} & $ \times (p=0.00) $ & $ \times (p=0.01) $ & $ \checkmark (p=1.00) $ & $ \times (p=0.00) $ & $ \times (p=0.00) $ & $ \times (p=0.00) $ \\
\midrule
\multirow{5}{*}{L} & \textit{Mean} & \textbf{0.00} & \textbf{0.48} & \textbf{0.00} & \textbf{0.51} & \textbf{0.97} & \textbf{0.62} \\
 & \textit{Std} & -0.00/+0.00 & -0.07/+0.06 & -0.00/+-0.00 & -0.06/+0.06 & -0.26/+0.25 & -0.11/+0.12 \\
 & \textit{IQM} & 0.00 & 0.48 & 0.00 & 0.51 & 0.98 & 0.61 \\
 & \textit{BCI} & [0.00, 0.00] & [0.48, 0.49] & [0.00, 0.00] & [0.50, 0.51] & [0.95, 0.99] & [0.61, 0.63] \\
 & \textit{Normal} & $ \checkmark (p=1.00) $ & $ \times (p=0.00) $ & $ \times (p=0.00) $ & $ \times (p=0.00) $ & $ \times (p=0.00) $ & $ \times (p=0.00) $ \\
\midrule
\multirow{5}{*}{M} & \textit{Mean} & \textbf{0.00} & \textbf{0.48} & \textbf{0.00} & \textbf{0.51} & \textbf{0.86} & \textbf{0.76} \\
 & \textit{Std} & -0.00/+0.00 & -0.07/+0.07 & -0.00/+0.00 & -0.06/+0.06 & -0.32/+0.39 & -0.22/+0.20 \\
 & \textit{IQM} & 0.00 & 0.48 & 0.00 & 0.51 & 0.80 & 0.76 \\
 & \textit{BCI} & [0.00, 0.00] & [0.47, 0.49] & [0.00, 0.00] & [0.51, 0.52] & [0.83, 0.89] & [0.74, 0.78] \\
 & \textit{Normal} & $ \checkmark (p=1.00) $ & $ \times (p=0.01) $ & $ \checkmark (p=1.00) $ & $ \times (p=0.00) $ & $ \times (p=0.00) $ & $ \times (p=0.00) $ \\
\midrule
\multirow{5}{*}{V} & \textit{Mean} & \textbf{0.00} & \textbf{0.49} & \textbf{0.00} & \textbf{0.51} & \textbf{1.01} & \textbf{0.73} \\
 & \textit{Std} & -0.00/+0.00 & -0.07/+0.06 & -0.00/+0.00 & -0.05/+0.04 & -0.26/+0.26 & -0.17/+0.16 \\
 & \textit{IQM} & 0.00 & 0.48 & 0.00 & 0.51 & 1.00 & 0.72 \\
 & \textit{BCI} & [0.00, 0.00] & [0.48, 0.49] & [0.00, 0.00] & [0.51, 0.52] & [0.98, 1.03] & [0.71, 0.74] \\
 & \textit{Normal} & $ \checkmark (p=1.00) $ & $ \times (p=0.00) $ & $ \checkmark (p=1.00) $ & $ \times (p=0.00) $ & \checkmark (p=0.07) & $ \times (p=0.00) $ \\
\midrule
\multicolumn{8}{c}{\textbf{MinVertex}} \\
\midrule
\multirow{5}{*}{S} & \textit{Mean} & \textbf{0.04} & \textbf{0.40} & \textbf{0.00} & \textbf{0.35} & \textbf{0.55} & \textbf{0.68} \\
 & \textit{Std} & -0.04/+-0.04 & -0.15/+0.16 & -0.00/+-0.00 & -0.18/+0.15 & -0.55/+0.45 & -0.30/+0.32 \\
 & \textit{IQM} & 0.00 & 0.39 & 0.00 & 0.33 & 0.40 & 0.70 \\
 & \textit{BCI} & [0.02, 0.05] & [0.39, 0.41] & [0.00, 0.01] & [0.33, 0.36] & [0.51, 0.59] & [0.66, 0.71] \\
 & \textit{Normal} & $ \times (p=0.00) $ & $ \times (p=0.00) $ & $ \times (p=0.00) $ & $ \times (p=0.00) $ & $ \times (p=0.00) $ & $ \times (p=0.00) $ \\
\midrule
\multirow{5}{*}{L} & \textit{Mean} & \textbf{0.09} & \textbf{0.41} & \textbf{0.24} & \textbf{0.39} & \textbf{0.15} & \textbf{0.17} \\
 & \textit{Std} & -0.09/+-0.07 & -0.16/+0.17 & -0.24/+0.33 & -0.16/+0.17 & -0.15/+0.32 & -0.17/+0.33 \\
 & \textit{IQM} & 0.00 & 0.40 & 0.13 & 0.36 & 0.00 & 0.04 \\
 & \textit{BCI} & [0.07, 0.11] & [0.40, 0.42] & [0.22, 0.27] & [0.37, 0.40] & [0.12, 0.18] & [0.15, 0.20] \\
 & \textit{Normal} & $ \times (p=0.00) $ & $ \times (p=0.00) $ & $ \times (p=0.00) $ & $ \times (p=0.00) $ & $ \times (p=0.00) $ & $ \times (p=0.00) $ \\
\midrule
\multirow{5}{*}{M} & \textit{Mean} & \textbf{0.06} & \textbf{0.39} & \textbf{0.15} & \textbf{0.37} & \textbf{0.26} & \textbf{0.20} \\
 & \textit{Std} & -0.06/+-0.06 & -0.16/+0.17 & -0.15/+0.35 & -0.17/+0.18 & -0.26/+0.61 & -0.20/+0.55 \\
 & \textit{IQM} & 0.00 & 0.36 & 0.00 & 0.36 & 0.06 & 0.02 \\
 & \textit{BCI} & [0.05, 0.08] & [0.37, 0.41] & [0.13, 0.17] & [0.36, 0.39] & [0.23, 0.30] & [0.17, 0.24] \\
 & \textit{Normal} & $ \times (p=0.00) $ & $ \times (p=0.00) $ & $ \times (p=0.00) $ & $ \times (p=0.00) $ & $ \times (p=0.00) $ & $ \times (p=0.00) $ \\
\midrule
\multirow{5}{*}{V} & \textit{Mean} & \textbf{0.27} & \textbf{0.42} & \textbf{0.18} & \textbf{0.40} & \textbf{0.35} & \textbf{0.17} \\
 & \textit{Std} & -0.27/+0.28 & -0.15/+0.18 & -0.18/+0.37 & -0.15/+0.16 & -0.35/+0.54 & -0.17/+0.33 \\
 & \textit{IQM} & 0.17 & 0.40 & 0.03 & 0.39 & 0.16 & 0.03 \\
 & \textit{BCI} & [0.25, 0.30] & [0.40, 0.43] & [0.15, 0.21] & [0.38, 0.41] & [0.31, 0.39] & [0.14, 0.19] \\
 & \textit{Normal} & $ \times (p=0.00) $ & $ \times (p=0.00) $ & $ \times (p=0.00) $ & $ \times (p=0.00) $ & $ \times (p=0.00) $ & $ \times (p=0.00) $ \\
\midrule
\bottomrule
\end{tabular}
\caption{Extended generalization results with confidence intervals and Shapiro-Wilk normality checks for TSP and MinVertex.}
\label{tab:generalization_extended1}
\end{table}

\begin{table}[ht]
\centering
\footnotesize
\setlength{\tabcolsep}{4pt}
\renewcommand{\arraystretch}{1.1}
\begin{tabular}{llcccccc}
\toprule
Regime & Metric & P-Discrete & P-Discrete-M & G-Discrete & G-Discrete-M & Iterative & Projection (ours) \\
\midrule
\multicolumn{8}{c}{\textbf{MaxCut}} \\
\midrule
\multirow{5}{*}{S} & \textit{Mean} & \textbf{0.90} & \textbf{0.91} & \textbf{0.20} & \textbf{0.20} & \textbf{0.91} & \textbf{0.92} \\
 & \textit{Std} & -0.02/+0.02 & -0.02/+0.02 & -0.20/+0.22 & -0.20/+0.19 & -0.06/+0.05 & -0.05/+0.04 \\
 & \textit{IQM} & 0.91 & 0.91 & 0.08 & 0.08 & 0.92 & 0.93 \\
 & \textit{BCI} & [0.90, 0.90] & [0.91, 0.91] & [0.18, 0.23] & [0.18, 0.22] & [0.91, 0.92] & [0.92, 0.93] \\
 & \textit{Normal} & $ \times (p=0.00) $ & $ \times (p=0.00) $ & $ \times (p=0.00) $ & $ \times (p=0.00) $ & $ \times (p=0.00) $ & $ \times (p=0.00) $ \\
\midrule
\multirow{5}{*}{L} & \textit{Mean} & \textbf{0.27} & \textbf{0.88} & \textbf{0.90} & \textbf{0.91} & \textbf{0.48} & \textbf{0.91} \\
 & \textit{Std} & -0.27/+0.60 & -0.04/+0.04 & -0.02/+0.02 & -0.02/+0.02 & -0.48/+0.45 & -0.03/+0.03 \\
 & \textit{IQM} & 0.07 & 0.88 & 0.90 & 0.91 & 0.33 & 0.91 \\
 & \textit{BCI} & [0.24, 0.31] & [0.87, 0.88] & [0.90, 0.90] & [0.90, 0.91] & [0.45, 0.52] & [0.91, 0.91] \\
 & \textit{Normal} & $ \times (p=0.00) $ & $ \times (p=0.00) $ & $ \times (p=0.00) $ & $ \times (p=0.00) $ & $ \times (p=0.00) $ & $ \times (p=0.00) $ \\
\midrule
\multirow{5}{*}{M} & \textit{Mean} & \textbf{0.56} & \textbf{0.88} & \textbf{0.88} & \textbf{0.89} & \textbf{0.41} & \textbf{0.94} \\
 & \textit{Std} & -0.56/+0.35 & -0.04/+0.04 & -0.03/+0.04 & -0.03/+0.03 & -0.41/+0.53 & -0.02/+0.02 \\
 & \textit{IQM} & 0.44 & 0.89 & 0.90 & 0.90 & 0.23 & 0.95 \\
 & \textit{BCI} & [0.52, 0.59] & [0.88, 0.88] & [0.87, 0.88] & [0.89, 0.89] & [0.38, 0.45] & [0.94, 0.94] \\
 & \textit{Normal} & $ \times (p=0.00) $ & $ \times (p=0.00) $ & $ \times (p=0.00) $ & $ \times (p=0.00) $ & $ \times (p=0.00) $ & $ \times (p=0.00) $ \\
\midrule
\multirow{5}{*}{V} & \textit{Mean} & \textbf{0.69} & \textbf{0.87} & \textbf{0.87} & \textbf{0.90} & \textbf{0.39} & \textbf{0.93} \\
 & \textit{Std} & -0.69/+0.22 & -0.04/+0.04 & -0.02/+0.04 & -0.02/+0.02 & -0.39/+0.53 & -0.02/+0.02 \\
 & \textit{IQM} & 0.88 & 0.88 & 0.89 & 0.90 & 0.21 & 0.94 \\
 & \textit{BCI} & [0.66, 0.73] & [0.87, 0.88] & [0.87, 0.88] & [0.89, 0.90] & [0.35, 0.43] & [0.93, 0.93] \\
 & \textit{Normal} & $ \times (p=0.00) $ & $ \times (p=0.00) $ & $ \times (p=0.00) $ & $ \times (p=0.00) $ & $ \times (p=0.00) $ & $ \times (p=0.00) $ \\
\midrule
\multicolumn{8}{c}{\textbf{Placement}} \\
\midrule
\multirow{5}{*}{S} & \textit{Mean} & \textbf{0.64} & \textbf{0.71} & \textbf{0.44} & \textbf{0.76} & \textbf{0.28} & \textbf{0.79} \\
 & \textit{Std} & -0.21/+0.22 & -0.15/+0.15 & -0.39/+0.39 & -0.23/+0.22 & -0.26/+0.26 & -0.21/+0.20 \\
 & \textit{IQM} & 0.67 & 0.73 & 0.43 & 0.79 & 0.26 & 0.80 \\
 & \textit{BCI} & [0.62, 0.66] & [0.70, 0.73] & [0.42, 0.47] & [0.73, 0.78] & [0.26, 0.31] & [0.77, 0.81] \\
 & \textit{Normal} & $ \times (p=0.00) $ & $ \times (p=0.00) $ & $ \times (p=0.00) $ & $ \times (p=0.00) $ & $ \times (p=0.00) $ & $ \times (p=0.01) $ \\
\midrule
\multirow{5}{*}{L} & \textit{Mean} & \textbf{0.58} & \textbf{0.70} & \textbf{0.63} & \textbf{0.73} & \textbf{0.26} & \textbf{0.76} \\
 & \textit{Std} & -0.17/+0.16 & -0.12/+0.14 & -0.19/+0.21 & -0.16/+0.16 & -0.24/+0.26 & -0.18/+0.17 \\
 & \textit{IQM} & 0.58 & 0.71 & 0.66 & 0.74 & 0.21 & 0.77 \\
 & \textit{BCI} & [0.57, 0.59] & [0.69, 0.71] & [0.61, 0.65] & [0.72, 0.75] & [0.24, 0.28] & [0.75, 0.78] \\
 & \textit{Normal} & \checkmark (p=0.09) & $ \times (p=0.00) $ & $ \times (p=0.00) $ & $ \times (p=0.00) $ & $ \times (p=0.00) $ & \checkmark (p=0.29) \\
\midrule
\multirow{5}{*}{M} & \textit{Mean} & \textbf{0.57} & \textbf{0.73} & \textbf{0.52} & \textbf{0.75} & \textbf{0.12} & \textbf{0.78} \\
 & \textit{Std} & -0.21/+0.21 & -0.14/+0.14 & -0.47/+0.31 & -0.17/+0.18 & -0.12/+0.17 & -0.22/+0.21 \\
 & \textit{IQM} & 0.59 & 0.74 & 0.59 & 0.76 & 0.02 & 0.79 \\
 & \textit{BCI} & [0.55, 0.59] & [0.72, 0.74] & [0.49, 0.55] & [0.73, 0.77] & [0.10, 0.14] & [0.76, 0.80] \\
 & \textit{Normal} & $ \times (p=0.00) $ & $ \times (p=0.00) $ & $ \times (p=0.00) $ & $ \times (p=0.00) $ & $ \times (p=0.00) $ & $ \times (p=0.01) $ \\
\midrule
\multirow{5}{*}{V} & \textit{Mean} & \textbf{0.60} & \textbf{0.71} & \textbf{0.67} & \textbf{0.72} & \textbf{0.35} & \textbf{0.78} \\
 & \textit{Std} & -0.16/+0.17 & -0.13/+0.14 & -0.21/+0.22 & -0.14/+0.14 & -0.34/+0.37 & -0.19/+0.17 \\
 & \textit{IQM} & 0.60 & 0.72 & 0.69 & 0.72 & 0.30 & 0.79 \\
 & \textit{BCI} & [0.58, 0.62] & [0.69, 0.72] & [0.65, 0.69] & [0.70, 0.73] & [0.32, 0.38] & [0.77, 0.80] \\
 & \textit{Normal} & \checkmark (p=0.30) & $ \times (p=0.00) $ & $ \times (p=0.00) $ & $ \times (p=0.03) $ & $ \times (p=0.00) $ & \checkmark (p=0.09) \\
\bottomrule
\end{tabular}
\caption{Extended generalization results with confidence intervals and Shapiro-Wilk normality checks for MaxCut and Placement.}
\label{tab:generalization_extended2}
\end{table}

\begin{table}[ht]
\centering
\footnotesize
\setlength{\tabcolsep}{4pt}
\renewcommand{\arraystretch}{1.1}
\begin{tabular}{llcccccc}
\toprule
Regime & Metric & P-Discrete & P-Discrete-M & G-Discrete & G-Discrete-M & Iterative & Projection (ours) \\
\midrule
\multicolumn{8}{c}{\textbf{Cyber-Path}} \\
\midrule
\multirow{5}{*}{S} & \textit{Mean} & \textbf{0.21} & \textbf{0.62} & \textbf{0.19} & \textbf{0.47} & \textbf{0.36} & \textbf{0.61} \\
 & \textit{Std} & -0.10/+0.10 & -0.12/+0.13 & -0.09/+0.08 & -0.17/+0.18 & -0.23/+0.20 & -0.11/+0.11 \\
 & \textit{IQM} & 0.19 & 0.61 & 0.18 & 0.45 & 0.35 & 0.61 \\
 & \textit{BCI} & [0.20, 0.22] & [0.61, 0.63] & [0.19, 0.20] & [0.45, 0.48] & [0.34, 0.38] & [0.60, 0.62] \\
 & \textit{Normal} & $ \times (p=0.00) $ & $ \times (p=0.03) $ & $ \times (p=0.00) $ & $ \times (p=0.00) $ & $ \times (p=0.00) $ & \checkmark (p=0.23) \\
\midrule
\multirow{5}{*}{L} & \textit{Mean} & \textbf{0.21} & \textbf{0.61} & \textbf{0.21} & \textbf{0.52} & \textbf{0.30} & \textbf{0.61} \\
 & \textit{Std} & -0.09/+0.09 & -0.11/+0.14 & -0.10/+0.09 & -0.15/+0.15 & -0.22/+0.39 & -0.11/+0.11 \\
 & \textit{IQM} & 0.19 & 0.61 & 0.18 & 0.51 & 0.20 & 0.61 \\
 & \textit{BCI} & [0.20, 0.22] & [0.60, 0.62] & [0.20, 0.21] & [0.50, 0.53] & [0.27, 0.32] & [0.60, 0.62] \\
 & \textit{Normal} & $ \times (p=0.00) $ & $ \times (p=0.00) $ & $ \times (p=0.00) $ & \checkmark (p=0.13) & $ \times (p=0.00) $ & \checkmark (p=0.09) \\
\midrule
\multirow{5}{*}{M} & \textit{Mean} & \textbf{0.21} & \textbf{0.64} & \textbf{0.20} & \textbf{0.47} & \textbf{0.31} & \textbf{0.64} \\
 & \textit{Std} & -0.10/+0.09 & -0.11/+0.11 & -0.09/+0.08 & -0.17/+0.17 & -0.20/+0.29 & -0.09/+0.10 \\
 & \textit{IQM} & 0.19 & 0.64 & 0.19 & 0.46 & 0.25 & 0.64 \\
 & \textit{BCI} & [0.20, 0.21] & [0.63, 0.65] & [0.19, 0.20] & [0.45, 0.48] & [0.29, 0.33] & [0.63, 0.65] \\
 & \textit{Normal} & $ \times (p=0.00) $ & $ \times (p=0.01) $ & $ \times (p=0.00) $ & $ \times (p=0.00) $ & $ \times (p=0.00) $ & $ \times (p=0.00) $ \\
\midrule
\multirow{5}{*}{V} & \textit{Mean} & \textbf{0.20} & \textbf{0.68} & \textbf{0.19} & \textbf{0.51} & \textbf{0.28} & \textbf{0.68} \\
 & \textit{Std} & -0.10/+0.11 & -0.13/+0.13 & -0.09/+0.10 & -0.18/+0.15 & -0.12/+0.12 & -0.10/+0.10 \\
 & \textit{IQM} & 0.18 & 0.68 & 0.18 & 0.51 & 0.25 & 0.68 \\
 & \textit{BCI} & [0.19, 0.21] & [0.67, 0.69] & [0.18, 0.20] & [0.49, 0.52] & [0.27, 0.29] & [0.67, 0.69] \\
 & \textit{Normal} & $ \times (p=0.00) $ & \checkmark (p=0.07) & $ \times (p=0.00) $ & $ \times (p=0.01) $ & $ \times (p=0.00) $ & $ \times (p=0.00) $ \\
\midrule
\multicolumn{8}{c}{\textbf{OSPF}} \\
\midrule
\multirow{5}{*}{S} & \textit{Mean} & \textbf{0.24} & \textbf{0.86} & \textbf{0.37} & \textbf{0.50} & \textbf{0.92} & \textbf{0.94} \\
 & \textit{Std} & -0.24/+0.37 & -0.07/+0.09 & -0.37/+0.30 & -0.38/+0.33 & -0.09/+0.08 & -0.06/+0.06 \\
 & \textit{IQM} & 0.08 & 0.86 & 0.37 & 0.51 & 0.94 & 0.95 \\
 & \textit{BCI} & [0.21, 0.26] & [0.85, 0.87] & [0.35, 0.40] & [0.47, 0.52] & [0.91, 0.93] & [0.93, 0.94] \\
 & \textit{Normal} & $ \times (p=0.00) $ & $ \times (p=0.00) $ & $ \times (p=0.00) $ & $ \times (p=0.00) $ & $ \times (p=0.00) $ & $ \times (p=0.00) $ \\
\midrule
\multirow{5}{*}{L} & \textit{Mean} & \textbf{0.16} & \textbf{0.84} & \textbf{0.33} & \textbf{0.73} & \textbf{0.90} & \textbf{0.89} \\
 & \textit{Std} & -0.16/+0.24 & -0.09/+0.11 & -0.33/+0.33 & -0.19/+0.18 & -0.09/+0.10 & -0.07/+0.11 \\
 & \textit{IQM} & 0.03 & 0.85 & 0.19 & 0.75 & 0.91 & 0.89 \\
 & \textit{BCI} & [0.14, 0.18] & [0.83, 0.85] & [0.30, 0.35] & [0.71, 0.74] & [0.89, 0.91] & [0.88, 0.90] \\
 & \textit{Normal} & $ \times (p=0.00) $ & $ \times (p=0.00) $ & $ \times (p=0.00) $ & $ \times (p=0.00) $ & $ \times (p=0.00) $ & $ \times (p=0.00) $ \\
\midrule
\multirow{5}{*}{M} & \textit{Mean} & \textbf{0.27} & \textbf{0.84} & \textbf{0.33} & \textbf{0.56} & \textbf{0.91} & \textbf{0.92} \\
 & \textit{Std} & -0.27/+0.41 & -0.10/+0.11 & -0.33/+0.33 & -0.33/+0.26 & -0.09/+0.09 & -0.07/+0.08 \\
 & \textit{IQM} & 0.12 & 0.86 & 0.20 & 0.59 & 0.93 & 0.92 \\
 & \textit{BCI} & [0.24, 0.30] & [0.83, 0.86] & [0.31, 0.36] & [0.54, 0.59] & [0.90, 0.92] & [0.91, 0.92] \\
 & \textit{Normal} & $ \times (p=0.00) $ & $ \times (p=0.00) $ & $ \times (p=0.00) $ & $ \times (p=0.00) $ & $ \times (p=0.00) $ & $ \times (p=0.00) $ \\
\midrule
\multirow{5}{*}{V} & \textit{Mean} & \textbf{0.14} & \textbf{0.80} & \textbf{0.58} & \textbf{0.75} & \textbf{0.91} & \textbf{0.90} \\
 & \textit{Std} & -0.14/+0.30 & -0.12/+0.12 & -0.28/+0.25 & -0.16/+0.15 & -0.08/+0.09 & -0.07/+0.10 \\
 & \textit{IQM} & 0.01 & 0.82 & 0.60 & 0.76 & 0.93 & 0.91 \\
 & \textit{BCI} & [0.12, 0.16] & [0.78, 0.81] & [0.56, 0.61] & [0.73, 0.76] & [0.90, 0.92] & [0.90, 0.91] \\
 & \textit{Normal} & $ \times (p=0.00) $ & $ \times (p=0.00) $ & $ \times (p=0.00) $ & $ \times (p=0.00) $ & $ \times (p=0.00) $ & $ \times (p=0.00) $ \\
\bottomrule
\end{tabular}
\caption{Extended generalization results with confidence intervals and Shapiro-Wilk normality checks for Cyber-Path and OSPF.}
\label{tab:generalization_extended3}
\end{table}

\begin{table}[ht]
\centering
\footnotesize
\setlength{\tabcolsep}{4pt}
\renewcommand{\arraystretch}{1.1}
\begin{tabular}{llcccccc}
\toprule
Regime & Metric & P-Discrete & P-Discrete-M & G-Discrete & G-Discrete-M & Iterative & Projection (ours) \\
\midrule
\multicolumn{8}{c}{\textbf{Traffic}} \\
\midrule
\multirow{5}{*}{S} & \textit{Mean} & \textbf{0.24} & \textbf{0.82} & \textbf{0.30} & \textbf{0.64} & \textbf{0.63} & \textbf{0.77} \\
 & \textit{Std} & -0.24/+0.31 & -0.07/+0.07 & -0.28/+0.33 & -0.25/+0.20 & -0.26/+0.25 & -0.11/+0.13 \\
 & \textit{IQM} & 0.12 & 0.81 & 0.24 & 0.69 & 0.67 & 0.79 \\
 & \textit{BCI} & [0.22, 0.27] & [0.81, 0.82] & [0.28, 0.33] & [0.62, 0.66] & [0.60, 0.65] & [0.76, 0.79] \\
 & \textit{Normal} & $ \times (p=0.00) $ & $ \times (p=0.00) $ & $ \times (p=0.00) $ & $ \times (p=0.00) $ & $ \times (p=0.00) $ & $ \times (p=0.00) $ \\
\midrule
\multirow{5}{*}{L} & \textit{Mean} & \textbf{0.28} & \textbf{0.81} & \textbf{0.49} & \textbf{0.79} & \textbf{0.44} & \textbf{0.83} \\
 & \textit{Std} & -0.28/+0.28 & -0.08/+0.08 & -0.15/+0.17 & -0.09/+0.08 & -0.29/+0.31 & -0.08/+0.08 \\
 & \textit{IQM} & 0.17 & 0.81 & 0.49 & 0.79 & 0.49 & 0.83 \\
 & \textit{BCI} & [0.26, 0.31] & [0.80, 0.82] & [0.48, 0.51] & [0.78, 0.80] & [0.41, 0.47] & [0.82, 0.84] \\
 & \textit{Normal} & $ \times (p=0.00) $ & $ \times (p=0.00) $ & \checkmark (p=0.05) & $ \times (p=0.00) $ & $ \times (p=0.00) $ & $ \times (p=0.00) $ \\
\midrule
\multirow{5}{*}{M} & \textit{Mean} & \textbf{0.45} & \textbf{0.80} & \textbf{0.55} & \textbf{0.64} & \textbf{0.52} & \textbf{0.85} \\
 & \textit{Std} & -0.37/+0.26 & -0.09/+0.09 & -0.26/+0.23 & -0.33/+0.20 & -0.25/+0.24 & -0.08/+0.08 \\
 & \textit{IQM} & 0.49 & 0.80 & 0.57 & 0.72 & 0.55 & 0.85 \\
 & \textit{BCI} & [0.43, 0.48] & [0.79, 0.81] & [0.53, 0.57] & [0.62, 0.66] & [0.49, 0.54] & [0.84, 0.86] \\
 & \textit{Normal} & $ \times (p=0.00) $ & $ \times (p=0.00) $ & $ \times (p=0.00) $ & $ \times (p=0.00) $ & $ \times (p=0.00) $ & $ \times (p=0.00) $ \\
\midrule
\multirow{5}{*}{V} & \textit{Mean} & \textbf{0.37} & \textbf{0.82} & \textbf{0.55} & \textbf{0.78} & \textbf{0.58} & \textbf{0.85} \\
 & \textit{Std} & -0.37/+0.30 & -0.08/+0.07 & -0.23/+0.23 & -0.09/+0.09 & -0.25/+0.25 & -0.08/+0.08 \\
 & \textit{IQM} & 0.38 & 0.81 & 0.58 & 0.78 & 0.61 & 0.84 \\
 & \textit{BCI} & [0.35, 0.40] & [0.81, 0.82] & [0.53, 0.58] & [0.77, 0.79] & [0.55, 0.61] & [0.84, 0.85] \\
 & \textit{Normal} & $ \times (p=0.00) $ & $ \times (p=0.00) $ & $ \times (p=0.00) $ & $ \times (p=0.00) $ & $ \times (p=0.00) $ & $ \times (p=0.00) $ \\
\bottomrule
\end{tabular}
\caption{Extended generalization results with confidence intervals and Shapiro-Wilk normality checks for Traffic.}
\label{tab:generalization_extended4}
\end{table}

\begin{table}[ht]
\centering
\footnotesize
\setlength{\tabcolsep}{4pt}
\renewcommand{\arraystretch}{1.1}
\begin{tabular}{llcccccc}
\toprule
Regime & Metric & P-Discrete & P-Discrete-M & G-Discrete & G-Discrete-M & Iterative & Projection (ours) \\
\midrule
\multicolumn{8}{c}{\textbf{TSP}} \\
\midrule
\multirow{5}{*}{S} & \textit{Mean} & \textbf{0.01} & \textbf{0.43} & \textbf{0.00} & \textbf{0.43} & \textbf{0.87} & \textbf{0.71} \\
 & \textit{Std} & -0.01/+-0.01 & -0.08/+0.08 & -0.00/+0.00 & -0.08/+0.08 & -0.27/+0.29 & -0.19/+0.19 \\
 & \textit{IQM} & 0.00 & 0.43 & 0.00 & 0.43 & 0.86 & 0.71 \\
 & \textit{BCI} & [0.01, 0.02] & [0.43, 0.43] & [0.00, 0.00] & [0.43, 0.43] & [0.86, 0.88] & [0.70, 0.72] \\
 & \textit{Normal} & $ \times (p=0.00) $ & $ \times (p=0.00) $ & $ \checkmark (p=1.00) $ & $ \times (p=0.00) $ & $ \times (p=0.00) $ & $ \times (p=0.00) $ \\
\midrule
\multirow{5}{*}{L} & \textit{Mean} & \textbf{0.00} & \textbf{0.43} & \textbf{0.00} & \textbf{0.43} & \textbf{0.92} & \textbf{0.52} \\
 & \textit{Std} & -0.00/+0.00 & -0.08/+0.07 & -0.00/+-0.00 & -0.07/+0.07 & -0.27/+0.26 & -0.13/+0.14 \\
 & \textit{IQM} & 0.00 & 0.43 & 0.00 & 0.43 & 0.93 & 0.51 \\
 & \textit{BCI} & [0.00, 0.00] & [0.42, 0.43] & [0.00, 0.00] & [0.42, 0.43] & [0.91, 0.93] & [0.51, 0.52] \\
 & \textit{Normal} & $ \checkmark (p=1.00) $ & $ \times (p=0.00) $ & $ \times (p=0.00) $ & $ \times (p=0.00) $ & $ \times (p=0.00) $ & $ \times (p=0.00) $ \\
\midrule
\multirow{5}{*}{M} & \textit{Mean} & \textbf{0.00} & \textbf{0.43} & \textbf{0.00} & \textbf{0.43} & \textbf{0.81} & \textbf{0.67} \\
 & \textit{Std} & -0.00/+0.00 & -0.07/+0.07 & -0.00/+0.00 & -0.08/+0.07 & -0.34/+0.42 & -0.23/+0.21 \\
 & \textit{IQM} & 0.00 & 0.43 & 0.00 & 0.43 & 0.75 & 0.68 \\
 & \textit{BCI} & [0.00, 0.00] & [0.42, 0.43] & [0.00, 0.00] & [0.42, 0.43] & [0.79, 0.82] & [0.66, 0.68] \\
 & \textit{Normal} & $ \checkmark (p=1.00) $ & $ \times (p=0.00) $ & $ \checkmark (p=1.00) $ & $ \times (p=0.00) $ & $ \times (p=0.00) $ & $ \times (p=0.00) $ \\
\midrule
\multirow{5}{*}{V} & \textit{Mean} & \textbf{0.00} & \textbf{0.43} & \textbf{0.00} & \textbf{0.43} & \textbf{0.96} & \textbf{0.64} \\
 & \textit{Std} & -0.00/+0.00 & -0.08/+0.08 & -0.00/+0.00 & -0.07/+0.08 & -0.28/+0.27 & -0.20/+0.19 \\
 & \textit{IQM} & 0.00 & 0.43 & 0.00 & 0.43 & 0.96 & 0.64 \\
 & \textit{BCI} & [0.00, 0.00] & [0.42, 0.43] & [0.00, 0.00] & [0.43, 0.43] & [0.95, 0.97] & [0.63, 0.64] \\
 & \textit{Normal} & $ \checkmark (p=1.00) $ & $ \times (p=0.00) $ & $ \checkmark (p=1.00) $ & $ \times (p=0.00) $ & $ \times (p=0.00) $ & $ \times (p=0.00) $ \\
\midrule
\multicolumn{8}{c}{\textbf{MinVertex}} \\
\midrule
\multirow{5}{*}{S} & \textit{Mean} & \textbf{0.01} & \textbf{0.26} & \textbf{0.00} & \textbf{0.26} & \textbf{0.55} & \textbf{0.57} \\
 & \textit{Std} & -0.01/+-0.01 & -0.15/+0.14 & -0.00/+-0.00 & -0.14/+0.14 & -0.55/+0.45 & -0.32/+0.31 \\
 & \textit{IQM} & 0.00 & 0.23 & 0.00 & 0.23 & 0.40 & 0.56 \\
 & \textit{BCI} & [0.01, 0.02] & [0.26, 0.27] & [0.00, 0.00] & [0.26, 0.27] & [0.53, 0.57] & [0.56, 0.58] \\
 & \textit{Normal} & $ \times (p=0.00) $ & $ \times (p=0.00) $ & $ \times (p=0.00) $ & $ \times (p=0.00) $ & $ \times (p=0.00) $ & $ \times (p=0.00) $ \\
\midrule
\multirow{5}{*}{L} & \textit{Mean} & \textbf{0.03} & \textbf{0.27} & \textbf{0.10} & \textbf{0.27} & \textbf{0.15} & \textbf{0.09} \\
 & \textit{Std} & -0.03/+-0.03 & -0.14/+0.15 & -0.10/+0.23 & -0.14/+0.13 & -0.15/+0.32 & -0.09/+0.16 \\
 & \textit{IQM} & 0.00 & 0.24 & 0.00 & 0.24 & 0.00 & 0.01 \\
 & \textit{BCI} & [0.02, 0.03] & [0.26, 0.27] & [0.10, 0.11] & [0.26, 0.27] & [0.14, 0.16] & [0.09, 0.10] \\
 & \textit{Normal} & $ \times (p=0.00) $ & $ \times (p=0.00) $ & $ \times (p=0.00) $ & $ \times (p=0.00) $ & $ \times (p=0.00) $ & $ \times (p=0.00) $ \\
\midrule
\multirow{5}{*}{M} & \textit{Mean} & \textbf{0.02} & \textbf{0.27} & \textbf{0.06} & \textbf{0.27} & \textbf{0.26} & \textbf{0.13} \\
 & \textit{Std} & -0.02/+-0.02 & -0.16/+0.17 & -0.06/+-0.06 & -0.15/+0.13 & -0.26/+0.61 & -0.13/+0.32 \\
 & \textit{IQM} & 0.00 & 0.25 & 0.00 & 0.24 & 0.06 & 0.00 \\
 & \textit{BCI} & [0.02, 0.03] & [0.27, 0.28] & [0.05, 0.07] & [0.26, 0.28] & [0.25, 0.28] & [0.12, 0.14] \\
 & \textit{Normal} & $ \times (p=0.00) $ & $ \times (p=0.00) $ & $ \times (p=0.00) $ & $ \times (p=0.00) $ & $ \times (p=0.00) $ & $ \times (p=0.00) $ \\
\midrule
\multirow{5}{*}{V} & \textit{Mean} & \textbf{0.13} & \textbf{0.27} & \textbf{0.08} & \textbf{0.27} & \textbf{0.35} & \textbf{0.09} \\
 & \textit{Std} & -0.13/+0.25 & -0.14/+0.13 & -0.08/+-0.08 & -0.14/+0.13 & -0.35/+0.55 & -0.09/+0.11 \\
 & \textit{IQM} & 0.02 & 0.24 & 0.00 & 0.24 & 0.16 & 0.00 \\
 & \textit{BCI} & [0.12, 0.14] & [0.26, 0.27] & [0.07, 0.09] & [0.26, 0.27] & [0.33, 0.37] & [0.08, 0.09] \\
 & \textit{Normal} & $ \times (p=0.00) $ & $ \times (p=0.00) $ & $ \times (p=0.00) $ & $ \times (p=0.00) $ & $ \times (p=0.00) $ & $ \times (p=0.00) $ \\
\bottomrule
\end{tabular}
\caption{Extended generalization results with confidence intervals and Shapiro-Wilk normality checks for TSP and MinVertex without the best selection per environment.}
\label{tab:generalization_extended1_nomax}
\end{table}

\begin{table}[ht]
\centering
\footnotesize
\setlength{\tabcolsep}{4pt}
\renewcommand{\arraystretch}{1.1}
\begin{tabular}{llcccccc}
\toprule
Regime & Metric & P-Discrete & P-Discrete-M & G-Discrete & G-Discrete-M & Iterative & Projection (ours) \\
\midrule
\multicolumn{8}{c}{\textbf{MaxCut}} \\
\midrule
\multirow{5}{*}{S} & \textit{Mean} & \textbf{0.87} & \textbf{0.88} & \textbf{0.16} & \textbf{0.15} & \textbf{0.91} & \textbf{0.91} \\
 & \textit{Std} & -0.05/+0.04 & -0.03/+0.04 & -0.16/+0.19 & -0.15/+0.16 & -0.06/+0.05 & -0.05/+0.05 \\
 & \textit{IQM} & 0.89 & 0.89 & 0.04 & 0.05 & 0.92 & 0.92 \\
 & \textit{BCI} & [0.87, 0.87] & [0.88, 0.88] & [0.15, 0.16] & [0.15, 0.16] & [0.91, 0.91] & [0.91, 0.91] \\
 & \textit{Normal} & $ \times (p=0.00) $ & $ \times (p=0.00) $ & $ \times (p=0.00) $ & $ \times (p=0.00) $ & $ \times (p=0.00) $ & $ \times (p=0.00) $ \\
\midrule
\multirow{5}{*}{L} & \textit{Mean} & \textbf{0.24} & \textbf{0.86} & \textbf{0.85} & \textbf{0.87} & \textbf{0.48} & \textbf{0.88} \\
 & \textit{Std} & -0.24/+0.58 & -0.05/+0.05 & -0.05/+0.06 & -0.04/+0.04 & -0.48/+0.45 & -0.05/+0.05 \\
 & \textit{IQM} & 0.05 & 0.88 & 0.87 & 0.88 & 0.33 & 0.89 \\
 & \textit{BCI} & [0.23, 0.26] & [0.86, 0.87] & [0.85, 0.85] & [0.86, 0.87] & [0.47, 0.50] & [0.88, 0.88] \\
 & \textit{Normal} & $ \times (p=0.00) $ & $ \times (p=0.00) $ & $ \times (p=0.00) $ & $ \times (p=0.00) $ & $ \times (p=0.00) $ & $ \times (p=0.00) $ \\
\midrule
\multirow{5}{*}{M} & \textit{Mean} & \textbf{0.52} & \textbf{0.85} & \textbf{0.79} & \textbf{0.83} & \textbf{0.41} & \textbf{0.92} \\
 & \textit{Std} & -0.52/+0.37 & -0.07/+0.06 & -0.12/+0.10 & -0.07/+0.08 & -0.41/+0.53 & -0.03/+0.03 \\
 & \textit{IQM} & 0.39 & 0.87 & 0.84 & 0.86 & 0.23 & 0.93 \\
 & \textit{BCI} & [0.50, 0.53] & [0.85, 0.85] & [0.79, 0.80] & [0.82, 0.83] & [0.40, 0.43] & [0.92, 0.92] \\
 & \textit{Normal} & $ \times (p=0.00) $ & $ \times (p=0.00) $ & $ \times (p=0.00) $ & $ \times (p=0.00) $ & $ \times (p=0.00) $ & $ \times (p=0.00) $ \\
\midrule
\multirow{5}{*}{V} & \textit{Mean} & \textbf{0.65} & \textbf{0.84} & \textbf{0.79} & \textbf{0.84} & \textbf{0.39} & \textbf{0.91} \\
 & \textit{Std} & -0.65/+0.25 & -0.06/+0.06 & -0.08/+0.10 & -0.06/+0.06 & -0.39/+0.53 & -0.04/+0.04 \\
 & \textit{IQM} & 0.82 & 0.86 & 0.84 & 0.87 & 0.21 & 0.92 \\
 & \textit{BCI} & [0.63, 0.67] & [0.83, 0.84] & [0.79, 0.80] & [0.84, 0.85] & [0.37, 0.41] & [0.90, 0.91] \\
 & \textit{Normal} & $ \times (p=0.00) $ & $ \times (p=0.00) $ & $ \times (p=0.00) $ & $ \times (p=0.00) $ & $ \times (p=0.00) $ & $ \times (p=0.00) $ \\
\midrule
\multicolumn{8}{c}{\textbf{Placement}} \\
\midrule
\multirow{5}{*}{S} & \textit{Mean} & \textbf{0.51} & \textbf{0.58} & \textbf{0.35} & \textbf{0.66} & \textbf{0.28} & \textbf{0.69} \\
 & \textit{Std} & -0.25/+0.24 & -0.19/+0.19 & -0.34/+0.35 & -0.28/+0.26 & -0.26/+0.26 & -0.24/+0.23 \\
 & \textit{IQM} & 0.53 & 0.59 & 0.31 & 0.70 & 0.26 & 0.70 \\
 & \textit{BCI} & [0.50, 0.52] & [0.57, 0.59] & [0.34, 0.37] & [0.65, 0.67] & [0.27, 0.29] & [0.68, 0.70] \\
 & \textit{Normal} & $ \times (p=0.00) $ & $ \times (p=0.00) $ & $ \times (p=0.00) $ & $ \times (p=0.00) $ & $ \times (p=0.00) $ & $ \times (p=0.00) $ \\
\midrule
\multirow{5}{*}{L} & \textit{Mean} & \textbf{0.43} & \textbf{0.56} & \textbf{0.49} & \textbf{0.61} & \textbf{0.26} & \textbf{0.62} \\
 & \textit{Std} & -0.20/+0.19 & -0.18/+0.18 & -0.25/+0.24 & -0.19/+0.20 & -0.24/+0.26 & -0.21/+0.20 \\
 & \textit{IQM} & 0.42 & 0.56 & 0.51 & 0.61 & 0.21 & 0.62 \\
 & \textit{BCI} & [0.42, 0.44] & [0.56, 0.57] & [0.48, 0.50] & [0.60, 0.61] & [0.25, 0.27] & [0.61, 0.63] \\
 & \textit{Normal} & $ \times (p=0.00) $ & $ \times (p=0.00) $ & $ \times (p=0.00) $ & $ \times (p=0.00) $ & $ \times (p=0.00) $ & $ \times (p=0.04) $ \\
\midrule
\multirow{5}{*}{M} & \textit{Mean} & \textbf{0.44} & \textbf{0.60} & \textbf{0.41} & \textbf{0.63} & \textbf{0.12} & \textbf{0.66} \\
 & \textit{Std} & -0.23/+0.23 & -0.19/+0.19 & -0.39/+0.30 & -0.20/+0.22 & -0.12/+0.17 & -0.25/+0.25 \\
 & \textit{IQM} & 0.44 & 0.60 & 0.43 & 0.64 & 0.02 & 0.66 \\
 & \textit{BCI} & [0.43, 0.45] & [0.59, 0.61] & [0.40, 0.42] & [0.62, 0.64] & [0.11, 0.13] & [0.65, 0.67] \\
 & \textit{Normal} & $ \times (p=0.00) $ & $ \times (p=0.00) $ & $ \times (p=0.00) $ & $ \times (p=0.00) $ & $ \times (p=0.00) $ & $ \times (p=0.00) $ \\
\midrule
\multirow{5}{*}{V} & \textit{Mean} & \textbf{0.45} & \textbf{0.57} & \textbf{0.55} & \textbf{0.58} & \textbf{0.35} & \textbf{0.65} \\
 & \textit{Std} & -0.20/+0.20 & -0.18/+0.18 & -0.25/+0.23 & -0.18/+0.19 & -0.34/+0.38 & -0.20/+0.20 \\
 & \textit{IQM} & 0.45 & 0.58 & 0.56 & 0.58 & 0.30 & 0.65 \\
 & \textit{BCI} & [0.44, 0.46] & [0.56, 0.58] & [0.54, 0.56] & [0.57, 0.59] & [0.34, 0.37] & [0.65, 0.66] \\
 & \textit{Normal} & $ \times (p=0.00) $ & $ \times (p=0.00) $ & $ \times (p=0.00) $ & $ \times (p=0.01) $ & $ \times (p=0.00) $ & $ \times (p=0.01) $ \\
\bottomrule
\end{tabular}
\caption{Extended generalization results with confidence intervals and Shapiro-Wilk normality checks for MaxCut and Placement without the best selection per environment.}
\label{tab:generalization_extended2_nomax}
\end{table}

\begin{table}[ht]
\centering
\footnotesize
\setlength{\tabcolsep}{4pt}
\renewcommand{\arraystretch}{1.1}
\begin{tabular}{llcccccc}
\toprule
Regime & Metric & P-Discrete & P-Discrete-M & G-Discrete & G-Discrete-M & Iterative & Projection (ours) \\
\midrule
\multicolumn{8}{c}{\textbf{Cyber-Path}} \\
\midrule
\multirow{5}{*}{S} & \textit{Mean} & \textbf{0.11} & \textbf{0.44} & \textbf{0.11} & \textbf{0.29} & \textbf{0.26} & \textbf{0.47} \\
 & \textit{Std} & -0.05/+0.05 & -0.19/+0.18 & -0.05/+0.05 & -0.17/+0.18 & -0.19/+0.19 & -0.15/+0.16 \\
 & \textit{IQM} & 0.08 & 0.44 & 0.08 & 0.25 & 0.21 & 0.48 \\
 & \textit{BCI} & [0.11, 0.11] & [0.43, 0.44] & [0.10, 0.11] & [0.28, 0.29] & [0.25, 0.27] & [0.46, 0.48] \\
 & \textit{Normal} & $ \times (p=0.00) $ & $ \times (p=0.00) $ & $ \times (p=0.00) $ & $ \times (p=0.00) $ & $ \times (p=0.00) $ & $ \times (p=0.00) $ \\
\midrule
\multirow{5}{*}{L} & \textit{Mean} & \textbf{0.11} & \textbf{0.43} & \textbf{0.11} & \textbf{0.32} & \textbf{0.23} & \textbf{0.46} \\
 & \textit{Std} & -0.05/+0.06 & -0.21/+0.18 & -0.05/+0.06 & -0.19/+0.18 & -0.17/+0.24 & -0.16/+0.15 \\
 & \textit{IQM} & 0.09 & 0.44 & 0.09 & 0.29 & 0.12 & 0.47 \\
 & \textit{BCI} & [0.11, 0.11] & [0.43, 0.44] & [0.11, 0.11] & [0.31, 0.33] & [0.22, 0.24] & [0.46, 0.47] \\
 & \textit{Normal} & $ \times (p=0.00) $ & $ \times (p=0.00) $ & $ \times (p=0.00) $ & $ \times (p=0.00) $ & $ \times (p=0.00) $ & $ \times (p=0.00) $ \\
\midrule
\multirow{5}{*}{M} & \textit{Mean} & \textbf{0.11} & \textbf{0.46} & \textbf{0.11} & \textbf{0.29} & \textbf{0.23} & \textbf{0.49} \\
 & \textit{Std} & -0.05/+0.05 & -0.17/+0.16 & -0.05/+0.06 & -0.18/+0.21 & -0.16/+0.23 & -0.16/+0.15 \\
 & \textit{IQM} & 0.08 & 0.47 & 0.09 & 0.26 & 0.16 & 0.51 \\
 & \textit{BCI} & [0.10, 0.11] & [0.45, 0.47] & [0.11, 0.11] & [0.29, 0.30] & [0.22, 0.24] & [0.49, 0.50] \\
 & \textit{Normal} & $ \times (p=0.00) $ & $ \times (p=0.00) $ & $ \times (p=0.00) $ & $ \times (p=0.00) $ & $ \times (p=0.00) $ & $ \times (p=0.00) $ \\
\midrule
\multirow{5}{*}{V} & \textit{Mean} & \textbf{0.11} & \textbf{0.47} & \textbf{0.10} & \textbf{0.33} & \textbf{0.20} & \textbf{0.53} \\
 & \textit{Std} & -0.05/+0.05 & -0.20/+0.19 & -0.05/+0.05 & -0.18/+0.17 & -0.10/+0.10 & -0.14/+0.15 \\
 & \textit{IQM} & 0.08 & 0.48 & 0.08 & 0.31 & 0.18 & 0.54 \\
 & \textit{BCI} & [0.10, 0.11] & [0.47, 0.48] & [0.10, 0.11] & [0.32, 0.34] & [0.19, 0.20] & [0.53, 0.54] \\
 & \textit{Normal} & $ \times (p=0.00) $ & $ \times (p=0.00) $ & $ \times (p=0.00) $ & $ \times (p=0.00) $ & $ \times (p=0.00) $ & $ \times (p=0.00) $ \\
\midrule
\multicolumn{8}{c}{\textbf{OSPF}} \\
\midrule
\multirow{5}{*}{S} & \textit{Mean} & \textbf{0.16} & \textbf{0.78} & \textbf{0.24} & \textbf{0.39} & \textbf{0.92} & \textbf{0.87} \\
 & \textit{Std} & -0.16/+0.21 & -0.12/+0.12 & -0.24/+0.25 & -0.33/+0.33 & -0.09/+0.08 & -0.12/+0.13 \\
 & \textit{IQM} & 0.04 & 0.79 & 0.19 & 0.36 & 0.94 & 0.89 \\
 & \textit{BCI} & [0.15, 0.17] & [0.77, 0.78] & [0.23, 0.25] & [0.37, 0.40] & [0.92, 0.92] & [0.86, 0.87] \\
 & \textit{Normal} & $ \times (p=0.00) $ & $ \times (p=0.00) $ & $ \times (p=0.00) $ & $ \times (p=0.00) $ & $ \times (p=0.00) $ & $ \times (p=0.00) $ \\
\midrule
\multirow{5}{*}{L} & \textit{Mean} & \textbf{0.08} & \textbf{0.77} & \textbf{0.19} & \textbf{0.57} & \textbf{0.90} & \textbf{0.79} \\
 & \textit{Std} & -0.08/+0.10 & -0.12/+0.12 & -0.19/+0.22 & -0.24/+0.23 & -0.09/+0.10 & -0.13/+0.13 \\
 & \textit{IQM} & 0.00 & 0.78 & 0.08 & 0.58 & 0.91 & 0.80 \\
 & \textit{BCI} & [0.07, 0.09] & [0.77, 0.78] & [0.18, 0.20] & [0.56, 0.58] & [0.90, 0.90] & [0.78, 0.79] \\
 & \textit{Normal} & $ \times (p=0.00) $ & $ \times (p=0.00) $ & $ \times (p=0.00) $ & $ \times (p=0.00) $ & $ \times (p=0.00) $ & $ \times (p=0.00) $ \\
\midrule
\multirow{5}{*}{M} & \textit{Mean} & \textbf{0.18} & \textbf{0.78} & \textbf{0.22} & \textbf{0.44} & \textbf{0.91} & \textbf{0.83} \\
 & \textit{Std} & -0.18/+0.27 & -0.12/+0.13 & -0.22/+0.26 & -0.30/+0.29 & -0.09/+0.09 & -0.12/+0.13 \\
 & \textit{IQM} & 0.05 & 0.80 & 0.10 & 0.43 & 0.93 & 0.85 \\
 & \textit{BCI} & [0.17, 0.19] & [0.78, 0.79] & [0.21, 0.23] & [0.43, 0.45] & [0.91, 0.92] & [0.83, 0.84] \\
 & \textit{Normal} & $ \times (p=0.00) $ & $ \times (p=0.00) $ & $ \times (p=0.00) $ & $ \times (p=0.00) $ & $ \times (p=0.00) $ & $ \times (p=0.00) $ \\
\midrule
\multirow{5}{*}{V} & \textit{Mean} & \textbf{0.08} & \textbf{0.77} & \textbf{0.42} & \textbf{0.60} & \textbf{0.91} & \textbf{0.80} \\
 & \textit{Std} & -0.08/+0.10 & -0.12/+0.13 & -0.28/+0.27 & -0.21/+0.19 & -0.08/+0.09 & -0.12/+0.13 \\
 & \textit{IQM} & 0.00 & 0.79 & 0.40 & 0.61 & 0.93 & 0.82 \\
 & \textit{BCI} & [0.07, 0.09] & [0.77, 0.78] & [0.41, 0.43] & [0.60, 0.61] & [0.91, 0.92] & [0.80, 0.81] \\
 & \textit{Normal} & $ \times (p=0.00) $ & $ \times (p=0.00) $ & $ \times (p=0.00) $ & $ \times (p=0.00) $ & $ \times (p=0.00) $ & $ \times (p=0.00) $ \\
\bottomrule
\end{tabular}
\caption{Extended generalization results with confidence intervals and Shapiro-Wilk normality checks for Cyber-Path and OSPF without the best selection per environment.}
\label{tab:generalization_extended3_nomax}
\end{table}

\begin{table}[ht]
\centering
\footnotesize
\setlength{\tabcolsep}{4pt}
\renewcommand{\arraystretch}{1.1}
\begin{tabular}{llcccccc}
\toprule
Regime & Metric & P-Discrete & P-Discrete-M & G-Discrete & G-Discrete-M & Iterative & Projection (ours) \\
\midrule
\multicolumn{8}{c}{\textbf{Traffic}} \\
\midrule
\multirow{5}{*}{S} & \textit{Mean} & \textbf{0.17} & \textbf{0.69} & \textbf{0.23} & \textbf{0.48} & \textbf{0.63} & \textbf{0.64} \\
 & \textit{Std} & -0.17/+0.23 & -0.14/+0.13 & -0.23/+0.26 & -0.31/+0.26 & -0.26/+0.25 & -0.19/+0.18 \\
 & \textit{IQM} & 0.07 & 0.70 & 0.17 & 0.51 & 0.67 & 0.67 \\
 & \textit{BCI} & [0.16, 0.18] & [0.68, 0.69] & [0.22, 0.24] & [0.47, 0.49] & [0.62, 0.64] & [0.63, 0.65] \\
 & \textit{Normal} & $ \times (p=0.00) $ & $ \times (p=0.00) $ & $ \times (p=0.00) $ & $ \times (p=0.00) $ & $ \times (p=0.00) $ & $ \times (p=0.00) $ \\
\midrule
\multirow{5}{*}{L} & \textit{Mean} & \textbf{0.15} & \textbf{0.69} & \textbf{0.28} & \textbf{0.65} & \textbf{0.44} & \textbf{0.71} \\
 & \textit{Std} & -0.15/+0.22 & -0.14/+0.13 & -0.25/+0.22 & -0.16/+0.15 & -0.29/+0.31 & -0.13/+0.13 \\
 & \textit{IQM} & 0.05 & 0.70 & 0.27 & 0.66 & 0.49 & 0.72 \\
 & \textit{BCI} & [0.14, 0.16] & [0.68, 0.69] & [0.27, 0.29] & [0.64, 0.65] & [0.43, 0.46] & [0.70, 0.72] \\
 & \textit{Normal} & $ \times (p=0.00) $ & $ \times (p=0.00) $ & $ \times (p=0.00) $ & $ \times (p=0.00) $ & $ \times (p=0.00) $ & $ \times (p=0.00) $ \\
\midrule
\multirow{5}{*}{M} & \textit{Mean} & \textbf{0.30} & \textbf{0.69} & \textbf{0.40} & \textbf{0.51} & \textbf{0.52} & \textbf{0.73} \\
 & \textit{Std} & -0.30/+0.29 & -0.14/+0.13 & -0.27/+0.26 & -0.32/+0.24 & -0.25/+0.24 & -0.13/+0.14 \\
 & \textit{IQM} & 0.27 & 0.70 & 0.40 & 0.55 & 0.55 & 0.74 \\
 & \textit{BCI} & [0.29, 0.31] & [0.68, 0.69] & [0.39, 0.41] & [0.50, 0.52] & [0.51, 0.53] & [0.72, 0.74] \\
 & \textit{Normal} & $ \times (p=0.00) $ & $ \times (p=0.00) $ & $ \times (p=0.00) $ & $ \times (p=0.00) $ & $ \times (p=0.00) $ & $ \times (p=0.00) $ \\
\midrule
\multirow{5}{*}{V} & \textit{Mean} & \textbf{0.24} & \textbf{0.69} & \textbf{0.39} & \textbf{0.64} & \textbf{0.58} & \textbf{0.73} \\
 & \textit{Std} & -0.24/+0.27 & -0.14/+0.13 & -0.27/+0.26 & -0.16/+0.15 & -0.25/+0.26 & -0.13/+0.13 \\
 & \textit{IQM} & 0.13 & 0.70 & 0.39 & 0.66 & 0.61 & 0.74 \\
 & \textit{BCI} & [0.23, 0.25] & [0.68, 0.69] & [0.37, 0.40] & [0.64, 0.65] & [0.57, 0.59] & [0.72, 0.73] \\
 & \textit{Normal} & $ \times (p=0.00) $ & $ \times (p=0.00) $ & $ \times (p=0.00) $ & $ \times (p=0.00) $ & $ \times (p=0.00) $ & $ \times (p=0.00) $ \\
\bottomrule
\end{tabular}
\caption{Extended generalization results with confidence intervals and Shapiro-Wilk normality checks for Traffic without the best selection per environment.}
\label{tab:generalization_extended4_nomax}
\end{table}

\end{document}